\documentclass{article}

\usepackage[utf8]{inputenc} 
\usepackage[T1]{fontenc}    
\usepackage{hyperref}       
\usepackage{url}            
\usepackage{booktabs}       
\usepackage{amsfonts}       
\usepackage{nicefrac}       
\usepackage{microtype}          
\usepackage{hhline}
\usepackage[table]{xcolor}
\usepackage{arydshln}
\usepackage{colortbl}
\usepackage{fullpage}
\usepackage{natbib}
\usepackage{tabularx}

\hypersetup{
    colorlinks=true,
    linkcolor=black, 
    filecolor=blue,      
    urlcolor=blue,
    citecolor=blue
}

\usepackage{amsmath, amsthm, amssymb}
\usepackage{amsfonts,bm}
\usepackage{dsfont}

\def\1{\bm{1}}

\def\eps{{\varepsilon}}

\def\rd{{\textnormal{d}}}

\def\vzero{{\bm{0}}}

\def\vmu{{\bm{\mu}}}

\def\vnu{{\bm{\nu}}}
\def\va{{\bm{a}}}

\def\vg{{\bm{g}}}
\def\vh{{\bm{h}}}

\def\vt{{\bm{t}}}
\def\vu{{\bm{u}}}
\def\vv{{\bm{v}}}
\def\vw{{\bm{w}}}
\def\vx{{\bm{x}}}
\def\vy{{\bm{y}}}

\def\mI{{\bm{I}}}

\DeclareMathAlphabet{\mathsfit}{\encodingdefault}{\sfdefault}{m}{sl}
\SetMathAlphabet{\mathsfit}{bold}{\encodingdefault}{\sfdefault}{bx}{n}

\def\gA{{\mathcal{A}}}

\def\gN{{\mathcal{N}}}

\def\gP{{\mathcal{P}}}

\DeclareMathOperator*{\argmin}{arg\,min}

\newcommand{\norm}[1]{\left\|#1\right\|}
\newcommand{\abs}[1]{\lvert#1\rvert}

\DeclareMathOperator{\poly}{\mathrm{poly}}

\newcommand{\anchorterm}[2]{%
  \hypertarget{#1}{\text{#2}}
  \expandafter\gdef\csname anchortext@#1\endcsname{\text{#2}}%
}

\DeclareRobustCommand{\termlink}[1]{%
  \hyperlink{#1}{\text{\csname anchortext@#1\endcsname}}%
}

\newcommand{\myge}[1]{\stackrel{(\text{#1})}{\ge}}
\newcommand{\myeq}[1]{\stackrel{(\text{#1})}{=}}
\newcommand{\myle}[1]{\stackrel{(\text{#1})}{\le}}
\newcommand{\mylesim}[1]{\stackrel{(\text{#1})}{\lesssim}}
\newcommand{\myapprox}[1]{\stackrel{(\text{#1})}{\approx}}

\renewcommand{\P}{\mathbb{P}}
\newcommand{\ind}{\mathds{1}}
\newcommand{\op}{\mathrm{op}}

\newcommand{\tldpi}{\widetilde{\pi}}
\newcommand{\deltacls}{\delta_{\rm close}}

\newcommand{\tldtau}{\widetilde{\tau}}

\makeatletter
    \newcommand{\anchortext@termtau}{\text{(I)}}
    \newcommand{\anchortext@challengeweight}{\text{Challenge 1}}
    \newcommand{\anchortext@termIIi}{\text{(II.i)}}
    \newcommand{\anchortext@termIIii}{\text{(II.ii)}}
\makeatother

\newcommand{\tldvpsi}{\bm{\widetilde{\psi}}}

\newcommand{\pimins}{{\pi_{\min}^*}}

\newcommand{\vdelta}{{\bm{\delta}}}

\usepackage{enumitem}
\newlist{tightenum}{enumerate}{1}
\setlist[tightenum]{
  label=\arabic*.,   
  leftmargin=13pt,
  labelindent=0pt,
  listparindent=0pt,
  itemsep=-3pt,
  topsep=-15pt,
}

\def\vZ{{\bm{Z}}}
\def\vX{{\bm{X}}}

\newcommand{\sfA}{\mathsf{A}}
\newcommand{\vV}{{\bm{V}}}
\usepackage{Configs/config}
\usepackage[capitalise,nameinlink,noabbrev]{cleveref}

\crefname{lemma}{Lemma}{Lemmas}
\Crefname{lemma}{Lemma}{Lemmas}

\crefname{theorem}{Theorem}{Theorems}
\Crefname{theorem}{Theorem}{Theorems}

\crefname{claim}{Claim}{Claims}
\Crefname{claim}{Claim}{Claims}

\crefname{proposition}{Proposition}{Propositions}
\Crefname{proposition}{Proposition}{Propositions}

\crefname{figure}{Figure}{Figures}
\Crefname{figure}{Figure}{Figures}

\crefname{section}{Section}{Sections}
\Crefname{section}{Section}{Sections}

\crefname{remark}{Remark}{Remarks}
\Crefname{remark}{Remark}{Remarks}

\crefname{assumption}{Assumption}{Assumptions}
\Crefname{assumption}{Assumption}{Assumptions}

\let\PolishL\L
\newcommand\blfootnote[1]{%
  \begingroup
  \renewcommand\thefootnote{}\footnote{#1}%
  \addtocounter{footnote}{-1}%
  \endgroup
}

\title{Global Convergence of Gradient EM for Over-Parameterized Gaussian Mixtures}

\author{
  Mo Zhou$^1$\thanks{Equal contribution.}\qquad 
  Weihang Xu$^1$\footnotemark[1]\qquad 
  Maryam Fazel$^{1,2}$\qquad 
  Simon S. Du$^1$\\
  $^1$University of Washington
  \qquad
  $^2$Amazon, Inc.\\
  \texttt{\{mozhou17,xuwh,ssdu\}@cs.washington.edu, mfazel@uw.edu}
}

\begin{document}

\maketitle

\blfootnote{%
Changes in v2:
We remove Assumptions 1 and 2 from v1 and revise and simplify the proofs in the appendix.
}

\begin{abstract}
    Learning Gaussian Mixture Models (GMMs) is a fundamental problem in statistics and machine learning, with the Expectation-Maximization (EM) algorithm and its popular variant gradient EM being arguably the most widely used algorithms in practice. In the exact-parameterized setting, where both the ground truth GMM and the learning model have the same number of components $m$, a vast line of work has aimed to establish rigorous recovery guarantees for EM. However, global convergence has only been proven for the case of $m=2$, and EM is known to fail to recover the ground truth when $m\geq 3$.
  
    In this paper, we consider the \emph{over-parameterized} setting, where the learning model uses $n>m$ components to fit an $m$-component ground truth GMM. In contrast to the exact-parameterized case, we provide a guarantee for convergence to the globally optimal solution (the ground truth) 
    for gradient EM. Specifically, for any well-separated GMMs, we prove that with only mild over-parameterization $n = \Omega(m\log m)$, randomly initialized gradient EM converges to the ground truth with polynomial time and samples. Our analysis proceeds in two stages and introduces a suite of novel tools for Gaussian Mixture analysis to study the dynamics of gradient EM and characterize the geometric landscape of the likelihood loss. This is the first global convergence and recovery result for EM or Gradient EM beyond the special case of $m=2$.
\end{abstract}

\section{Introduction}\label{Sec: intro}

Gaussian Mixture Models (GMMs) \citep{pearson1894contributions,titterington1985statistical} are widely used across machine learning applications, where their ability to model complex, multimodal data distributions makes them a fundamental tool in probabilistic
modeling.
In this paper, we consider the canonical problem of learning Gaussian mixtures with isotropic covariances. Here the ground truth distribution, a $d$-dimensional $m$-component Gaussian mixture model ($m$-GMM), is given by the probability density function $p_*$:
\begin{equation}\label{eq: ground truth GMM}
    p_*(\vx) = \sum_{i=1}^m \pi_i^* \phi(\vmu_i^*;\vx),
\end{equation}
where $\pi_i^*\in \R_+$ are the mixing weights with $\sum_{i}\pi_i^*=1$, $\vmu_i^*\in \R^d$ are the ground truth mean vectors, and $\phi(\vmu;\vx)\coloneqq (2\pi)^{-d/2}\exp(-\norm{\vx-\vmu}_2^2/2)$  is the p.d.f. of a Gaussian with mean $\vmu\in\R^d$ and covariance $\mI_d$. The goal is to recover the true distribution $p_*$ with observed data sampled from it.

\paragraph*{The EM and gradient EM algorithms} 
For learning hidden variable models such as GMMs, perhaps the most popular algorithm used in practice is the Expectation-Maximization (EM) algorithm~\citep{dempster1977maximum}. Using a $n$-GMM model defined as 
\begin{equation}\label{eq: over-parameterized GMM}
    p(\vx) = \sum_{i=1}^n \pi_i \phi(\vmu_i;\vx),
\end{equation}
the EM algorithm iteratively updates the mixing weights $\vpi\coloneqq (\pi_1, \ldots, \pi_n)^{\top}$ and model means $\vmu\coloneqq (\vmu_1^{\top},\ldots, \vmu_n^{\top})^{\top}$ to learn the ground truth distribution $p_*$. Note that the number of component Gaussians in the model ($n$) and the ground truth ($m$) can differ. In practice, a popular lightweight variant of EM, called gradient EM, is often used when vanilla EM update is too computationally expensive. In the $t^{\text{th}}$ iteration, population EM and gradient EM updates consist of two steps:

\begin{itemize}
    \item Expectation (E) step: For $
    i=1,\ldots, n$, compute the membership weight function $\psi_i: \R^d\to \R^+$, the posterior probability of $\vx\in \R^d$ being sampled from the $i^{\text{th}}$ model Gaussian:
    \[\psi_i(\vx) 
    = \Pr(\vx|i)
    =\frac{\pi_i^{(t)}\phi(\vmu_i^{(t)};\vx)}{\sum_{k=1}^n \pi_k^{(t)}\phi(\vmu_k^{(t)};\vx)}.\]
    \item Maximization (M) step: Define $Q$ function
    \[
        Q(\bm{\vpi, \vmu} |\vpi^{(t)}, \bm{\mu}^{(t)})
        = \E_{\vx\sim p_*} \left[\sum_{i=1}^n \psi_i(\vx;\vpi^{(t)},\vmu^{(t)})\left(\log\pi_i - \frac{1}{2}\norm{\vx-\vmu_i}_2^2\right)\right].
    \]
    Standard EM updates $\vpi, \vmu$ as the maximizer of $Q$ function: \[(\vpi^{(t+1)}, \vmu^{(t+1)}) = \arg\max_{\vpi, \vmu} Q(\bm{\vpi, \vmu} |\vpi^{(t)}, \bm{\mu}^{(t)}).\] Gradient EM replaces this sometimes costly maximization step with one gradient step on $Q$. Let $\eta\in\R^+$ be the learning rate. The update rule is:
    \begin{equation}\label{eq: vanilla gradient EM update}
        (\vpi^{(t+1)}, \vmu^{(t+1)}) \leftarrow (\vpi^{(t)}, \vmu^{(t)}) + \eta \nabla_{\vpi, \vmu} Q(\bm{\vpi, \vmu} |\vpi^{(t)}, \bm{\mu}^{(t)}).
    \end{equation}
\end{itemize}
Here $\vpi^{(t)}, \vmu^{(t)}$ denotes the collection of training parameters at $t^{\text{th}}$ iteration and $\pi_i^{(t)}, \vmu_i^{(t)}$ are similarly defined. 
To focus the optimization analysis, we primarily study the population update in infinite sample limit without statistical errors (see, $e.g.$, \cite{Xu2016GlobalAO, xu2024toward}). We later extend our results to the finite-sample setting with polynomial sample complexity in \cref{thm: sample complexity}.

\paragraph*{Exact-parameterized GMM} Establishing a rigorous recovery guarantee for (gradient) EM on GMM has been an important problem in machine learning theory, due to its popularity in practice. The most natural setting here is exact-parameterization, where the model uses the same number of component Gaussians as the ground truth ($n=m$). Since the seminal result \cite{balakrishnan2017statistical} provided the first convergence guarantee for $2$-GMMs, a long line of work has followed  \citep{daskalakis17TenSteps, Xu2016GlobalAO, kwon_em_2020, yan_convergence_2017}. However, proving global convergence for general $m$-GMMs with $m\geq 3$, even in the well-separated setting, turns out to be intractable due to the existence of bad local minima that (gradient) EM can get trapped \citep{Jin2016LocalMI}.

\paragraph*{Over-parameterized GMM} Motivated by the impossibility result in the exact-parameterized setting, another series of papers study the behavior of (gradient) EM in over-parameterized setting where the model uses more component Gaussians than in the ground truth ($n>m$) \citep{dasgupta2013two,Dwivedi2018SingularityMA, Dwivedi2019SharpAO, chen2024local, xu2024toward}. 
This line of work is motivated by the implicit belief that over-parameterization might improve the optimization landscape of GMMs, thus help (gradient) EM avoid spurious local minima. 
The following important conjecture (see Section~5 of~\cite{chen2024local}) lies at the core of this line of research:

\begin{conjecture*}
    Overparametrization enables (gradient) EM to converge to the ground truth from random initialization on Gaussian Mixtures.
\end{conjecture*}
In this paper, we prove the above conjecture by showing that gradient EM with random initialization converges globally when using an overparameterized GMM to recover the underlying (well-separated) GMM. Our main result can be stated as follows:

\begin{theorem}[Main result, informal version of Theorem \ref{thm: main}]
    Assume the ground truth Gaussian means $\{\vmu_i^*\}_{i=1}^m$ are well separated (\cref{assump: delta}). 
    Consider the problem of using over-parameterized GMM \eqref{eq: over-parameterized GMM} to learn the ground truth GMM \eqref{eq: ground truth GMM}. Suppose we use randomly initialized population gradient EM with near-optimal weight updates (Algorithm \ref{alg}) with polynomially small step size $\eta$. If the amount of over-parameterization is at least logarithmically large, then with high probability, gradient EM converges globally from random initialization to ground truth with rate $\O{\frac{1}{T^2}}$.
\end{theorem}
For the detailed assumptions and algorithmic setup we refer the readers to Section \ref{Section: prelim}.
We compare our result with prior art in Table~\ref{tab: comparison}. 
In this paper, because the problem is nonconvex, we use the term \textit{global convergence} to mean that the algorithm converges to the ground truth starting from a \textit{random initialization}, rather than starting from arbitrary or carefully chosen initializations.

\paragraph*{Main Contributions}
The major contributions of this paper can be summarized as follows:
\begin{itemize}
    \item We prove that over-parameterized gradient EM converges globally from random initialization to recover well-separated GMMs. This is the first global convergence result of EM or gradient EM for learning general $m$-GMMs. Comparisons of our result with previous analysis of EM/gradient EM are listed in Table \ref{tab: comparison}.
    \item We introduce novel tools from other fields and apply them to the analysis of gradient EM:
    \begin{itemize}
        \item 
        We use the idea of test functions to characterize the geometric optimization landscape of GMM learning using gradient EM.

        \item We use homotopy method between \(p\), \(p_*\) to analyze the local landscape and dynamics.
    \end{itemize}
    Test-function arguments have been used in hypothesis testing and in the analysis of shallow neural networks \citep{gretton2012kernel,zhou2021local,zhou2024how}, while homotopy ideas have a long history in analysis \citep{liao2012homotopy}.  In our setting, they provide a bridge between the optimization dynamics of gradient EM and robust identifiability for Gaussian mixtures.
\end{itemize}

\begin{table}[t]
    \arrayrulecolor{black}
    \centering
    \begin{tabular}{|c|c|c|}
    \hline
        Paper       & Setting & Convergence \\ \hline
        &&\\[-11pt]\hline
        \cite{balakrishnan2017statistical} & $m=n=2$ & Local \\ \hline
        \cite{daskalakis17TenSteps, Xu2016GlobalAO, wu2019randomly} & $m=n=2$ & Global \\ \hline
        \cite{yan_convergence_2017, kwon_em_2020, segol_improved_2021}& $ m=n$ arbitrary & Local\\ \hline
        \cite{Dwivedi2018SingularityMA, Dwivedi2019SharpAO} & $m=1, n=2$ & Global\\\hline
        \cite{Dwivedi2018TheoreticalGF} & $m \in \{2,3\}, n=2$ & Local \\\hline
        \cite{xu2024toward} & $m=1, n$ arbitrary & Global\\ \hline
        \cite{Jin2016LocalMI} & $m=n \ge 3$ & Suboptimal\\ \hline
        \cellcolor{gray!20}This work & \cellcolor{gray!20}$m$ arbitrary, $n\geq\Omega(m \log m)$ & \cellcolor{gray!20}Global\\
        \hline
    \end{tabular}
    \caption{Comparison of our result with prior art on gradient EM for GMMs. ``Suboptimal'' denotes the negative result showing high-probability convergence to suboptimal local minima. For local convergence, initialization is close to the ground truth. For global convergence and suboptimal convergence, the initialization is random.}
    \label{tab: comparison}
    \vspace{-5pt}
\end{table}

\subsection{Related works}

\paragraph*{EM and gradient EM algorithm for learning GMMs}
The literature on theoretical analysis of (gradient) EM is long and vast \citep{wu1983convergence,xu1996convergence}. Here we only focus on the existing analysis for GMMs.

A series of works have established local and global convergence results for exactly-parameterized 2-GMMs \citep{balakrishnan2017statistical, daskalakis17TenSteps, Xu2016GlobalAO, wu2019randomly, Weinberger2021TheEA}. These studies often consider a symmetric setting where the true means are $\pm\vmu^*$, and EM is used to recover them via estimates $\pm\vmu$. Notably, EM can recover $\vmu^*$ regardless of $\norm{\vmu^*}$, even when it is small (often called the low signal-to-noise regime). When $\norm{\vmu^*}=0$, the problem becomes overparameterized: using a 2-GMM with symmetric means to learn a single Gaussian. We note that the symmetric parameterization often simplifies the analysis. In contrast, our work considers the general, non-symmetric setting, which poses additional challenges.

Going beyond 2-GMMs, another line of work focused on local convergence for general exact-parameterized $m$-GMMs \citep{yan_convergence_2017, zhao2020statistical, kwon_em_2020,segol_improved_2021}. These results typically show that, starting from a good initialization (in the sense that initial estimate is within a certain distance of the true cluster means), EM iteratively contracts the parameter error and converges to the ground truth. However, a general global convergence analysis from random initialization in exact-parameterized regime is intractable due to the spurious local minima, even for well-separated GMMs \cite{Jin2016LocalMI} . 

In the over-parameterized regime, existing works primarily studied the global convergence of using multi-component GMMs to learn a single Gaussian ($m=1$) \citep{Dwivedi2018SingularityMA, Dwivedi2019SharpAO, Dwivedi2018TheoreticalGF, xu2024toward}. Providing evidence for benefits of over-parameterization, \cite{dasgupta2013two} proposed a variant of EM with pruning and \cite{chen2024local} studied the geometric landscape of likelihood loss for GMMs. Yet none of the above works imply global convergence of (gradient) EM beyond the special case of $m=2$.

Most of the above works focus only on learning and updating the means $\vmu$. This is natural in the exact-parameterized setting, where the weights can be assumed known, and in the overparameterized case of learning 1-GMM, where the total weight sums to 1. However, in the general overparameterized setting ($m \ge 2$) that we study, it becomes necessary to also update the weights $\vpi$, as assigning accurate weights a priori is difficult even when the ground-truth weights are known.

\paragraph*{Other methods for provable learning GMMs}
The first estimators for the GMMs (learning univariate 2-GMM) were based on the method-of-moments, as introduced by \cite{pearson1894contributions}. The general idea of method-of-moments is to match the moments of the model to the empirical moments of the data. Solving these polynomial equations often would reveal useful information about the ground-truth model. This idea has been used in many of the results discussed below.

Since \cite{dasgupta1999learning}, algorithmic approaches other than EM for GMM learning have been developed in the theoretical computer science and machine learning community, leading to a long and rich line of work. The high-level idea is typically based on clustering: first cluster the data in a principled way, then estimate the means. Later works extended these ideas to more general covariance structures \citep{sanjeev2001learning} and improved the required separation conditions using spectral methods such as PCA \citep{vempala2004spectral, achlioptas2005spectral, brubaker2008isotropic}. The best-known separation bounds in this line require at least $\Omega(d^{1/4})$ (or $\Omega(m^{1/4})$ after reducing the dimension to the subspace spanned by the true means).

For learning $m$-GMMs with fixed $m$ (i.e., $m=O(1)$), another line of work developed algorithms under minimal separation assumptions, requiring only that the Gaussians are statistically distinguishable \citep{kalai2010efficiently, belkin2010polynomial, moitra2010settling, diakonikolas2020small}. However, these results often come at the cost of runtime or sample complexity that is exponential in $m$, which is shown to be tight in \cite{hardt2015tight}.

A separate line of research focuses on tensor decomposition methods \citep{hsu2012learningmixturessphericalgaussians, anandkumar2014tensor, AndersonBGRV14, ge2015learning,bhaskara2014smoothed}. These methods typically rely on algebraic non-degeneracy conditions, often justified via smoothed analysis, and use low-order moment tensors (usually third or fourth order) to recover parameters. 

Recently, a series of works develop theoretical algorithms with sum-of-squares proofs \citep{hopkins2017mixturemodelsrobustnesssum, kothari2018robust} and method-of-moments \citep{diakonikolas2018list}, using higher-order moments to relax the separation requirement to $\Delta = \Omega(m^{1/\gamma})$, with runtime and sample complexity $O(d^{\poly(1/\gamma)})$. This was further improved to $\Omega(\sqrt{\log m})$ separation in \cite{liu2022clustering, diakonikolas2024implicit}, which matches the information-theoretic lower bound when using only polynomial samples \citep{regev2017learning}.

While these algorithms achieve strong or even optimal theoretical guarantees, they tend to be more complex and less practical than (gradient) EM. They typically require estimating high-order moments, which demands a large number of samples. In practice, EM remains the most widely used method due to its simplicity and efficiency.
Our work aims at proving theoretical guarantees for gradient EM and is orthogonal to these algorithmic results. Nevertheless, our analysis connects the technical tools developed in these works with the analysis of gradient EM.

\section{Preliminaries}\label{Section: prelim}

\paragraph*{Notation}
Define $[n]\coloneqq \{1,2,\ldots,n\}$. For any vector $\vv\in\R^d$, we use $\norm{\vv}$ as its $\ell_2$ norm and $\Bar{\vv}\coloneqq\frac{\vv}{\norm{\vv}}$ to denote its normalization. Denote $\pi_{\min}^*\coloneqq\min_{i\in [m]} \pi_i^*$ as the minimum weight of ground truth Gaussian Mixtures. We denote the collection of model mixing weights and means with $\vpi\coloneqq (\pi_1, \ldots, \pi_n)^{\top}$ and $\vmu\coloneqq (\vmu_1^{\top},\ldots, \vmu_n^{\top})^{\top}$, respectively. We use the shorthand $ \E_{x\sim j}[\cdot]\coloneqq \E_{x\in \N(\vmu^*_j, \mI)} [\cdot]$ as the expectation over the $j^{th}$ ground truth distribution.
We use $\mathcal{O}, \Omega, \Theta$ to hide constants and $a\lesssim b$ to denote there exists an absolute constant $C$ such that $a\leq Cb$.

\paragraph*{KL divergence loss}
A well known property of gradient EM is its equivalence with gradient methods on the population likelihood. We first have the following definition of KL divergence loss.
\begin{definition}[Loss function]
Define the loss function $\L$ as the Kullback–Leibler (KL) divergence between the ground truth GMM and the model GMM as
    \begin{equation}\label{eq: loss}
        \L(\vpi,\vmu) 
    \coloneqq \kl(p_*||p)
    = \E_{\vx\sim p_*}[\ell(p(\vx))] - \E_{\vx\sim p_*}[\ell(p^*(\vx))],
    \end{equation}
where $\ell(z) = -\log(z)$ is the negative log loss. 
\end{definition}
Since the minimum of $\kl(p_*||p)$ is achieved when $p=p_*$, global minimum of the loss $\L$ is obtained if and only if the ground truth GMM is recovered. Then the following property connects gradient EM to gradient descent on $\L$.
\begin{claim}\label{claim: Q function to KL loss}
For any $\vpi^{(t)}, \vmu^{(t)}$, we have 
       $ \left.\nabla_{\vpi, \vmu} Q(\bm{\vpi, \vmu} |\vpi^{(t)}, \bm{\mu}^{(t)})\right|_{(\vpi,\vmu)=(\vpi^{(t)},\vmu^{(t)})} = -\nabla_{\vpi, \vmu}\L(\vpi^{(t)}, \vmu^{(t)}).$
\end{claim}

Claim \ref{claim: Q function to KL loss} directly implies the equivalence between the gradient step on the $Q$ function in gradient EM and one gradient step on the KL loss $\L$. Its proof is by direct algebraic calculation (see \cite{dempster1977maximum,Jin2016LocalMI, xu2024toward}). Therefore, gradient EM could be viewed as gradient descent on the loss function $\L$. Our analysis crucially relies on this property and revolves around various properties of $\L$.

Moreover, the gradient can be written in the following form:
\begin{restatable}{claim}{lemgradientform}
    The gradient $\nabla_{\vmu_i} \L(\vmu,\vpi)$ can be written as
    \begin{align}\label{eq: gradient form}
        \nabla_{\vmu_i} \L(\vmu,\vpi)
        = \sum_{j\in [m]} \pi_j^* \E_{\vx\sim j}\left[\psi_i(\vx)\sum_{k\in[n]} \psi_k(\vx)(\vmu_k-\vmu_j^*) \right].
    \end{align}
\end{restatable}

\paragraph*{Algorithmic setup}
Our algorithm for the population setting is presented in Algorithm~\ref{alg}. Since we primarily focus on the population case, we defer the finite-sample version to Algorithm~\ref{alg: fixed sample} in Appendix~\ref{appendix: sample complexity}. The only difference between the two algorithms is that Algorithm~\ref{alg: fixed sample} replaces the expectation in the population update of Algorithm~\ref{alg} with an empirical average over samples.
\begin{algorithm}
\caption{Population gradient-EM with near-optimal weight updates}\label{alg}
\begin{algorithmic}[1]
\STATE \textbf{Input:} Stepsize $\eta$, 
iterations $T$, target error $\eps$, weight subproblem error $\eps_\pi$
\STATE \textbf{(Random)  Initialization:}  $\forall i\in [n], \vmu_i^{(0)} \overset{\text{i.i.d.}}{\sim} p^*, \pi_i^{(0)} = \frac{1}{n}$.
\FOR{$t = 0$ to $T-1$}
    \STATE $\vpi^{(t+1)} \gets \eps_\pi$-\text{optimal}\footnotemark
    \text{ solution of convex subproblem} $\arg\min_{\vpi \in \gP} \mathcal{L}(\vpi, \vmu^{(t)})$ 
    \STATE $\vmu^{(t+1)} \gets \vmu^{(t)} + \eta\nabla_{\vmu} Q(\bm{\vpi, \vmu} |\vpi^{(t+1)}, \bm{\mu}^{(t)})=\vmu^{(t)} - \eta \nabla_{\vmu} \mathcal{L}(\vpi^{(t+1)}, \vmu^{(t)})$
\ENDFOR
\STATE \textbf{Output:} $\vmu^{(T)}, \vpi^{(T)}$
\end{algorithmic}
\end{algorithm}
\footnotetext{By $\eps$-optimal we mean in the sense of first-order KKT optimality and loss is non-increasing that $\mathcal{L}(\vpi^{(t+1)}, \vmu^{(t)})\le \mathcal{L}(\vpi^{(t)}, \vmu^{(t)})$, see more precise definition in Appendix~\ref{appendix: alg}.}

For initialization, we consider a standard random initialization scheme \citep{dasgupta2013two,srebro2006investigation, Jin2016LocalMI} where the mean vectors $\vmu_i^{(0)}$ are initialized as i.i.d. samples from the observed ground truth data. We also assume the mixing weights are initialized as equal for simplicity: $\pi_i=\frac{1}{n}$ for all $i\in [n]$. This initialization is intuitively reasonable and has been observed to perform well in practice. Note that in the population setting, this is equivalent to sampling $\vmu_i^{(0)}$ directly from the ground-truth GMM $p_*$, as in Algorithm~\ref{alg}.

For the iterative updates, careful readers might notice that the classical definition of gradient EM update described in \eqref{eq: vanilla gradient EM update} is actually not well-defined for the mixing weights $\vpi$, as they reside in a 
simplex $\gP\coloneqq\{\vpi: 0\le\pi_i\le1, \forall i, \sum_i \pi_i=1\}$ while gradient EM update is defined for unconstrained problems. Note that obtaining the maximum likelihood estimate w.r.t $\vpi$ is a convex problem and can be efficiently solved by classic convex optimization approaches \citep{beck2003mirror,jaggi2013revisiting,ryu2016primer}. Therefore, we assume for simplicity that $\vpi$ is updated in each iteration with a $\eps_\pi$-optimal solution of the convex subproblem $\min_{\vpi \in \gP} \mathcal{L}(\vpi, \vmu^{(t)})$ where $\eps_\pi$ is the target error of this subproblem to be specified. Meanwhile $\vmu$ is still updated with standard gradient EM. 

\begin{remark}
    Although Algorithm~\ref{alg} considers near-optimal weight updates for simplicity, it also can be seen as a limiting case of standard gradient EM with different step sizes for $\vpi$ and $\vmu$. Specifically, by setting the step size $\eta_\pi$ much larger than $\eta_\mu$---that is, updating $\vpi$ significantly faster than $\vmu$---Algorithm~\ref{alg} is recovered in the limit as $\eta_\mu / \eta_\pi \to 0$ in each iteration. Similar algorithmic reductions are often used for the analysis of neural networks \citep{borkar2008stochastic,marion2023leveraging,takakura2024meanfield,berthier2024learning,barboni2025ultra}. 
\end{remark}

\paragraph*{Assumptions}
We make the following standard assumptions on the ground-truth GMM.

Define the separation of the true centers by
\[
    \Delta_{\max}
    :=
    \max_{i\ne j\in[m]}\norm{\vmu_i^*-\vmu_j^*},
    \quad
    \Delta
    :=\min_{i\ne j \in[m]}\norm{\vmu_i^*-\vmu_j^*}.
\]
By translation equivariance of the likelihood and of the gradient EM updates,
we work in coordinates such that
$\max_{\ell\in[m]}\norm{\vmu_\ell^*}\le \Delta_{\max}$.
For instance, this holds if the origin is chosen as \(m^{-1}\sum_{\ell\in[m]}\vmu_\ell^*\).

\begin{assumption}[Well-separatedness]\label{assump: delta}
    The separation between the ground truth Gaussians satisfy the following: there is a large enough constant $C_{\rm sep}$ such that
    \footnote{Technically, $\Delta$ is not well-defined when $m=1$, so this separation assumption is imposed only when \(m\ge2\).  When \(m=1\), there are no cross-cluster terms, so in the proof one should view all cross-group bound of $e^{-\Delta^2}$ type as 0 and $\Delta_{\max}=0$.}
    \[\Delta\ge C_{\rm sep}\max\left\{\sqrt{\log(dnm\Delta_{\max}/\pimins)}, \sqrt{dn}\right\}.\]
\end{assumption}

\begin{remark}
    The separation condition in Assumption~\ref{assump: delta}, while stronger than the minimal $\Omega(\sqrt{\log m})$ requirement \citep{regev2017learning}, allows us to focus on the well-separated regime and simplifies the analysis. We emphasized that as discussed in the introduction, it is known that EM algorithm would even fail under such condition in \cref{assump: delta} when exact-parametrized. We believe this is a natural starting point for future research aimed at further weakening the separation requirement under the overparameterized setting. 
    
    A simple effective-dimension trick can further reduce this requirement from $\Omega(\sqrt{d})$ to $\Omega(\sqrt{m})$ (see Section~7.3.1 of \cite{chen2024local} for details).
    Roughly speaking, the idea is to first identify the subspace spanned by the ground-truth means and initialize the algorithm within this subspace. This subspace only has dimension $m$ instead of $d$. As a result, the optimization dynamics remain in this subspace, effectively reducing the problem’s dimensionality from $d$ to $m$.
\end{remark}

As an example, all the above assumptions hold when $\vmu_i^* \sim \gN(\vzero, \Delta^2 \mI_d)$ for sufficiently large $\Delta$, in the high-dimensional regime where $m \ll d$. For simplicity, casual readers may assume $\Delta_{\max}= \Theta(\Delta) \gtrsim \sqrt{d}$, $\max_i \pi_i^* / \min_i \pi_i^* = \Theta(1)$, and $n = \Theta(m \log m) \ll d$.

\section{Main result}\label{sec: main result}

We now present our main result. It shows that with only logarithmic overparametrization, gradient EM can recover ground-truth model in polynomial time.
\begin{restatable}[Main result]{theorem}{thmmain}\label{thm: main}
    Under \cref{assump: delta}, fix any \(\delta\in(e^{-\sqrt d},1)\)
    and target error \(\eps>0\). Suppose
    $n\gtrsim \frac1{\pimins}\log\frac{2m}{\delta\pimins}$.
    There exist polynomial quantities \(A_{\rm rate},A_{\eta},A_{\pi}\)
    and an inverse-polynomial threshold \(\eps_0\) satisfying
    \[
        A_{\rm rate},A_{\eta},A_{\pi}
        =
        \poly\left(d,m,n,\frac1{\pimins}\right),
        \qquad
        \eps_0
        =
        \frac{1}{\poly\left(d,m,n,\frac1{\pimins}\right)}
    \]
    such that the following holds. Run \cref{alg} with stepsize $\eta\le \frac{c}{A_{\eta}}$ and weight subproblem error $\eps_\pi\le \frac{c}{A_{\pi}}\min\{\eps,\eps_0\}$.
    Then, with probability at least \(1-\delta\) over the random
    initialization, the population loss satisfies
    \[
        \L(\vpi^{(T)},\vmu^{(T)})\le \eps
    \]
    after
    \[
        T=
        \begin{cases}
        \displaystyle
        O\left(\frac{dn}{\eta\eps}\right),
        & \text{if } \eps\ge \eps_0,
        \\[1.2ex]
        \displaystyle
        O\!\left(\frac{dn}{\eta\eps_0}\right)
        +
        \frac{A_{\rm rate}}{\eta}
        \sqrt{\frac1\eps-\frac1{\eps_0}},
        & \text{if } \eps<\eps_0 .
        \end{cases}
    \]
    Moreover, as \(\eps\to0\), we converge to the  solution that for every \(i\in[n]\), either there exists
    \(\ell\in[m]\) such that \(\vmu_i=\vmu_\ell^*\), or \(\pi_i=0\).
    Equivalently, the learned GMM recovers the ground-truth GMM
    \(p^*\).
\end{restatable}
When $\frac{\max_i \pi_i^*}{\min_i \pi_i^*}=\Theta(1)$ (or equivalently $\pi_i^*=\Theta(1/m)$), the above result implies that we only need $n=\Theta(m\log m)$ components to recover ground-truth $p_*$. Since at least $m$ components are required to represent the ground-truth $p_*$, this suggests that a small logarithmic amount of overparameterization suffices to guarantee convergence. This is the first global convergence and recovery result for EM or gradient EM beyond the special case of $m=2$. We believe both the convergence rate and the separation requirement in our assumptions can be improved, and we leave these refinements for future work.

Some previous works \citep{dasgupta2013two} consider variants of the EM algorithm utilizing a pruning process to remove redundant component Gaussians. Our proof in fact shows that gradient EM automatically prunes the redundant centers $\vmu_i$ as their weights $\pi_i$ converge to $0$.

The proof follows a two-stage analysis: first the loss decreases below the
local threshold \(\eps_0\), and then it converges to an arbitrary target
error \(\eps\). See more details in Section~\ref{sec: proof sketch}.

In contrast to our result, it is known that with exact-parametrized GMM, gradient EM does not converge to global minima, even for well-separated GMMs satisfying Assumption \ref{assump: delta}.
\begin{theorem}[Negative result in \cite{Jin2016LocalMI}, rephrased]\label{thm: counter example}
    Consider the case $m=n$ and $\pi_i=\pi_i^*=\tfrac1m$, and using first-order EM algorithm with stepsize $\eta <1$ under random initialization. There exists a universal constant $c$, for any $m \geq 3$ and any constant $C_{\rm{gap}} > 0$, such that there is a well-separated $m$-$\mathrm{GMM}(\vmu^*)$ with
    \[
    \mathbb{P}\left( \forall t \geq 0,\; \L(\mu^{(t)}) \geq \L(\mu^*) + C_{\rm{gap}} \right) \geq 1 - e^{-cm}.
    \]
\end{theorem}
The main idea behind Theorem \ref{thm: counter example} from \cite{Jin2016LocalMI} is that a randomly initialized exact-parameterized GMM will mismatch model means $\vmu$ with the ground truth with high probability. In contrast, the coupon collector's problem implies that a randomly initialized over-parameterized GMM will, with high probability, assign at least one $\vmu_i$ to each ground truth Gaussian, which is key to ensuring global convergence. Theorem~\ref{thm: main} and \ref{thm: counter example} together demonstrate the key role over-parameterization plays in global convergence of gradient EM.

\paragraph*{Sample complexity}
Our results naturally extend to the setting where only a polynomial number of samples is available. We use a fixed batch of \(N\) samples throughout the run and minimize the empirical negative log-likelihood on these data. See
\cref{alg: fixed sample} in \cref{appendix: sample complexity}.
\begin{restatable}[Sample complexity]{theorem}{thmsamplecomplexity}
\label{thm: sample complexity}
    Under the same setting as \cref{thm: main}, fix any target error
    \(\eps>0\), and set \(\eps_{\rm stat}:=\min\{\eps,\eps_0\}\).  There
    exists a polynomial quantity
    \[
        \sfA_{\rm samp}
        :=
        \poly\left(
            d,m,n,\frac1{\pimins},\Delta_{\max}
        \right)
    \]
    such that the following holds.  Run \cref{alg: fixed sample} with
    \(\eta\le c/A_\eta\) and
    \(\eps_\pi\le cA_\pi^{-1}\eps_{\rm stat}\).  If
    \begin{align}
        N
        \ge
        C\sfA_{\rm samp}\eps_{\rm stat}^{-2}
        \log\frac{C\sfA_{\rm samp}}{\eps_{\rm stat}\delta},
        \label{eq:fixed-sample-size}
    \end{align}
    then, with probability at least \(1-\delta\) over the initialization and data, \cref{alg: fixed sample} outputs an iterate satisfying
    \[
        \L(\vpi^{(T)},\vmu^{(T)})\le \eps
    \]
    after
    \[
        T=
        \begin{cases}
        \displaystyle
        O\!\left(\frac{dn}{\eta\eps}\right),
        & \text{if } \eps\ge \eps_0,\\[1.2ex]
        \displaystyle
        O\!\left(\frac{dn}{\eta\eps_0}\right)
        +
        \frac{A_{\rm rate}}{\eta}
        \sqrt{\frac1\eps-\frac1{\eps_0}},
        & \text{if } \eps<\eps_0 .
        \end{cases}
    \]
    In particular, for \(\eps<\eps_0\), the number of samples is
    \(N=\widetilde O(\sfA_{\rm samp}\eps^{-2})\).
\end{restatable}

This result shows that overparametrized gradient EM can efficiently learn GMMs with a polynomial-size sample $\eps^{-2}$ and polynomial runtime $\eps^{-1/2}$ under our assumptions, ignoring other dependencies to achieve population loss $\L\le\eps$. Our goal is to establish polynomial bounds without aiming for tightness, and we leave the refinement of these estimates to future work.

\begin{figure*}[t!]
    \centering
    \begin{subfigure}[t]{0.48\linewidth}
        \centering
        \raisebox{-0.05in}{\includegraphics[width=\linewidth]{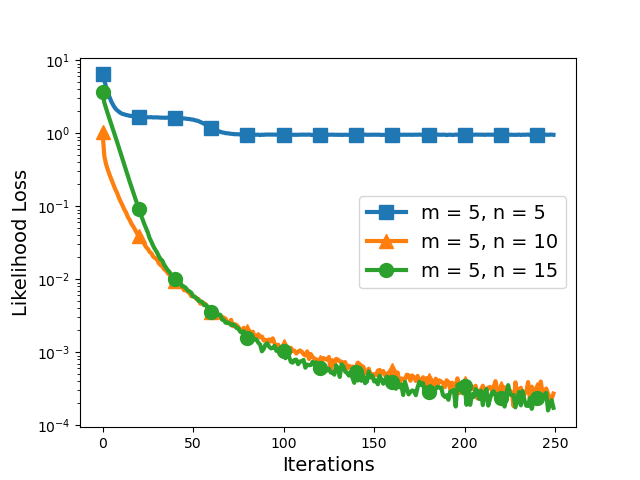}}
        \label{fig: kl loss}
    \end{subfigure}
    \hfill\hspace{-17pt}
        \begin{subfigure}[t]{0.5\linewidth}
        \centering
        \includegraphics[width=\linewidth]{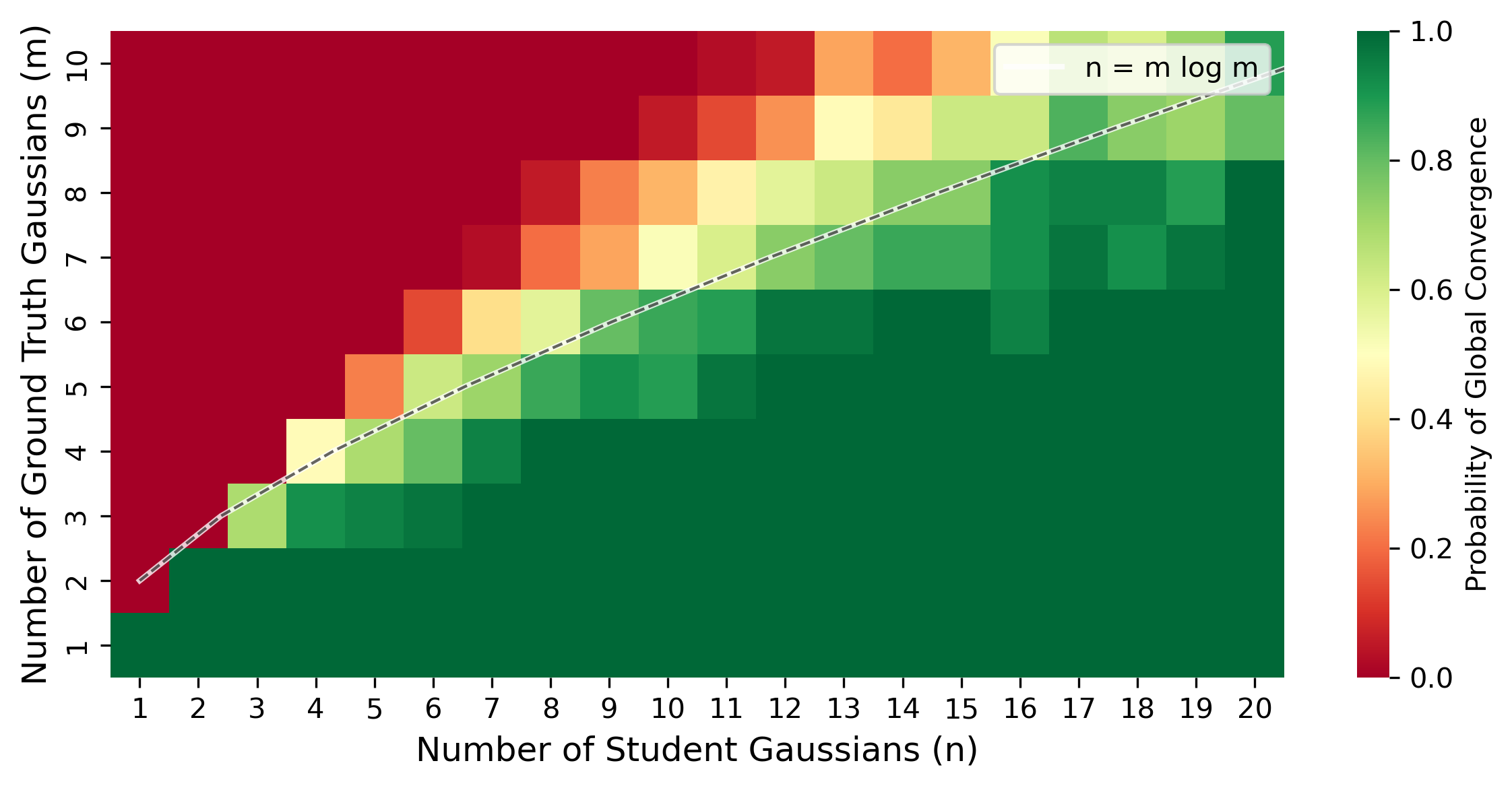}
        \label{fig: convergence probability graph}
    \end{subfigure}

    \vspace{0.2cm}
    \begin{subfigure}[t]{0.45\linewidth}
       \centering
        \includegraphics[width=\linewidth]{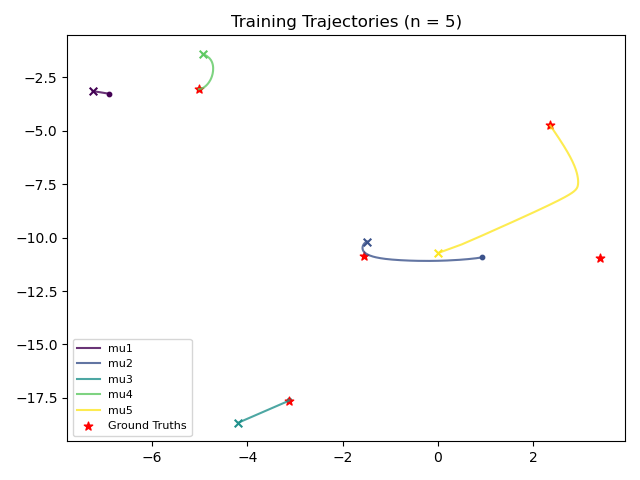}
        \label{fig: exact-parameterized trajectory}
    \end{subfigure}
        \hfill
    \begin{subfigure}[t]{0.45\linewidth}
        \centering
        \includegraphics[width=\linewidth]{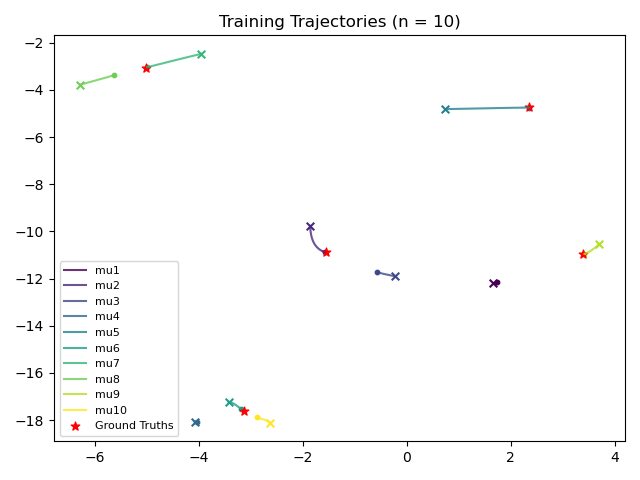}
        \label{fig: over-parameterized trajectory}
    \end{subfigure}
    \caption{
    \bftext{Top Left}: Dynamics of likelihood loss $\L$, exact-parameterization v.s. over-parameterization;
    \bftext{Top Right}: Impact of over-parameterization on global convergence probability. 
    \bftext{Bottom Left}: Exact-parameterized gradient EM converges to spurious local minima; \bftext{Bottom Right}: Global convergence trajectory of over-parameterized gradient EM.}    \label{fig: main}
\end{figure*}

\paragraph*{Experiments}
Empirical simulations validating our result are provided in Figure \ref{fig: main}.
The top left panel shows that over-parameterized gradient EM converges to the ground truth while exact-parameterized gradient EM does not. The result is reported for randomly initialized gradient EM on a well separated ground truth $5$-GMM with model size $n=5,10,15$. The top right panel provides a more fine-grained record of the global convergence probability with differently over-parameterized $(m,n)$ pairs. We run randomly initialized gradient EM for each $(m,n)$ pair $35$ times and report the portion of runs converging to the global optimum. It can be observed that larger $n$ improves the probability of recovery\footnote{Recovery is empirically measured by whether gradient EM enters a sufficiently small neighborhood of the ground truth.}. The phase transition around $n = m$ highlights necessity of over-parameterization. The transition region roughly aligns with the reference line \(n=m\log m\), which is consistent with the logarithmic overparameterization predicted by \cref{thm: main} (consider $\pimins=\Theta\left(1/m\right)$).

The bottom left and right panels plot the training trajectory in exact v.s. over parameterized regimes. In the exact-parameterized case (bottom left panel) gradient EM gets trapped in a spurious local minima where $\vmu_2$ converges to a local minimizer between two ground truth Gaussians. This corresponds to the "one-fits-many" local minima structure \citep{chen2024local}. In the over-parameterized case (bottom right panel) each ground truth is fitted with at least one $\vmu_i$. In both regimes there are some $\vmu_i$ not converge to any of the ground truths. Instead, their mixing weights converge to $0$. This phenomenon corresponds to the dynamic weight challenge in our theoretical analysis (see~\termlink{challengeweight}).

\section{Proof sketch: 2-stage analysis}\label{sec: proof sketch}

We provide a sketch of the proof in the population setting (Theorem~\ref{thm: main}). It follows a two-stage analysis: global convergence (loss drops below a threshold $\eps_0$, Sections~\ref{sec: proof sketch global}) and local convergence (loss reaches the target error $\eps$, Section~\ref{sec: proof sketch local}).

We begin by introducing some notations. Let the components $\vmu_i$ be partitioned as $[n] = \cup_{\ell \in [m]} S_\ell$, where $S_\ell := \left\{ i \in [n] \;\middle|\; \norm{\vmu_i - \vmu_\ell^*}_2 \le \norm{\vmu_i - \vmu_j^*}_2 \;\; \forall j \ne \ell \right\}$
is the set of components closest to $\vmu_\ell^*$ among all ground truths. Define the potential function $U(\vmu) = \sum_{\ell \in [m]} \sum_{i \in S_\ell} \norm{\vmu_i - \vmu_\ell^*}_2^2,$
which measures the total distance of components from their nearest ground truths. We assume each $S_\ell$ is non-empty when discussing $U$. We can show that $U$ remains bounded during training. Also, let $\hpi_\ell := \sum_{i \in S_\ell} \pi_i$ denote the total weight in $S_\ell$.

\subsection{Key challenges in analysis}

Our analysis faces two main challenges due to overparameterization and multiple true components:

\bftext{\anchorterm{challengeweight}{Challenge 1}: Dynamic weights.}    
Unlike prior works that often keep weights $\vpi$ fixed, we update both weights $\pi_i$ and means $\vmu_i$. In the overparameterized setting ($n>m$), some weights $\pi_i$ may converge to zero, preventing the corresponding $\vmu_i$ from converging to any ground-truth components. This means identifiability cannot be guaranteed per component, only collectively over component groups. This contrasts with much of the existing analysis, which often focuses on the parametric distance between the ground-truth and model parameters (e.g., \cite{balakrishnan2017statistical,Dwivedi2018SingularityMA,yan_convergence_2017}).

\bftext{\anchorterm{challenge: cross term}{Challenge 2}: Cross terms.}
In the multi-cluster setting, a key challenge is controlling cross-term interactions between different ground-truth components: for example, terms like $\psi_i$ and $\psi_k$ that appear in the gradient \eqref{eq: gradient form} for $\vmu_i$ and $\vmu_k$ from different groups. These terms persist even at the global optimum, typically at the scale $\exp(-\Theta(\Delta^2))$, making it hard to prove convergence to arbitrarily small error $\eps$.  
Existing results only guarantee error up to this scale, even when analyzing the global minima \citep{chen2024local}.

To overcome these challenges, we combine three ingredients.  First, we use bounded localized test functions to turn small KL loss into quantitative identifiability of the overparameterized parameters at the group level instead of individual weights. Second, in the local phase we expand each nearby model density around its assigned true center and use the resulting zeroth-, first-, and second-order cancellations to show that cross-cluster errors are proportional to the current loss.  Third, we use a homotopy comparison through \(p_\theta=\theta p+(1-\theta)p_*\), together with certain Gaussian estimates, to control these Taylor remainders despite the small denominators that appear in the KL geometry.

\subsection{Global convergence analysis}\label{sec: proof sketch global}

The goal of this part is to show that, under random initialization, the loss decreases below a local threshold \(\eps_0\) within polynomial time.  This threshold is inverse-polynomial in the relevant problem parameters.

We begin with random initialization. With only logarithmic overparameterization, we show that with high probability, each ground-truth center $\vmu_\ell^*$ has at least one $\vmu_i$ that is closer to it than to any other ground-truth center. The proof follows from a standard coupon collector argument combined with concentration inequalities and is deferred to \cref{appendix: global}.

\begin{restatable}[Initialization]{lemma}{leminit}\label{lem: init}
    Under \cref{assump: delta} and random initialization that $\vmu_i \overset{\text{i.i.d.}}{\sim} p_*$,
    fix any $\delta\in(e^{-\sqrt d},1)$. Suppose
    $n \gtrsim \frac{1}{\pimins}\log \frac{2m}{\delta}$.
    Then, with probability at least \(1-\delta\), we have
    \begin{enumerate}
        \item For each cluster $\ell\in[m]$, there exists at least one component
        $\vmu_i$ such that $\vmu_i\in S_\ell$ and
        $\norm{\vmu_i-\vmu_\ell^*}_2\lesssim \sqrt{d}$.
        \item The initial potential satisfies
        \(U(\vmu^{(0)})\lesssim dn\).
        \item The initial loss satisfies
        $\L^{(0)}\lesssim \log(n)+d$.
    \end{enumerate}
\end{restatable}

We now present the main result for global phase below. All conditions are satisfied by random initialization above with high probability. The full proof is deferred to \cref{appendix: global}, and we give a proof sketch below.
\begin{theorem}[Main result for global phase]
\label{thm: global main text}
    Under \cref{assump: delta}, consider \cref{alg} with random initialization event in \cref{lem: init}. Let \(L_0\asymp d+\log n\) and \(B^2\asymp dn\) be chosen large enough so that
    $\L(\vpi^{(0)},\vmu^{(0)})\le L_0$ and 
    $U(\vmu^{(0)})\le \frac{B^2}{2}$.
    There are universal constants \(c,C>0\) and polynomial quantities $\sfA_{\rm g},\sfA_{\eta}$
    such that the following holds. Let \(\eps_0\) be the local threshold in \cref{thm: main}. 
    
    If \(\eta\le c/\sfA_{\eta}\) and $\eps_\pi
        \le
        c\sfA_{\rm g}^{-1}\eps_0$, then for any
    \(\eps\ge \eps_0\), the global phase reaches \(\L\le \eps\) within
    \[
        T_1
        \le
        \frac{C B^2}{\eta\eps}
    \]
    iterations.  Moreover, before the global phase stops,
    we have $U(\vmu^{(t)})\le \frac{B^2}{2}.$
\end{theorem}

We show that the loss drops below the threshold $\eps_0$ within polynomial time. As discussed in \termlink{challengeweight}, there is no simple parametric distance suitable for proving convergence in our setting. Instead, we focus directly on the loss and demonstrate that it decreases over time.
The proof idea is to establish a lower bound on the gradient norm (i.e., the Kurdyka \PolishL{}ojasiewicz (KL) condition $\norm{\nabla \L}\ge \L$), which ensures that the algorithm decreases the loss at each step.

To prove this lower bound, we construct a descent direction and show it has non-trivial correlation with the gradient. The descent direction is intuitive: we move each $\vmu_i \in S_\ell$ toward its nearest ground-truth mean $\vmu_\ell^*$.

\begin{lemma}[Descent Direction]
\label{lem:global-full-potential-linear-main-text}
    Under \cref{thm: global main text}, suppose
    $U(\vmu)\le B^2/2$ and
    $\eps_0\le \L(\vpi,\vmu)\le L_0.$
    Then
    \[
        \sum_{\ell=1}^m\sum_{i\in S_\ell}
        \left\langle
            \nabla_{\vmu_i}\L(\vpi,\vmu),
            \vmu_i-\vmu_\ell^*
        \right\rangle
        \gtrsim
        \L(\vpi,\vmu).
    \]
\end{lemma}

To lower bound the above gradient projection term on LHS, note that Assumption~\ref{assump: delta} and $U(\vmu)\le B^2/2$ imply that for $i\in S_\ell$, $\norm{\vmu_i-\vmu_\ell^*}_2 \le B \ll \Delta \le \norm{\vmu_i-\vmu_j^*}_2$. Intuitively, this allows us to bound the cross-group interaction (\termlink{challenge: cross term}) since Gaussian density decays exponentially fast: 
\begin{lemma}[Item 2 in \cref{lem:global-separated-overlap}]
\label{lem: psi cross term global informal}
For $i\in S_\ell$ and $j\ne \ell$, we have cross term
    $
        \E_{\vx\sim j}\left[
            \psi_i(\vx)
        \right]
        \lesssim \exp(-\Theta(\Delta^2)).
    $
\end{lemma}
Therefore, after treating all cross terms as small error terms, the gradient projection essentially decomposes into independent same-group quantities.  For each group, define the normalized within-group mixture $q_\ell$, the corresponding within-group posterior weights $\psi_i^{(\ell)}$ and $\tldvpsi_\ell(\vx)$ as
\[
    q_\ell(\vx):=
        \sum_{i\in S_\ell}
        \frac{\pi_i}{\hat\pi_\ell}\phi(\vmu_i;\vx),
    \quad
    \psi_i^{(\ell)}(\vx):=
        \frac{\frac{\pi_i}{\hpi_\ell}\phi(\vmu_i;\vx)}{q_\ell(\vx)},
    \quad
    \tldvpsi_\ell(\vx):=
        \sum_{i\in S_\ell}
        \psi_i^{(\ell)}(\vx)(\vmu_i-\vmu_\ell^*).
\]
Then the full gradient projection is lower bounded, up to exponentially small cross-group errors:
    \begin{align*}
        &\sum_{\ell=1}^m\sum_{i\in S_\ell}
        \left\langle
            \nabla_{\vmu_i}\L(\vpi,\vmu),
            \vmu_i-\vmu_\ell^*
        \right\rangle
        \ge \sum_{\ell\in [m]} \pi_\ell^* \E_{\vx\sim \ell}\left[\norm{\tldvpsi_\ell(\vx) }^2\right]
        - \exp(-\Theta(\Delta^2)).
    \end{align*}

At this point, the problem has been reduced to understanding how well the components inside a single group \(S_\ell\) fit the single Gaussian \(\phi(\vmu_\ell^*)\). Similar forms appear in prior work \citep{xu2024toward} under the single-component setting ($m = 1$). Their approach introduces a dependency on $\pimin:=\min_i \pi_i$, which is problematic in our case since $\pimin$ may be zero due to dynamic updates of $\vpi$ (see \termlink{challengeweight}). Moreover, we improve this lower bound which leads to better convergence rate of $1/t$ than $1/\sqrt{t}$ in \cite{xu2024toward}. To achieve this, we show that
\begin{lemma}[\cref{lem:one-cluster-entropy-production} and \cref{lem:within-group-weight-entropy}]\label{lem: tldpsi lower bound global informal}
    Under \cref{thm: global main text}, for any $\ell\in[m]$ we have
    \begin{align*}
        \E_{\vx\sim\ell}\norm{\tldvpsi_\ell(\vx)}^2
        &\ge
        2\kl(\phi(\vmu_\ell^*)\|q_\ell)
        -
        C\left(
            \eps_\pi
            +
            \sfA_{\rm g}e^{-c\Delta^2}
        \right).
    \end{align*}
\end{lemma}
Combining all above, we already have a gradient lower bound in terms of the RHS of Lemma~\ref{lem: tldpsi lower bound global informal}. 
In \cref{lem:global-loss-comparison} we further relate these terms to the loss $\L$ to obtain the desired gradient lower bound in \cref{lem:global-full-potential-linear-main-text}.

\subsection{Local convergence analysis}\label{sec: proof sketch local}

In previous section, we show the loss can go below $\eps_0$. Here, we show that once below this threshold, the loss will further converge to any $\eps>0$ within time $\poly(d,m,n,1/\eps)$. The full proof is deferred to \cref{appendix: local}.

\begin{theorem}[Main result for local phase]
\label{thm: local main text}
    Under \cref{assump: delta}, there exist
    universal constants \(c,C>0\) and polynomial
    quantities $\sfA_{\eta},\sfA_{\rm rate}$ such that the following holds. 
    Suppose that at time \(T_1\),
        $\L(\vpi^{(T_1)},\vmu^{(T_1)})
        \le
        \eps_0$ and 
        $U(\vmu^{(T_1)})\le B^2/2$, where the local threshold \(\eps_0\) is small enough in \cref{thm: main}.
    If \(\eta\le c/\sfA_{\eta}\) and $\eps_\pi\le c\eps$, then for
    every \(t\ge T_1\) until loss \(\L\le\eps\),
    \[
        \L(\vpi^{(t)},\vmu^{(t)})
        \le
        \frac{A_{\rm rate}^2}
        {(\eta(t-T_1))^2+A_{\rm rate}^2/\eps_0},
        \qquad
        U(\vmu^{(t)})\le B^2.
    \]
\end{theorem}

The proof idea is same as in the global part to lower bound the gradient norm.

\begin{lemma}[Gradient norm lower bound]
\label{lem: grad norm lower bound main text}
    Under \cref{thm: local main text}, we have
    \[
        \norm{\nabla_{\vmu}\L}_F
        \gtrsim
        \sqrt{\frac{\pimins}{n\Xi_{\rm id}}}
        \L(\vpi,\vmu)^{3/4},
    \]
    where $\Xi_{\rm id}
    \lesssim m\left(
        1+d+\log\left(e+\frac{100mn(d+1)}{\pimins}\right)
    \right)^{3/2}$ as in \cref{thm:local-identifiability-log-separation}.
\end{lemma}

Similar to the global part, we construct a descent direction by moving all nearby components $\vmu_i$ toward their closest ground-truth center $\vmu_\ell^*$. Different from global phase, we only focus on near components in local phase because it is possible that some $\vmu_i$ remain far from all $\vmu_\ell^*$ and weight $\pi_i$ go to 0.

\begin{lemma}[Descent direction]
\label{lem: local descent dir main text}
    Under \cref{thm: local main text}, set
    $\tldpi_\ell:=\sum_{i\in S_\ell(\deltacls)}\pi_i$ as the total weights that is $\deltacls$-close to $\vmu_\ell^*$ with $\deltacls:=
    2\sqrt{\frac{\Xi_{\rm id}}{\pimins}}\eps^{1/4}$.
    Then
    \[
        \sum_{\ell=1}^m
        \sum_{i\in S_\ell(\deltacls)}
        \frac{\pi_\ell^*}{\tldpi_\ell}
        \left\langle
            \nabla_{\vmu_i}\L,\vmu_i-\vmu_\ell^*
        \right\rangle
        \ge \frac12\L(\vpi,\vmu).
    \]
\end{lemma}

To handle the gradient projection above, unlike Lemma~\ref{lem: psi cross term global informal} in the global part, the local analysis requires more care. We can now only tolerate perturbations proportional to the loss $\L$, rather than $\exp(-\Theta(\Delta^2))$, since $\L$ may be much smaller (see \termlink{challenge: cross term}). Moreover, cross terms do not vanish even at the global optimum, as $\E_{\vx \sim j}[\phi(\vmu_\ell^*; \vx)/p_*(\vx)] = \exp(-\Theta(\Delta^2))$ for $j \ne \ell$.

To obtain such a bound, we first control the cross terms via a Taylor expansion below.  For a group \(S_\ell\), we expand $\phi(\vmu_i;\vx)$ at its corresponding ground-truth $\vmu_\ell^*$ as
\[
    \sum_{i\in S_\ell}\pi_i\phi(\vmu_i;\vx)
    =
    \hpi_\ell\phi(\vmu_\ell^*;\vx)
    +
    \phi(\vmu_\ell^*;\vx)
    \left\langle
        \vx-\vmu_\ell^*,
        \sum_{i\in S_\ell}\pi_i(\vmu_i-\vmu_\ell^*)
    \right\rangle
    +
    \text{second-order remainder},
\]
where $\hpi_\ell = \sum_{i\in S_\ell}\pi_i$ is the total weights.
Therefore the mismatch between the model group and the true component is controlled by three aggregate quantities:
\[
    |\hpi_\ell-\pi_\ell^*|,
    \qquad
    \left\|
        \sum_{i\in S_\ell}\pi_i(\vmu_i-\vmu_\ell^*)
    \right\|,
    \qquad
    \sum_{i\in S_\ell}\pi_i\norm{\vmu_i-\vmu_\ell^*}^2 .
\]
These are exactly the quantities that replace the usual one-to-one
parametric error in the overparameterized setting.

This expansion also explains why the local dynamics are more delicate
under overparameterization.  The first-order term $\sum_{i \in S_\ell} \pi_i (\vmu_i-\vmu_\ell^*)$ may cancel even when the loss is nonzero, because several components in the same group can move in opposite directions.  When this cancellation occurs, the second-order quantity
$\sum_{i\in S_\ell}\pi_i\norm{\vmu_i-\vmu_\ell^*}^2$
becomes the dominant signal.  This higher-order behavior is one reason
the local convergence rate is slower in the overparameterized regime.  Similar phenomena appear in the
analysis of shallow neural networks
\citep{zhou2021local,zhou2024how,xu2023over} and 1-GMM
\citep{xu2024toward}, and also explain the slowdown in symmetric 2-GMMs
\citep{wu2019randomly,Dwivedi2018SingularityMA}, where the average
component error is zero by symmetry.

The formal version of the preceding Taylor comparison has one additional difficulty.  The terms in the KL gradient contain denominators involving the current model density \(p(\vx)\), and a direct expansion with \(1/p(\vx)\) is not uniformly stable in the tails.  To make the comparison rigorous, we interpolate between the truth and the current model by
\[
    p_\theta=(1-\theta)p_*+\theta p .
\]
For small fixed \(\theta\), the denominator \(p_\theta\) remains comparable to \(p_*\), so the Taylor remainders can first be controlled with \(p_\theta\) in place of \(p\).  We then transfer the estimate back to \(p\) using the homotopy \(\chi^2\)-type bound
\[
    \E_{p_\theta}\left[
        \left(\frac{p-p_*}{p_\theta}\right)^2
    \right]
    \lesssim
    \frac{\L(\vpi,\vmu)}{\theta}.
\]
This homotopy argument is the technique used in \cref{lem:local-linearization-remainders}.

After this reduction, the remaining task is an identifiability statement: small KL loss must imply that the three group-level quantities above are small.  This is precisely the content of the following theorem.  We write \(O_*(1)\) to hide polynomial factors in \(d,m,n,1/\pimins\); see \cref{thm:local-identifiability-log-separation} for the explicit dependence.

\begin{theorem}[Identifiability]\label{thm: id informal}
    Suppose separation $\Delta\gtrsim \sqrt{\log(mnd/\pimins)}$ and there exists partition $[n]=\bigcup_{l\in[m]} S_l$ such that $\forall i\in S_l, \norm{\vmu_i-\vmu_l^*}\leq B$. Then there exists a threshold $\eps_{\text{id}}=O_*(1)$ such that if $\L(\vpi, \vmu)\leq \eps\leq \eps_{\text{id}}$, then the following conditions are true:
    \begin{itemize}
        \item Weighted distance is small: For $\forall j\in[m]$,
        \[
            \sum_{i\in S_j}\pi_i\norm{\vmu_i-\vmu_j^*}^2\leq O_*(\sqrt{\eps}).
        \]
        \item Group weight is close: For $\forall j\in[m]$, 
        \[
            \left|\sum_{i\in S_j} \pi_i-\pi_j^*\right|\leq O_*(\sqrt{\eps}).
        \]
        \item Close-by weight is large: For each $j\in[m]$, let $S_j(\delta)\coloneqq\{i\in[n]: i\in S_j, \norm{\vmu_i-\vmu_j^*}\leq \delta\}$ be the set of components that are $\delta$-close to $\vmu_j^*$. There exists $\deltacls\coloneqq O_*(\eps^{1/4})$ such that
        \[
            \sum_{i\in S_j(\delta_{\rm close})} \pi_i\geq \frac{1}{2}\pi_j^*.
        \]
        \item Average component is close: For $\forall j\in[m]$,
        \[
            \norm{\sum_{i\in S_j}\pi_i(\vmu_i-\vmu_j^*)}{\leq}O_*(\sqrt{\eps}).
        \]
    \end{itemize}
\end{theorem}
We emphasize here that this separation requirement is much weaker than \cref{assump: delta}, and in fact matches the lower bound requirement on separation as in \cite{regev2017learning}. 

While this result seems natural to be expected, recall \termlink{challengeweight} that in the overparametrized regime ($n>m$) there is no classical one-to-one mapping as in the typical exact-parametrized regime ($m=n$). In particular, some $\vmu_i$ may never converge to $\vmu_\ell^*$, but instead $\pi_i\to 0$. Thus, it is challenging to show such identifiability result in the overparametrized regime. 

The $\sqrt{\eps}$ dependency in these bounds is important for our analysis when we do the analysis to control the cross term (See \cref{appendix: local}).
We prove this result by using a test function idea to draw a connection to other works on learning GMMs \citep{regev2017learning}, which may be of independent interest. We discuss more details in Section~\ref{sec: id}.

\section{Identifiability in overparametrized regime}\label{sec: id}

In this section, we outline the main ideas behind the proof of the identifiability result (Theorem~\ref{thm: id informal}), which plays a key role in local convergence proof. The detailed proof is presented in \cref{appendix:identifiability-new}.

Informally, identifiability means that when the loss is small, the parameters $\vpi, \vmu$ are expected to be close to the ground truth $\vpi^*,\vmu^*$. However, as discussed earlier, there is no one-to-one correspondence between parameters in the overparameterized setting. To address this, we prove the identifiability of GMMs by lower bounding the loss in terms of several augmented parameter errors that arise naturally in overparameterized setting. At a high level, we use the concept of \textit{test functions} to connect the KL loss to the quantities of interest, showing its identifiability.

While tensor decomposition and the method of moments have been widely used to provably learn GMMs \citep{anandkumar2014tensor, hsu2012learningmixturessphericalgaussians,liu2022clustering,kane2021robust,liu2023robustly}, our key contribution is connecting EM (minimizing KL divergence) to these problems, which is new to our knowledge.
Our approach is closely related in
spirit to the localizing equations used in \cite{regev2017learning}, but here the goal is different: we use these tests to extract parameter-level consequences from small KL loss.

\paragraph*{Test function}
Test functions allow us to extract information about a function $f$ by pairing it with a carefully chosen $g$ and analyzing their inner product $\langle f, g\rangle$. This idea appears in various fields (e.g., weak solutions in PDEs, hypothesis testing \citep{gretton2012kernel}, neural network analysis \citep{zhou2021local,zhou2024how}).

In our setting, where the loss $\L$ is small, we aim to construct a test function $g$ such that:
\begin{align*}
    \sqrt{\L} \norm{g}
    \myge{a} 
    \left|\int_\vx (p(\vx) - p_*(\vx)) g(\vx) \rd\vx\right|
    = \left|\E_{\vx\sim p}[g(\vx)] - \E_{\vx\sim p_*}[g(\vx)]\right|
    \myapprox{b}
    \text{Target term/property/...} 
\end{align*}
for a suitable norm $\norm{g}$ and target terms, for example, as appeared in Theorem~\ref{thm: id informal}. We explain each step in more details below.

\paragraph*{(a) Challenge in KL divergence}
The main difficulty in step (a) lies in handling the KL divergence. Prior work \citep{zhou2021local,zhou2024how} typically focuses on the squared loss. However, the KL divergence is harder to handle analytically. In this paper, we choose to consider bounded test function. By Pinsker's inequality, small \(\kl(p_*\|p)\) controls the total variation
distance between \(p_*\) and \(p\).  Therefore every bounded test function has close expectations under the two distributions:
\begin{restatable}{lemma}{lemboundedtestfun}
\label{lem:bounded-test-functions}
Recall \(\eps=\kl(p_*\|p)\). For any bounded measurable test function
\(f:\R^d\to\R^q\) with \(q\ge 1\), we have
\[
    \norm{\E_p f-\E_{p_*}f}
    \le
    \sqrt{2}\,\norm{f}_\infty\sqrt{\eps},
    \qquad
    \norm{f}_\infty:=\sup_{\vx\in\R^d}\norm{f(\vx)}.
\]
\end{restatable}

We can now conclude step (a) by observing that:
\begin{align*}
    \left|\int_\vx (p(\vx) - p_*(\vx)) g(\vx) \rd\vx\right|
    = \left|\E_{\vx\sim p}g(\vx) - \E_{\vx\sim p_*}g(\vx)\right|
    \le 
        \sqrt{2\L}
        \norm{g}_\infty.
\end{align*}
This enables the use of test function in the next step and we believe similar idea may generalize to other loss functions beyond KL loss and square loss.

\paragraph*{(b) Choice of test function.}
We next explain the choice of test functions.
To illustrate the idea, we consider the test function $g_{\ell,0}(\vx):=\ind_{\Omega_\ell}(\vx)$, and show how to use it to bound $|\hpi_\ell - \pi_\ell^*|$. Here $R=\Theta(\sqrt{d+\log(mnd/\pimins)})$ is large enough and the region of interest is 
\[
    \Omega_\ell:=V_\ell\cap B(\vmu_\ell^*,R),
    \quad
    V_\ell
    :=
    \left\{
        \vx:
        \norm{\vx-\vmu_\ell^*}
        \le
        \norm{\vx-\vmu_j^*}
        \quad
        \forall j\neq \ell
    \right\}.
\]
Intuitively, $\Omega_\ell$ captures almost all samples from the \(\ell\)-th true Gaussian and has exponentially small overlap $e^{-\Delta^2}$ with the other true Gaussians. Under logarithmic separation, this exponentially small
quantity absorbs the polynomial factors that appear in the estimates. Therefore, we have
\begin{align*}
    \int_\vx (p(\vx) - p_*(\vx)) g_{\ell,0}(\vx) \rd\vx
    = \E_{\vx\sim p}[g_{\ell,0}(\vx)] - \E_{\vx\sim p_*}[g_{\ell,0}(\vx)]
    \approx \hpi_\ell - \pi_\ell^*
    .
\end{align*}
See \cref{lem:local-gaussian-estimates-detailed} for the precise estimates. Then combining step (a)(b), we have
\begin{align*}
    \sqrt{\L} 
    \gtrsim 
    \left|\int_\vx (p(\vx) - p_*(\vx)) g_{\ell,0}(\vx) \rd\vx\right|
    \gtrsim \left|\hpi_\ell - \pi_\ell^*\right|.
\end{align*}
Similarly, we could get the desired bound for other terms using test functions. See \cref{appendix:identifiability-new} for details.

\section{Conclusion}

In this paper, we present the first global convergence result for gradient EM applied to general overparametrized Gaussian mixtures. Our analysis introduces new techniques, such as the use of test functions and homotopy method, for studying gradient EM in the overparameterized regime. We believe this work represent an important first step toward a deeper understanding of the EM algorithm to broader settings. Moreover, we expect that our techniques can be extended to other related settings, including learning latent variable models and single- or multi-index models. Our analysis is limited in the sense that it focuses on  well-separated GMMs. Extending the result to settings with weaker separation conditions, improved time and sample complexity, or the classical EM algorithm itself are interesting future directions.

\section*{Acknowledgements}
MF thanks Jason Altschuler for helpful discussions. SSD acknowledges the support of  NSF DMS 2134106, NSF IIS 2143493, NSF IIS 2229881, Sloan Fellowship, and the AI2050 program at Schmidt Sciences.
MF, MZ, and WX acknowledge the support of NSF TRIPODS II DMS 2023166. The work of MF was also supported by awards NSF CCF 2212261 and NSF CCF 2312775.

\newpage

\appendix

\numberwithin{theorem}{section}

\makeatletter
\renewcommand{\theHtheorem}{appendix.\theHsection.\arabic{theorem}}
\renewcommand{\theHlemma}{appendix.\theHsection.\arabic{lemma}}
\renewcommand{\theHcorollary}{appendix.\theHsection.\arabic{corollary}}
\renewcommand{\theHclaim}{appendix.\theHsection.\arabic{claim}}
\renewcommand{\theHproposition}{appendix.\theHsection.\arabic{proposition}}
\renewcommand{\theHremark}{appendix.\theHsection.\arabic{remark}}
\makeatother

\renewcommand{\theequation}{\thesection.\arabic{equation}}
\renewcommand{\theHequation}{appendix.\thesection.\arabic{equation}}

\numberwithin{equation}{section}

\section{Additional background and preliminaries}
In this section, we present several useful facts and the proof of the claims in main text. We give gradient related facts in Appendix~\ref{appendix: gradient prelim}, and some comments on Algorithm~\ref{alg} in Appendix~\ref{appendix: alg}.

\subsection{Gradient related facts}\label{appendix: gradient prelim}
We start with the classic Stein's lemma.
\begin{lemma}[\cite{stein1981estimation}]\label{lem: stein identity}
    For $\vx \sim \gN(\vmu, \sigma^2 \mI_d)$ and differentiable function $g : \R^d \to \R$ we have
    \[
    \E_\vx[g(\vx)(\vx - \vmu)] = \sigma^2 \E_\vx[\nabla_\vx g(\vx)],\] 
    if the two expectations in the above identity exist.
\end{lemma} 

We use Stein's lemma to derive an alternative form of the gradient, similar to the approach in \cite{xu2024toward} for the 1-GMM case. This new formulation is better suited for our analysis.
\lemgradientform*
\begin{proof}
    We have
    \begin{align*} 
        \nabla_{\vmu_i} \L(\vmu,\vpi) 
        = \sum_{j\in [m]} \pi_j^* \E_{\vx\sim j}\left[\psi_i(\vx)(\vmu_i-\vx)\right].
    \end{align*}
    We use Stein's identity (Lemma~\ref{lem: stein identity}) to transform each summation term in RHS as
    \begin{align*}
         \E_{x\sim j}\left[\psi_i(\vx)(\vmu_i-\vx)\right] 
         &= -\E_{\vx\sim j}\left[\psi_i(\vx)(\vx-\vmu_j^*)\right]+\E_{\vx\sim j}\left[\psi_i(\vx)(\vmu_i-\vmu_j^*)\right]\\
         &=-\E_{\vx\sim j}\left[\nabla_\vx \psi_i(\vx)\right]+(\vmu_i-\vmu_j^*)\E_{\vx\sim j}\left[\psi_i(\vx)\right].
    \end{align*}
    
    Next we calculate
    \begin{align*}
        \nabla_x \psi_i(x)
        &=\psi_i(\vx)(\vmu_i-\vx)-\psi_i(\vx)\sum_{k\in[n]}\psi_k(\vx)(\vmu_k-\vx)
        =\psi_i(\vx)\left(\vmu_i-\sum_{k\in[n]}\psi_k(\vx)\vmu_k\right),
    \end{align*}
    so
    \begin{align*}
        \E_{\vx\sim j}\left[\psi_i(\vx)(\vmu_i-\vx)\right]
        &=\E_{\vx\sim j}\left[\psi_i(\vx)\left(\sum_{k\in[n]} \psi_k(\vx)\vmu_k-\vmu_i+(\vmu_i-\vmu_j^*)\right)\right]\\
        &=\E_{\vx\sim j}\left[\psi_i(\vx)\sum_{k\in[n]} \psi_k(\vx)(\vmu_k-\vmu_j^*)\right].
    \end{align*}
    
    The gradient could thus be rewritten as
    \begin{align*}
        \nabla_{\vmu_i} \L(\vmu,\vpi)
        = \sum_{j\in [m]} \pi_j^* \E_{\vx\sim j}\left[\psi_i(\vx)(\vmu_i-\vx)\right] = \sum_{j\in [m]} \pi_j^* \E_{\vx\sim j}\left[\psi_i(\vx)\sum_{k\in[n]} \psi_k(\vx)(\vmu_k-\vmu_j^*) \right].
    \end{align*}
    
\end{proof}

\subsection{Comments on the optimality certificate in the algorithm}
\label{appendix: alg}

Here, we clarify what we mean by ``optimality'' in line 4 of Algorithm~\ref{alg}.  For fixed
\(\vmu\), the weight subproblem
\[
    \min_{\vpi\in\gP}\L(\vpi,\vmu)
\]
is convex.  We measure its accuracy by the simplex dual gap
\begin{equation}
    G_\pi(\vpi,\vmu)
    :=
    \max_{\bar\vpi\in\gP}
    \left\langle
        \nabla_{\vpi}\L(\vpi,\vmu),
        \vpi-\bar\vpi
    \right\rangle .
    \label{eq:weight-dual-gap}
\end{equation}
Thus, at a fixed \(\vmu\), the condition \(G_\pi(\vpi,\vmu)\le \eps\)
is our first-order certificate for the convex weight subproblem. Along the
algorithm, when we say that the weight update is \(\eps\)-optimal, we require
both
\[
    G_\pi(\vpi^+,\vmu)\le \eps,
    \qquad
    \L(\vpi^+,\vmu)\le \L(\vpi,\vmu),
\]
where \(\vpi\) is the current weight vector and \(\vpi^+\) is the updated one.
The second condition is the monotonicity condition used in the local and global
phase proofs. It is automatic for an exact minimizer, and can be enforced for
standard convex solvers by accepting only monotone iterates, for example by a
line search.

Since $\nabla_{\pi_i}\L(\vpi,\vmu)
    =
    -\E_{p^*}\frac{\phi(\vmu_i;\vx)}{p(\vx)}$ and $\sum_i\pi_i\nabla_{\pi_i}\L(\vpi,\vmu)=-1$,
the simplex structure gives the explicit form
\[
    G_\pi(\vpi,\vmu)
    =
    -1-\min_{i\in[n]}\nabla_{\pi_i}\L(\vpi,\vmu)
    =
    \max_{i\in[n]}
    \left[-\left(\nabla_{\pi_i}\L(\vpi,\vmu)+1\right)\right].
\]
Thus \(G_\pi(\vpi,\vmu)\le\eps\) is equivalent to the one-sided approximate
KKT condition
\begin{equation}
    \nabla_{\pi_i}\L(\vpi,\vmu)+1\ge -\eps
    \qquad\text{for every }i\in[n].
    \label{eq:weight-dual-gap-one-sided}
\end{equation}

\begin{lemma}[Consequences of the weight dual gap]
\label{lem:weight-dual-gap-consequences}
    Fix \(\vmu\), and let \(\vpi\in\gP\).  Suppose
    \(G_\pi(\vpi,\vmu)\le\eps\).  Then
    \begin{equation*}
        \L(\vpi,\vmu)
        \le
        \inf_{\bar\vpi\in\gP}\L(\bar\vpi,\vmu)+\eps.
    \end{equation*}
    Moreover, if \(\psi_i\) denotes the posterior under \((\vpi,\vmu)\), then
    \begin{equation*}
        \sum_{i=1}^n
        \abs{\pi_i-\E_{p^*}\psi_i}
        =
        \sum_{i=1}^n
        \pi_i
        \abs{\nabla_{\pi_i}\L(\vpi,\vmu)+1}
        \le
        2\eps.
    \end{equation*}
\end{lemma}

\begin{proof}
    Let
    \[
        g_i:=\nabla_{\pi_i}\L(\vpi,\vmu),
        \qquad
        a_i:=g_i+1 .
    \]
    By \eqref{eq:weight-dual-gap-one-sided}, \(a_i\ge-\eps\) for every
    \(i\).  Also \(\sum_i\pi_i a_i=0\), because
    \(\sum_i\pi_i g_i=-1\).

    The objective-gap bound follows directly from convexity: for every
    \(\bar\vpi\in\gP\),
    \[
        \L(\vpi,\vmu)-\L(\bar\vpi,\vmu)
        \le
        \left\langle
            \nabla_{\vpi}\L(\vpi,\vmu),
            \vpi-\bar\vpi
        \right\rangle
        \le
        G_\pi(\vpi,\vmu)
        \le
        \eps .
    \]
    Taking the infimum over \(\bar\vpi\in\gP\) gives
    the first claim.

    For the complementarity bound, since \(a_i\ge-\eps\), we have $\sum_i\pi_i(a_i)_-\le \eps$.
    The weighted average of \(a_i\) is zero, so the positive and negative
    parts have the same weighted mass: $\sum_i\pi_i(a_i)_+
        =
        \sum_i\pi_i(a_i)_-$.
    Hence
    \[
        \sum_i\pi_i|a_i|
        \le
        2\eps .
    \]
    Finally,
    \[
        \pi_i(\nabla_{\pi_i}\L+1)
        =
        \pi_i-\E_{p^*}\psi_i ,
    \]
    which proves the second claim.
\end{proof}

\section{Local robust identifiability}
\label{appendix:identifiability-new}

In this section we prove a local robust identifiability result under only a logarithmic separation
condition. This is weaker than \cref{assump: delta} that requires the well-separation. Such logarithmic separation
condition is in fact the best one could hope for parameter recovery under polynomial sample complexity \citep{regev2017learning}.

Our argument below is based on bounded test functions localized around the true centers. The
main result is \cref{thm:local-identifiability-log-separation}; the form used in the main text follows
immediately from \cref{cor:local-identifiability-consequences}.

\paragraph{Notation.}
Recall the Voronoi partition of the model components by assigning each component to a nearest true center, with any deterministic rule to break ties:
\[
    S_\ell
    :=
    \left\{
        i\in[n]:
        \norm{\vmu_i-\vmu_\ell^*}
        \le
        \norm{\vmu_i-\vmu_j^*}
        \text{ for every }j\in[m]
    \right\}.
\]
Thus \(S_\ell\) consists of the model components whose nearest true center is \(\vmu_\ell^*\).  We further define the near group \(\bar S_\ell\) by keeping
only the components within the identifiability radius \(B_{\rm id}\) (defined later in \cref{thm:local-identifiability-log-separation}), and let
\(\bar S_0\) collect all remaining components:
\[
    \bar S_\ell
    :=
    \{i\in S_\ell:\norm{\vmu_i-\vmu_\ell^*}\le B_{\rm id}\},
    \qquad
    \bar S_0:=[n]\setminus\bigcup_{\ell=1}^m\bar S_\ell .
\]
Let \(\hpi_\ell:=\sum_{i\in\bar S_\ell}\pi_i\) be the weight of the
\(\ell\)-th near group, and let
\(\pi_0:=\sum_{i\in\bar S_0}\pi_i\) be the null-group weight.  For each
\(\ell\in[m]\), define the zeroth-, first-, and second-order centered
near-group errors by
\[
    \Gamma_{\ell,0}:=\hpi_\ell-\pi_\ell^*,
    \qquad
    \Gamma_{\ell,1}:=
    \sum_{i\in\bar S_\ell}\pi_i(\vmu_i-\vmu_\ell^*),
    \qquad
    \Gamma_{\ell,2}:=
    \sum_{i\in\bar S_\ell}
    \pi_i\norm{\vmu_i-\vmu_\ell^*}^2 .
\]
For a vector-valued map \(F:\R^d\to\R^r\), we write
\(\nabla F\) for its Jacobian, whose \(a\)-th row is \(\nabla F_a\).  We
write \(\nabla^2F\) for the componentwise Hessian, viewed as a bilinear map
from \(\R^d\times\R^d\) to \(\R^r\), with the corresponding operator norm.
For scalar-valued maps this agrees with the usual gradient and Hessian.

We now present the main result of this section.
\begin{theorem}[Identifiability]
\label{thm:local-identifiability-log-separation}
There exist universal constants \(b_0,r_0,C_{\rm sep},C>0\) such that the
following holds.  Let
\(\Lambda:=e+\frac{100mn(d+1)}{\pimins}\),
\(B_{\rm id}:=b_0\sqrt{\log\Lambda}\), and
\(R:=r_0\sqrt{d+B_{\rm id}^2}\). If \(m\ge2\), assume
\[
    \Delta\ge C_{\rm sep}\sqrt{\log\Lambda}
    \asymp \sqrt{\log(mnd/\pimins)}.
\]
When \(m=1\), this separation assumption is vacuous.
Then, for loss \(\eps:=\L(\vpi,\vmu)=\kl(p_*\|p)\), we have
\[
    \pi_0+
    \sum_{\ell=1}^m
    \left(
        |\Gamma_{\ell,0}|+\norm{\Gamma_{\ell,1}}+\Gamma_{\ell,2}
    \right)
    \le
    \Xi_{\rm id}\sqrt{\eps},
\]
where
\[
    \Xi_{\rm id}
    :=Cm(1+R)^3
    \lesssim
    m\left(
        1+d+\log\left(e+\frac{100mn(d+1)}{\pimins}\right)
    \right)^{3/2}.
\]
\end{theorem}

As a direct corollary, \cref{thm:local-identifiability-log-separation}
implies the full-partition estimates used in the local convergence proof.
The statement is written in terms of the full Voronoi groups \(S_\ell\); the
near/null decomposition is used only in the proof.

\begin{corollary}[\cref{thm: id informal} restated, identifiability consequences for the local phase]
\label{cor:local-identifiability-consequences}
    Under the assumptions of
    \cref{thm:local-identifiability-log-separation}, let
    \(\Xi_{\rm id}\) be the constant in that theorem, so that
    \[
        \Xi_{\rm id}
        \lesssim
        m\left(
            1+d+\log\left(e+\frac{100mn(d+1)}{\pimins}\right)
        \right)^{3/2}.
    \]
    Suppose the loss \(\eps:=\L(\vpi,\vmu)=\kl(p_*\|p)\) is small enough
    that \(\Xi_{\rm id}\sqrt{\eps}\le \pimins/4\).  Suppose also that every
    component in a full Voronoi group is within distance \(B\) of its assigned
    true center; that is,
    \[
        \norm{\vmu_i-\vmu_\ell^*}\le B
        \qquad
        \text{for every }\ell\in[m]\text{ and every }i\in S_\ell .
    \]
    Then the following estimates hold.

    \begin{itemize}
        \item (Group weight.)  Each full group has almost the correct total
        weight:
        \begin{equation}
        \label{eq: Identifiability weight bound}
            \left|
                \sum_{i\in S_\ell}\pi_i-\pi_\ell^*
            \right|
            \lesssim
            \Xi_{\rm id}\sqrt{\eps}
            \qquad\text{for every }\ell\in[m].
        \end{equation}

        \item (Weighted distance.)  The weighted squared distance to the
        corresponding true center is small in each full group:
        \begin{equation}
        \label{eq: Identifiability 2nd order term bound}
            \sum_{i\in S_\ell}
            \pi_i\norm{\vmu_i-\vmu_\ell^*}^2
            \lesssim
            \Xi_{\rm id}(1+B^2)\sqrt{\eps}
            \qquad\text{for every }\ell\in[m].
        \end{equation}

        \item (Average component.)  The weighted average displacement in each
        full group is small:
        \begin{equation}
        \label{eq: Identifiability linear term bound}
            \norm{
                \sum_{i\in S_\ell}
                \pi_i(\vmu_i-\vmu_\ell^*)
            }
            \lesssim
            \Xi_{\rm id}(1+B^2)\sqrt{\eps}
            \qquad\text{for every }\ell\in[m].
        \end{equation}

        \item (Close-by weight.)  For every \(\ell\in[m]\), define
        \(S_\ell(\delta):=\{i\in S_\ell:
        \norm{\vmu_i-\vmu_\ell^*}\le\delta\}\), and set
        \(\delta_{\rm close}:=
        2\sqrt{\Xi_{\rm id}/\pimins}\,\eps^{1/4}\).  Then each true center
        has at least half of its true weight represented by close components:
        \begin{equation}
        \label{eq: Identifiability close points}
            \sum_{i\in S_\ell(\delta_{\rm close})}\pi_i
            \ge
            \frac12\pi_\ell^*
            \qquad\text{for every }\ell\in[m].
        \end{equation}
    \end{itemize}
\end{corollary}

\begin{proof}
    We transfer the near-group estimates from
    \cref{thm:local-identifiability-log-separation} to the full groups
    \(S_\ell\).  Let \(N_\ell:=S_\ell\setminus\bar S_\ell\) be the part of
    the full group classified as null by the identifiability theorem.  Then
    \(N_\ell\subset\bar S_0\), and the theorem gives
    \[
        \pi_0+
        \sum_{\ell=1}^m
        \left(
            \abs{\Gamma_{\ell,0}}
            +\norm{\Gamma_{\ell,1}}
            +\Gamma_{\ell,2}
        \right)
        \le
        \Xi_{\rm id}\sqrt{\eps},
        \qquad
        \pi_0:=\sum_{i\in\bar S_0}\pi_i .
    \]
    Here the \(\Gamma_{\ell,k}\)'s are the centered errors over the near
    group \(\bar S_\ell\).

    We first add the null components back into the full groups.  Since every
    component in \(S_\ell\) is within distance \(B\) of \(\vmu_\ell^*\), for
    each \(\ell\),
    \[
    \begin{aligned}
        \left|\sum_{i\in S_\ell}\pi_i-\pi_\ell^*\right|
        &\le
        \abs{\Gamma_{\ell,0}}
        +
        \sum_{i\in N_\ell}\pi_i,\\
        \norm{
            \sum_{i\in S_\ell}
            \pi_i(\vmu_i-\vmu_\ell^*)
        }
        &\le
        \norm{\Gamma_{\ell,1}}
        +
        B\sum_{i\in N_\ell}\pi_i,\\
        \sum_{i\in S_\ell}
        \pi_i\norm{\vmu_i-\vmu_\ell^*}^2
        &\le
        \Gamma_{\ell,2}
        +
        B^2\sum_{i\in N_\ell}\pi_i .
    \end{aligned}
    \]
    The identifiability bound, together with
    \(\sum_{i\in N_\ell}\pi_i\le\pi_0\), proves the group-weight,
    weighted-distance, and average-component estimates.  For the average
    component bound we also use \(B\le 1+B^2\).

    It remains to prove the close-by weight lower bound.  The near-group
    estimates give
    \[
        \left|
            \sum_{i\in\bar S_\ell}\pi_i-\pi_\ell^*
        \right|
        \le
        \Xi_{\rm id}\sqrt{\eps},
        \qquad
        \sum_{i\in\bar S_\ell}
        \pi_i\norm{\vmu_i-\vmu_\ell^*}^2
        \le
        \Xi_{\rm id}\sqrt{\eps}.
    \]
    If \(\eps=0\), then all positive mass in \(\bar S_\ell\) lies at
    distance zero from \(\vmu_\ell^*\), and the claim follows.  Otherwise,
    Markov's inequality and the definition of \(\delta_{\rm close}\) give
    \[
        \sum_{\substack{i\in\bar S_\ell:
        \norm{\vmu_i-\vmu_\ell^*}>\delta_{\rm close}}}\pi_i
        \le
        \frac{\Xi_{\rm id}\sqrt{\eps}}{\delta_{\rm close}^2}
        =
        \frac{\pimins}{4}
        \le
        \frac{\pi_\ell^*}{4}.
    \]
    Hence
    \[
    \begin{aligned}
        \sum_{i\in S_\ell(\delta_{\rm close})}\pi_i
        &\ge
        \sum_{\substack{i\in\bar S_\ell:
        \norm{\vmu_i-\vmu_\ell^*}\le\delta_{\rm close}}}\pi_i
        \ge
        \pi_\ell^*
        -
        \Xi_{\rm id}\sqrt{\eps}
        -
        \frac14\pi_\ell^*
        \ge
        \frac12\pi_\ell^*,
    \end{aligned}
    \]
    where the last step uses
    \(\Xi_{\rm id}\sqrt{\eps}\le\pimins/4\le\pi_\ell^*/4\).
\end{proof}

\subsection{Proof of \cref{thm:local-identifiability-log-separation}}
The proof uses bounded test functions to connect the KL loss to the desired grouped quantities. We first show a general bounded-test estimate, then introduce the localized tests, and finally prove the local Gaussian estimates needed to evaluate these tests near each true center.

Throughout this section, for any test function \(f\), we write
\[
    \E_\vmu f
    :=
    \E_{\vZ\sim\N(0,\mI_d)} f(\vmu+\vZ),
    \qquad
    \E_\ell f
    :=
    \E_{\vZ\sim\N(0,\mI_d)} f(\vmu_\ell^*+\vZ).
\]

\paragraph{Bounded test functions.}
The following lemma turns small KL divergence into control of bounded test functions. The proof is deferred to \cref{sec: id omit proof}.

\lemboundedtestfun*

\paragraph{Choose test functions as localized tests.}
We use the following bounded test functions:
\[
    g_{\ell,0}(\vx):=\ind_{\Omega_\ell}(\vx),
    \qquad
    g_{\ell,1}(\vx):=(\vx-\vmu_\ell^*)\ind_{\Omega_\ell}(\vx),
    \qquad
    g_{\rm quad}(\vx):=\min\{D(\vx)^2,R^2\}-d.
\]
Here \(g_{\ell,0}\) measures the local mass near \(\vmu_\ell^*\), \(g_{\ell,1}\) measures the local
centered first moment around \(\vmu_\ell^*\), and \(g_{\rm quad}\) measures the clipped squared
distance to the nearest true center. The local region and nearest-center distance are defined by
\[
    \Omega_\ell:=V_\ell\cap B(\vmu_\ell^*,R),
    \qquad
    D(\vx):=\min_{j\in[m]}\norm{\vx-\vmu_j^*}.
\]
The set \(V_\ell\) is the Voronoi cell of \(\vmu_\ell^*\), with deterministic tie-breaking applied to \(\argmin_j \norm{\vx-\vmu_j^*}\):
\[
    V_\ell
    :=
    \left\{
        \vx:
        \norm{\vx-\vmu_\ell^*}
        \le
        \norm{\vx-\vmu_j^*}
        \quad
        \forall j\neq \ell
    \right\}.
\]
Thus \(V_1,\ldots,V_m\) are disjoint and cover \(\R^d\). Since
\(\Omega_\ell\subset B(\vmu_\ell^*,R)\) and \(R^2\ge d\), we have
\[
    \norm{g_{\ell,0}}_\infty\le 1,
    \qquad
    \norm{g_{\ell,1}}_\infty\le R,
    \qquad
    \norm{g_{\rm quad}}_\infty\le R^2.
\]
Therefore, by \cref{lem:bounded-test-functions}, for every \(\ell\in[m]\),
\[
\begin{aligned}
    \left|
        \E_p g_{\ell,0}-\E_{p_*}g_{\ell,0}
    \right|
    &\lesssim
    \sqrt{\eps},
    \qquad
    \left\|
        \E_p g_{\ell,1}-\E_{p_*}g_{\ell,1}
    \right\|
    \lesssim
    R\sqrt{\eps},
    \qquad
    \left|
        \E_p g_{\rm quad}-\E_{p_*}g_{\rm quad}
    \right|
    \lesssim
    R^2\sqrt{\eps}.
\end{aligned}
\]

We next show that the localized tests recover the corresponding full Gaussian moments for same-group components up to a small error, while far-group contributions are negligible. The proof is deferred to \cref{sec: id omit proof}.
\begin{restatable}[Local Gaussian estimates]{lemma}{lemlocalgaussianestimate}
\label{lem:local-gaussian-estimates-detailed}
There is a quantity
\[
    \tau
    :=
    C(1+R)^8
    \left[
        e^{-c(R-B_{\rm id})^2}
        +
        \ind_{\{m\ge2\}}\,
        m e^{-c(\Delta-2B_{\rm id})^2}
        +
        m e^{-cB_{\rm id}^2}
    \right].
\]
By choosing
\(b_0,r_0,C_{\rm sep}\) sufficiently large in the definitions of \(B_{\rm id},R,\Delta\), we may assume
\(m(1+R)\tau\le c\) for a sufficiently small universal constant \(c>0\).

For every \(\ell\in[m]\) and every \(\norm{\vh}\le B_{\rm id}\), with \(\vZ\sim\N(0,\mI_d)\), define the
same-group quantities
\[
    G_{\ell,0}(\vh)
    :=
    \E_\vZ\left[g_{\ell,0}(\vmu_\ell^*+\vh+\vZ)\right],
    \quad
    G_{\ell,1}(\vh)
    :=
    \E_\vZ\left[g_{\ell,1}(\vmu_\ell^*+\vh+\vZ)\right],
    \quad
    G_{\ell,2}(\vh)
    :=
    \E_\vZ\left[g_{\rm quad}(\vmu_\ell^*+\vh+\vZ)\right].
\]
Then
\begin{align*}
    G_{\ell,0}(\vh)
    &=
    1+\bigl(G_{\ell,0}(0)-1\bigr)
    +
    \langle \nabla G_{\ell,0}(0),\vh\rangle
    +
    \rho_{\ell,0}(\vh),\\
    G_{\ell,1}(\vh)
    &=
    \vh
    +
    G_{\ell,1}(0)
    +
    \bigl(\nabla G_{\ell,1}(0)-\mI_d\bigr)\vh
    +
    \rho_{\ell,1}(\vh),\\
    G_{\ell,2}(\vh)
    &=
    G_{\ell,2}(0)
    +
    \langle \nabla G_{\ell,2}(0),\vh\rangle
    +
    \norm{\vh}^2
    +
    \rho_{\ell,2}(\vh),
\end{align*}
where
\begin{align*}
    |G_{\ell,0}(0)-1|+\norm{\nabla G_{\ell,0}(0)}
    &\le \tau,
    \qquad
    |\rho_{\ell,0}(\vh)|\le \tau\norm{\vh}^2,\\
    \norm{G_{\ell,1}(0)}
    +
    \norm{\nabla G_{\ell,1}(0)-\mI_d}_{\rm op}
    &\le
    \tau,
    \qquad
    \norm{\rho_{\ell,1}(\vh)}
    \le
    \tau\norm{\vh}^2,\\
    |G_{\ell,2}(0)|+\norm{\nabla G_{\ell,2}(0)}
    &\le \tau,
    \qquad
    |\rho_{\ell,2}(\vh)|\le \tau\norm{\vh}^2.
\end{align*}

For \(j\neq \ell\), define the far-group quantities
\[
    G_{\ell\leftarrow j,0}(\vh)
    :=
    \E_\vZ\left[g_{\ell,0}(\vmu_j^*+\vh+\vZ)\right],
    \qquad
    G_{\ell\leftarrow j,1}(\vh)
    :=
    \E_\vZ\left[g_{\ell,1}(\vmu_j^*+\vh+\vZ)\right].
\]
Then, for every \(\norm{\vh}\le B_{\rm id}\),
\begin{align*}
    |G_{\ell\leftarrow j,0}(0)|
    +
    \norm{\nabla G_{\ell\leftarrow j,0}(0)}
    +
    \sup_{\norm{\vt}\le B_{\rm id}}\norm{\nabla^2G_{\ell\leftarrow j,0}(\vt)}_{\rm op}
    &\le
    \tau,\\
    \norm{G_{\ell\leftarrow j,1}(0)}
    +
    \norm{\nabla G_{\ell\leftarrow j,1}(0)}_{\rm op}
    +
    \sup_{\norm{\vt}\le B_{\rm id}}\norm{\nabla^2G_{\ell\leftarrow j,1}(\vt)}_{\rm op}
    &\le
    \tau.
\end{align*}

Finally, if \(\min_{i\in[m]}\norm{\vmu-\vmu_i^*}>B_{\rm id}\), then
\[
    \E_\vmu g_{\rm quad}\gtrsim B_{\rm id}^2.
\]
\end{restatable}

Now we are ready to prove \cref{thm:local-identifiability-log-separation}. The proof uses the three
localized tests above to control the three grouped errors: the mass test controls the zeroth-order
errors \(\Gamma_{\ell,0}\), the local first-moment test controls the centered first moments \(\Gamma_{\ell,1}\), and the clipped quadratic
test controls both the second moments \(\Gamma_{\ell,2}\) and the null mass $\pi_0$.
\begin{proof}[Proof of \cref{thm:local-identifiability-log-separation}]
The argument is a stability estimate for the localized tests. The bounded-test lemma \cref{lem:bounded-test-functions} first controls
the observable discrepancies of these tests under \(p\) and \(p_*\). The local Gaussian estimates \cref{lem:local-gaussian-estimates-detailed}
then identify the leading terms of these discrepancies with the grouped errors, up to \(O(\tau)\)
remainders, which will be absorbed at the end.

Define the total error
\[
    \mathcal E
    :=
    \pi_0+
    \sum_{\ell=1}^m
    \left(
        |\Gamma_{\ell,0}|+\norm{\Gamma_{\ell,1}}+\Gamma_{\ell,2}
    \right).
\]
For \(i\in \bar{S}_j\), write \(\vh_i^{(j)}:=\vmu_i-\vmu_j^*\). By construction,
\(\norm{\vh_i^{(j)}}\le B_{\rm id}\).

\paragraph{0th-order $\Gamma_{\ell,0}$ bound.}
We first use the localized mass test \(g_{\ell,0}=\ind_{\Omega_\ell}\) to control the zeroth-order
group errors \(\Gamma_{\ell,0}=\hpi_\ell-\pi_\ell^*\). Let
\(\delta_{\ell,0}:=\E_p g_{\ell,0}-\E_{p_*}g_{\ell,0}\). By
\cref{lem:bounded-test-functions}, \(|\delta_{\ell,0}|\lesssim\sqrt{\eps}\). Decompose
\[
\begin{aligned}
    \delta_{\ell,0}
    &=
    \left[
        \sum_{i\in \bar{S}_\ell}\pi_i\E_{\vmu_i}g_{\ell,0}
        -
        \pi_\ell^*\E_\ell g_{\ell,0}
    \right]
    +
    \sum_{j\neq \ell}
    \left[
        \sum_{i\in \bar{S}_j}\pi_i\E_{\vmu_i}g_{\ell,0}
        -
        \pi_j^*\E_j g_{\ell,0}
    \right]
    +
    N_{\ell,0},
\end{aligned}
\]
where \(N_{\ell,0}:=\sum_{i\in \bar{S}_0}\pi_i\E_{\vmu_i}g_{\ell,0}\ge 0\) is the null contribution to
the local mass test. The same-group contribution is
\[
\begin{aligned}
    \sum_{i\in \bar{S}_\ell}\pi_i\E_{\vmu_i}g_{\ell,0}
    -
    \pi_\ell^*\E_\ell g_{\ell,0}
    &=
    \sum_{i\in \bar{S}_\ell}\pi_i G_{\ell,0}(\vh_i^{(\ell)})
    -
    \pi_\ell^*G_{\ell,0}(0)\\
    &=
    \Gamma_{\ell,0}
    +
    \bigl(G_{\ell,0}(0)-1\bigr)\Gamma_{\ell,0}
    +
    \langle \nabla G_{\ell,0}(0),\Gamma_{\ell,1}\rangle
    +
    \sum_{i\in \bar{S}_\ell}\pi_i\rho_{\ell,0}(\vh_i^{(\ell)})\\
    &\myeq{a}
    \Gamma_{\ell,0}
    \pm
    O(\tau)\left(
        |\Gamma_{\ell,0}|+\norm{\Gamma_{\ell,1}}+\Gamma_{\ell,2}
    \right).
\end{aligned}
\]
where (a) uses the same-group expansion of \(G_{\ell,0}\) from
\cref{lem:local-gaussian-estimates-detailed}. For \(j\neq\ell\), the far-group contribution is small:
\[
\begin{aligned}
    \sum_{i\in \bar{S}_j}\pi_i\E_{\vmu_i}g_{\ell,0}
    -
    \pi_j^*\E_j g_{\ell,0}
    &=
    \sum_{i\in \bar{S}_j}\pi_i G_{\ell\leftarrow j,0}(\vh_i^{(j)})
    -
    \pi_j^*G_{\ell\leftarrow j,0}(0)\\
    &=
    \Gamma_{j,0}G_{\ell\leftarrow j,0}(0)
    +
    \langle \nabla G_{\ell\leftarrow j,0}(0),\Gamma_{j,1}\rangle
    \pm
    O(\tau\Gamma_{j,2})\\
    &\myeq{a}
    \pm
    O(\tau)\left(
        |\Gamma_{j,0}|+\norm{\Gamma_{j,1}}+\Gamma_{j,2}
    \right).
\end{aligned}
\]
where (a) uses the far-group derivative bounds for \(G_{\ell\leftarrow j,0}\) from
\cref{lem:local-gaussian-estimates-detailed}. Combining the
same-group, far-group, and null contributions gives
\[
    \delta_{\ell,0}=\Gamma_{\ell,0}+N_{\ell,0}+e_{\ell,0},
    \qquad
    \sum_{\ell=1}^m |e_{\ell,0}|\lesssim m\tau\mathcal E,
    \qquad
    \sum_{\ell=1}^m |N_{\ell,0}|\le \pi_0.
\]
The last inequality follows from the disjointness of the regions \(\Omega_\ell\), which implies
\(\sum_{\ell=1}^m g_{\ell,0}(\vx)\le 1\). Summing over \(\ell\) and using
\(|\delta_{\ell,0}|\lesssim\sqrt{\eps}\), we obtain
\[
    \sum_{\ell=1}^m |\Gamma_{\ell,0}|
    \lesssim
    m\sqrt{\eps}
    +
    \pi_0
    +
    m\tau\mathcal E.
\]

\paragraph{1st-order $\Gamma_{\ell,1}$ bound.}
We next use the localized first-moment test
\(g_{\ell,1}=(\vx-\vmu_\ell^*)\ind_{\Omega_\ell}(\vx)\) to control the first-order group errors
\(\Gamma_{\ell,1}=\sum_{i\in \bar{S}_\ell}\pi_i(\vmu_i-\vmu_\ell^*)\). Let \(\delta_{\ell,1}:=\E_p g_{\ell,1}-\E_{p_*}g_{\ell,1}\). By
\cref{lem:bounded-test-functions}, \(\norm{\delta_{\ell,1}}\lesssim R\sqrt{\eps}\). Decompose
\[
\begin{aligned}
    \delta_{\ell,1}
    &=
    \left[
        \sum_{i\in \bar{S}_\ell}\pi_i\E_{\vmu_i}g_{\ell,1}
        -
        \pi_\ell^*\E_\ell g_{\ell,1}
    \right]
    +
    \sum_{j\neq \ell}
    \left[
        \sum_{i\in \bar{S}_j}\pi_i\E_{\vmu_i}g_{\ell,1}
        -
        \pi_j^*\E_j g_{\ell,1}
    \right]
    +
    N_{\ell,1},
\end{aligned}
\]
where \(N_{\ell,1}:=\sum_{i\in \bar{S}_0}\pi_i\E_{\vmu_i}g_{\ell,1}\) is the null contribution to the
local first-moment test. The same-group contribution is
\[
\begin{aligned}
    \sum_{i\in \bar{S}_\ell}\pi_i\E_{\vmu_i}g_{\ell,1}
    -
    \pi_\ell^*\E_\ell g_{\ell,1}
    &=
    \sum_{i\in \bar{S}_\ell}\pi_i G_{\ell,1}(\vh_i^{(\ell)})
    -
    \pi_\ell^*G_{\ell,1}(0)\\
    &=
    \Gamma_{\ell,1}
    +
    G_{\ell,1}(0)\Gamma_{\ell,0}
    +
    \bigl(\nabla G_{\ell,1}(0)-\mI_d\bigr)\Gamma_{\ell,1}
    +
    \sum_{i\in \bar{S}_\ell}\pi_i\rho_{\ell,1}(\vh_i^{(\ell)})\\
    &\myeq{a}
    \Gamma_{\ell,1}
    \pm
    O(\tau)\left(
        |\Gamma_{\ell,0}|+\norm{\Gamma_{\ell,1}}+\Gamma_{\ell,2}
    \right).
\end{aligned}
\]
where (a) uses the same-group expansion of \(G_{\ell,1}\) from
\cref{lem:local-gaussian-estimates-detailed}. For \(j\neq\ell\), the far-group
contribution is small:
\[
\begin{aligned}
    \sum_{i\in \bar{S}_j}\pi_i\E_{\vmu_i}g_{\ell,1}
    -
    \pi_j^*\E_j g_{\ell,1}
    &=
    \sum_{i\in \bar{S}_j}\pi_i G_{\ell\leftarrow j,1}(\vh_i^{(j)})
    -
    \pi_j^*G_{\ell\leftarrow j,1}(0)\\
    &=
    \Gamma_{j,0}G_{\ell\leftarrow j,1}(0)
    +
    \nabla G_{\ell\leftarrow j,1}(0)\Gamma_{j,1}
    \pm
    O(\tau\Gamma_{j,2})\\
    &\myeq{a}
    \pm
    O(\tau)\left(
        |\Gamma_{j,0}|+\norm{\Gamma_{j,1}}+\Gamma_{j,2}
    \right).
\end{aligned}
\]
where (a) uses the far-group derivative bounds for \(G_{\ell\leftarrow j,1}\) from
\cref{lem:local-gaussian-estimates-detailed}. Combining the
same-group, far-group, and null contributions gives
\[
    \delta_{\ell,1}=\Gamma_{\ell,1}+N_{\ell,1}+e_{\ell,1},
    \qquad
    \sum_{\ell=1}^m\norm{e_{\ell,1}}\lesssim m\tau\mathcal E,
    \qquad
    \sum_{\ell=1}^m\norm{N_{\ell,1}}\le R\pi_0.
\]
The last inequality again uses disjointness: since \(g_{\ell,1}\) is supported on \(\Omega_\ell\) and
\(\norm{g_{\ell,1}}\le R\), we have
\(\sum_{\ell=1}^m\norm{g_{\ell,1}(\vx)}\le R\). Thus, summing over \(\ell\) and using
\(\norm{\delta_{\ell,1}}\lesssim R\sqrt{\eps}\),
\[
    \sum_{\ell=1}^m\norm{\Gamma_{\ell,1}}
    \lesssim
    mR\sqrt{\eps}
    +
    R\pi_0
    +
    m\tau\mathcal E.
\]

\paragraph{2nd-order $\Gamma_{\ell,2}$ and the null mass $\pi_0$.}
It remains to control \(\sum_\ell\Gamma_{\ell,2}\) and \(\pi_0\). The advantage of the clipped
quadratic test is that near components contribute their squared distances, while null components
contribute a positive amount of order \(B_{\rm id}^2\). Let
\(\delta_{\rm quad}:=\E_p g_{\rm quad}-\E_{p_*}g_{\rm quad}\). By
\cref{lem:bounded-test-functions}, \(|\delta_{\rm quad}|\lesssim R^2\sqrt{\eps}\). We first write
the exact decomposition
\[
\begin{aligned}
    \delta_{\rm quad}
    &=
    \sum_{\ell=1}^m
    \left[
        \sum_{i\in \bar{S}_\ell}\pi_i\E_{\vmu_i}g_{\rm quad}
        -
        \pi_\ell^*\E_\ell g_{\rm quad}
    \right]
    +
    \sum_{i\in \bar{S}_0}\pi_i\E_{\vmu_i}g_{\rm quad}.
\end{aligned}
\]
For the near group \(\bar{S}_\ell\), the expansion of \(G_{\ell,2}\) gives
\[
\begin{aligned}
    \sum_{i\in \bar{S}_\ell}\pi_i\E_{\vmu_i}g_{\rm quad}
    -
    \pi_\ell^*\E_\ell g_{\rm quad}
    &=
    \sum_{i\in \bar{S}_\ell}\pi_i G_{\ell,2}(\vh_i^{(\ell)})
    -
    \pi_\ell^*G_{\ell,2}(0)\\
    &=
    \Gamma_{\ell,2}
    +
    G_{\ell,2}(0)\Gamma_{\ell,0}
    +
    \langle \nabla G_{\ell,2}(0),\Gamma_{\ell,1}\rangle
    +
    \sum_{i\in \bar{S}_\ell}\pi_i\rho_{\ell,2}(\vh_i^{(\ell)})\\
    &\myeq{a}
    \Gamma_{\ell,2}
    \pm
    O(\tau)\left(
        |\Gamma_{\ell,0}|+\norm{\Gamma_{\ell,1}}+\Gamma_{\ell,2}
    \right).
\end{aligned}
\]
where (a) uses the same-group expansion of \(G_{\ell,2}\) from
\cref{lem:local-gaussian-estimates-detailed}. 

For every \(i\in \bar{S}_0\), the null lower bound in
\cref{lem:local-gaussian-estimates-detailed} gives
\(\E_{\vmu_i}g_{\rm quad}\gtrsim B_{\rm id}^2\). Hence
\[
    \sum_{i\in \bar{S}_0}\pi_i\E_{\vmu_i}g_{\rm quad}
    \gtrsim
    B_{\rm id}^2\pi_0.
\]
Combining the above bounds gives
\[
    R^2\sqrt{\eps}
    \gtrsim
    \abs{\delta_{\rm quad}}
    \ge
    \delta_{\rm quad}
    \gtrsim
    \sum_{\ell=1}^m\Gamma_{\ell,2}
    +
    B_{\rm id}^2\pi_0
    -
    \tau\mathcal E.
\]
In particular,
\[
    \pi_0
    \lesssim
    \frac{R^2}{B_{\rm id}^2}\sqrt{\eps}
    +
    \frac{\tau}{B_{\rm id}^2}\mathcal E,
    \qquad
    \sum_{\ell=1}^m\Gamma_{\ell,2}
    \lesssim
    R^2\sqrt{\eps}
    +
    \tau\mathcal E.
\]

\paragraph{Closing the inequalities.}
Combining the estimates for \(\Gamma_{\ell,0}\), \(\Gamma_{\ell,1}\), \(\Gamma_{\ell,2}\), and
\(\pi_0\), we get
\[
\begin{aligned}
    \mathcal E
    &=
    \pi_0+
    \sum_{\ell=1}^m
    \left(
        |\Gamma_{\ell,0}|+\norm{\Gamma_{\ell,1}}+\Gamma_{\ell,2}
    \right)\\
    &\lesssim
    m(1+R)\sqrt{\eps}
    +
    (1+R)\pi_0
    +
    R^2\sqrt{\eps}
    +
    m\tau\mathcal E\\
    &\myle{a}
    m(1+R)^3\sqrt{\eps}
    +
    m(1+R)\tau\mathcal E.
\end{aligned}
\]
where (a) substitutes
\(\pi_0\lesssim (R^2/B_{\rm id}^2)\sqrt{\eps}+(\tau/B_{\rm id}^2)\mathcal E\), uses \(B_{\rm id}\ge 1\), and absorbs
\((1+R)\tau/B_{\rm id}^2\) into \(m(1+R)\tau\). By the choice of
\(b_0,r_0,C_{\rm sep}\), we have \(m(1+R)\tau\le c\) for a sufficiently small universal constant.
Absorbing the last term into the left-hand side gives
\[
    \mathcal E
    \lesssim
    m(1+R)^3\sqrt{\eps}.
\]
This proves the theorem.
\end{proof}

\subsection{Omitted proofs}\label{sec: id omit proof}
We now prove \cref{lem:bounded-test-functions} and \cref{lem:local-gaussian-estimates-detailed}.

\lemboundedtestfun*
\begin{proof}
By Pinsker's inequality,
\[
    \frac12\int_{\R^d}|p(\vx)-p_*(\vx)|\rd\vx
    =
    {\rm TV}(p_*,p)
    \le
    \sqrt{\frac{\kl(p_*\|p)}{2}}
    =
    \sqrt{\frac{\eps}{2}}.
\]
For scalar-valued \(f\),
\[
\begin{aligned}
    \left|\E_p f-\E_{p_*}f\right|
    &=
    \left|
        \int_{\R^d} f(\vx)(p(\vx)-p_*(\vx))\rd\vx
    \right|
    \le
    2\norm{f}_\infty{\rm TV}(p_*,p)
    \le
    \sqrt{2}\,\norm{f}_\infty\sqrt{\eps}.
\end{aligned}
\]
For vector-valued \(f\), the claim follows by applying the scalar bound to
\(f_{\vu}(\vx):=\langle \vu,f(\vx)\rangle\) and taking the supremum over all unit vectors \(\vu\).
\end{proof}

\lemlocalgaussianestimate*
\begin{proof}
The proof has four parts. First, we prove a rare-event moment bound for a Gaussian whose mean is
within distance \(B_{\rm id}\) of \(\vmu_\ell^*\). This bound is then used for the same-group Taylor
expansions of the maps \(G_{\ell,0},G_{\ell,1},G_{\ell,2}\). Then we prove the far-group estimates,
where the Gaussian is centered near a different true mean. Finally, we prove the lower bound for
the clipped statistic \(g_{\rm quad}\) for null components.

\paragraph{Step 1: the main rare-event estimate.}
The goal of this step is to prove that, uniformly over \(\norm{\vh}\le B_{\rm id}\),
\[
    \E\left[
        (1+\norm{\vh+\vZ})^8
        \ind\{\vmu_\ell^*+\vh+\vZ\notin\Omega_\ell\}
    \right]
    \le
    C(1+R)^8
    \left[
        e^{-c(R-B_{\rm id})^2}
        +
        \ind_{\{m\ge2\}}\,
        m e^{-c(\Delta-2B_{\rm id})^2}
    \right].
\]
This estimate will be used to show that localized Gaussian moments differ from the corresponding
full Gaussian moments only by \(O(\tau)\).

Since \(\Omega_\ell=V_\ell\cap B(\vmu_\ell^*,R)\), we have
\[
\begin{aligned}
    \P(\vmu_\ell^*+\vh+\vZ\notin\Omega_\ell)
    &\le
    \P(\norm{\vh+\vZ}>R)
    +
    \P(\vmu_\ell^*+\vh+\vZ\notin V_\ell)\\
    &\le
    e^{-c(R-B_{\rm id})^2}
    +
    \ind_{\{m\ge2\}}\,
    m e^{-c(\Delta-2B_{\rm id})^2}.
\end{aligned}
\]
For the first term, \(\norm{\vh+\vZ}>R\) implies
\(\norm{\vZ}>R-B_{\rm id}\). By choosing \(r_0\) large enough, \(R-B_{\rm id}\gtrsim\sqrt{d+B_{\rm id}^2}\), and the standard
\(\chi_d^2\) tail bound gives
\[
    \P(\norm{\vh+\vZ}>R)
    \le
    \P(\norm{\vZ}>R-B_{\rm id})
    \le
    e^{-c(R-B_{\rm id})^2}.
\]
For the second term, there is nothing to prove when \(m=1\).  When
\(m\ge2\), if
\(\vmu_\ell^*+\vh+\vZ\notin V_\ell\), then for some \(j\neq \ell\), this point is closer to
\(\vmu_j^*\) than to \(\vmu_\ell^*\). Thus, with
$\vv_{\ell j}:=(\vmu_j^*-\vmu_\ell^*)/\norm{\vmu_j^*-\vmu_\ell^*}$,
\[
    \langle \vh+\vZ,\vv_{\ell j}\rangle
    \ge
    \frac12\norm{\vmu_j^*-\vmu_\ell^*}
    \quad\Rightarrow\quad
    \langle \vZ,\vv_{\ell j}\rangle
    \ge
    \frac{\Delta}{2}-B_{\rm id},
\]
and the one-dimensional Gaussian tail bound plus a union bound over \(j\neq\ell\) gives the bound.

Let $\gA:=\{\vmu_\ell^*+\vh+\vZ\notin\Omega_\ell\}$. By Cauchy--Schwarz,
\[
\begin{aligned}
    \E\left[
        (1+\norm{\vh+\vZ})^8\ind_{\gA}
    \right]
    &\le
    \left(\E(1+\norm{\vh+\vZ})^{16}\right)^{1/2}
    \P(\gA)^{1/2}\\
    &\le
    C(1+B_{\rm id}+\sqrt d)^8
    \P(\gA)^{1/2}\\
    &\le
    C(1+R)^8
    \left[
        e^{-c(R-B_{\rm id})^2}
        +
        \ind_{\{m\ge2\}}\,
        m e^{-c(\Delta-2B_{\rm id})^2}
    \right].
\end{aligned}
\]
Here the last line uses \(R=r_0\sqrt{d+B_{\rm id}^2}\) and absorbs square roots of tail probabilities into
the constants \(C,c\). This proves the rare-event moment bound.

\paragraph{Step 2: same-group estimates.}
We now use the rare-event bound to prove the expansions for \(G_{\ell,0},G_{\ell,1},G_{\ell,2}\).
The idea is that, on \(\Omega_\ell\), the localized tests agree with the corresponding full Gaussian
moments, and after subtracting the full moments the error is supported on the rare region. Let
$\mathcal R_\ell:=\Omega_\ell-\vmu_\ell^*=\{\vmu-\vmu_\ell^*:\vmu\in\Omega_\ell\}$.

We first present the consequence of Step 1 that will be used throughout this step. Uniformly over
\(\norm{\vh}\le B_{\rm id}\),
\[
    \E\left[
        (1+R+\norm{\vh+\vZ})^8
        \ind\{\vh+\vZ\in\mathcal R_\ell^c\}
    \right]
    \lesssim
    \tau.
\]
Indeed, the event \(\{\vh+\vZ\in\mathcal R_\ell^c\}\) is contained in the union of
\(\{\norm{\vh+\vZ}>R\}\) and
\(\{\vmu_\ell^*+\vh+\vZ\notin V_\ell,\ \norm{\vh+\vZ}\le R\}\). On the first event,
\(1+R+\norm{\vh+\vZ}\lesssim 1+\norm{\vh+\vZ}\), so Step 1 gives the required contribution. On the
second event, the integrand is at most \(C(1+R)^8\), while the probability is at most
\(m e^{-c(\Delta-2B_{\rm id})^2}\). Combining these two bounds and using the definition of \(\tau\) proves
the display. Together with \cref{lem:shifted-gaussian-derivatives}, this estimate controls the
value and the first two derivatives of all rare-region Gaussian averages below.

For \(G_{\ell,0}\), since the full Gaussian mass equals one,
\[
    G_{\ell,0}(\vh)-1
    =
    -\int_{\mathcal R_\ell^c}\phi(0;\vy-\vh)\rd\vy.
\]
Thus the error in replacing the localized mass by the full Gaussian mass is supported only on
\(\mathcal R_\ell^c\). Applying \cref{lem:shifted-gaussian-derivatives} with \(q\equiv 1\) gives
\[
    |G_{\ell,0}(0)-1|
    +
    \norm{\nabla G_{\ell,0}(0)}
    +
    \sup_{\norm{\vh}\le B_{\rm id}}
    \norm{\nabla^2G_{\ell,0}(\vh)}_{\rm op}
    \le
    \tau.
\]
Taylor's theorem gives the expansion for \(G_{\ell,0}\).

For \(G_{\ell,1}\), the corresponding full-space identity is
$\int_{\R^d}\vy\,\phi(0;\vy-\vh)\rd\vy=\vh$.
Since \(g_{\ell,1}(\vmu_\ell^*+\vy)=\vy\) on \(\mathcal R_\ell\), the difference
\(G_{\ell,1}(\vh)-\vh\) is supported on the same rare region:
\[
    G_{\ell,1}(\vh)-\vh
    =
    \int_{\mathcal R_\ell^c}
    \bigl(g_{\ell,1}(\vmu_\ell^*+\vy)-\vy\bigr)
    \phi(0;\vy-\vh)\rd\vy .
\]
On \(\mathcal R_\ell^c\), the integrand has norm at most \(\norm{\vy}\). Thus
\cref{lem:shifted-gaussian-derivatives} and the rare-event moment bound give
\[
    \norm{G_{\ell,1}(0)}
    +
    \norm{\nabla G_{\ell,1}(0)-\mI_d}_{\rm op}
    +
    \sup_{\norm{\vh}\le B_{\rm id}}
    \norm{\nabla^2G_{\ell,1}(\vh)}_{\rm op}
    \le
    \tau.
\]
Taylor's theorem gives the expansion for \(G_{\ell,1}\).

It remains to treat \(G_{\ell,2}\). On \(\mathcal R_\ell\), we have
\(g_{\rm quad}(\vmu_\ell^*+\vy)=\norm{\vy}^2-d\), while the full Gaussian quadratic moment satisfies
\(\int_{\R^d}(\norm{\vy}^2-d)\phi(0;\vy-\vh)\rd\vy=\norm{\vh}^2\). Hence
\[
    G_{\ell,2}(\vh)-\norm{\vh}^2
    =
    \int_{\mathcal R_\ell^c}
    \bigl(g_{\rm quad}(\vmu_\ell^*+\vy)-(\norm{\vy}^2-d)\bigr)
    \phi(0;\vy-\vh)\rd\vy.
\]
The error is again supported only on \(\mathcal R_\ell^c\), and
$\left|
        g_{\rm quad}(\vmu_\ell^*+\vy)-(\norm{\vy}^2-d)
    \right|
    \lesssim
    R^2+\norm{\vy}^2+d$.
Using \cref{lem:shifted-gaussian-derivatives} and the rare-event moment bound,
\[
    |G_{\ell,2}(0)|
    +
    \norm{\nabla G_{\ell,2}(0)}
    +
    \sup_{\norm{\vh}\le B_{\rm id}}
    \norm{\nabla^2G_{\ell,2}(\vh)-2\mI_d}_{\rm op}
    \le
    \tau.
\]
Taylor's theorem gives the expansion for \(G_{\ell,2}\).

\paragraph{Step 3: far-group estimates.}
When \(m=1\), there are no far-group estimates.  We therefore assume
\(m\ge2\) in this step and consider a Gaussian centered within distance
\(B_{\rm id}\) of a different true center \(\vmu_j^*\), where \(j\neq\ell\). The goal is to show that its contribution to the local tests
associated with \(\Omega_\ell\) is negligible, including after differentiating the shifted Gaussian
density.

Fix \(\norm{\vh}\le B_{\rm id}\). If \(\vmu_j^*+\vh+\vZ\in\Omega_\ell\subset V_\ell\), then the point must cross
the Voronoi hyperplane between \(\vmu_j^*\) and \(\vmu_\ell^*\). The same one-dimensional margin
calculation as in Step 1 gives
\[
    \P(\vmu_j^*+\vh+\vZ\in\Omega_\ell)
    \le
    e^{-c(\Delta-2B_{\rm id})^2}.
\]
Moreover, the corresponding fixed-degree moment bound also holds, after decreasing \(c\):
\[
    \E\left[
        (1+R+\norm{\vZ})^8
        \ind\{\vmu_j^*+\vh+\vZ\in\Omega_\ell\}
    \right]
    \le
    C(1+R)^8e^{-c(\Delta-2B_{\rm id})^2}.
\]
Since \(g_{\ell,0}\) and \(g_{\ell,1}\) are supported on \(\Omega_\ell\), and since
\(\norm{g_{\ell,0}}_\infty\le 1\), \(\norm{g_{\ell,1}}_\infty\le R\), applying
\cref{lem:shifted-gaussian-derivatives} to the far-group response maps gives
\[
\begin{aligned}
    |G_{\ell\leftarrow j,0}(0)|
    +
    \norm{\nabla G_{\ell\leftarrow j,0}(0)}
    +
    \sup_{\norm{\vt}\le B_{\rm id}}
    \norm{\nabla^2G_{\ell\leftarrow j,0}(\vt)}_{\rm op}
    &\le
    \tau,\\
    \norm{G_{\ell\leftarrow j,1}(0)}
    +
    \norm{\nabla G_{\ell\leftarrow j,1}(0)}_{\rm op}
    +
    \sup_{\norm{\vt}\le B_{\rm id}}
    \norm{\nabla^2G_{\ell\leftarrow j,1}(\vt)}_{\rm op}
    &\le
    \tau.
\end{aligned}
\]
This proves the far-group estimates.

\paragraph{Step 4: null group lower bound.}
It remains to prove that a Gaussian whose center is farther than \(B_{\rm id}\) from all true centers has
positive lower bound. Let
\(\delta:=\min_{i\in[m]}\norm{\vmu-\vmu_i^*}>B_{\rm id}\) and
\(\vh_s:=\vmu-\vmu_s^*\). For \(\vX=\vmu+\vZ\),
\[
    D(\vX)^2
    =
    \min_s\norm{\vZ+\vh_s}^2
    =
    \norm{\vZ}^2+
    \min_s\left(\norm{\vh_s}^2+2\langle \vh_s,\vZ\rangle\right).
\]
Define
\[
    \mathcal G
    :=
    \bigcap_{s=1}^m
    \left\{
        2\langle \vh_s,\vZ\rangle\ge -\frac12\norm{\vh_s}^2
    \right\}.
\]
Since \(\norm{\vh_s}\ge\delta>B_{\rm id}\), the one-dimensional Gaussian tail bound and a union bound give
\(\P(\mathcal G^c)\le m e^{-cB_{\rm id}^2}\). On \(\mathcal G\),
\(D(\vX)^2\ge\norm{\vZ}^2+\delta^2/2\).

We split according to whether the null center is within the clipping radius. First suppose
\(\delta\le R\). On \(\mathcal G\), we have
\(D(\vX)^2\ge \norm{\vZ}^2+\delta^2/2\). Therefore,
\[
\begin{aligned}
    \E\min\{D(\vX)^2,R^2\}
    &\ge
    \E\min\left\{\norm{\vZ}^2+\frac12\delta^2,R^2\right\}
    -
    R^2\P(\mathcal G^c)\\
    &\ge
    d+\frac12\delta^2
    -
    \E\left[
        \left(
            \norm{\vZ}^2-\frac12R^2
        \right)_+
    \right]
    -
    R^2m e^{-cB_{\rm id}^2}\\
    &\ge
    d+cB_{\rm id}^2 .
\end{aligned}
\]
The last line follows from the
standard \(\chi^2_d\) tail bound, choosing \(r_0\) large enough so that the truncation error is a
sufficiently small multiple of \(B_{\rm id}^2\), and then choosing \(b_0\) large enough so that
\(R^2m e^{-cB_{\rm id}^2}\) is also a sufficiently small multiple of \(B_{\rm id}^2\). Since \(\delta>B_{\rm id}\), the positive
term \(\delta^2/2\) leaves a contribution of order \(B_{\rm id}^2\).

It remains to consider \(\delta>R\). On \(\mathcal G\), we have
\(D(\vX)^2\ge \norm{\vZ}^2+\delta^2/2\ge R^2/2\), and hence
\[
\begin{aligned}
    \E\min\{D(\vX)^2,R^2\}
    &\ge
    \E\left[
        \min\{D(\vX)^2,R^2\}\ind_{\mathcal G}
    \right]\\
    &\ge
    \frac12R^2\P(\mathcal G)
    \ge
    \frac12R^2-R^2m e^{-cB_{\rm id}^2}
    \ge
    d+cB_{\rm id}^2 .
\end{aligned}
\]
The last inequality follows by choosing \(r_0\) large enough and then \(b_0\) large enough.
Subtracting \(d\) in the two cases proves \(\E_\vmu g_{\rm quad}\gtrsim B_{\rm id}^2\).
\end{proof}

The following lemma justifies the differentiation of shifted Gaussian averages used in \cref{lem:local-gaussian-estimates-detailed} above. The test function \(q\) need not be smooth, the required smoothness comes from convolution with the Gaussian density.
\begin{lemma}
\label{lem:shifted-gaussian-derivatives}
Let \(\mathcal A\subset\R^d\) be measurable and let \(q:\R^d\to\R^r\) be measurable with
polynomial growth. Define
\[
    F(\vh):=\int_{\mathcal A} q(\vy)\phi(0;\vy-\vh)\rd\vy .
\]
Then \(F\) is smooth, and differentiation can be passed to the shifted
Gaussian density.  For vector-valued \(F\), the following identities are
understood componentwise, with \(\nabla F\) denoting the Jacobian:
\begin{align*}
    \nabla F(\vh)
    &=
    \int_{\mathcal A}q(\vy)(\vy-\vh)^\top\phi(0;\vy-\vh)\rd\vy,\\
    \nabla^2F(\vh)
    &=
    \int_{\mathcal A}q(\vy)
    \left((\vy-\vh)(\vy-\vh)^\top-\mI_d\right)
    \phi(0;\vy-\vh)\rd\vy.
\end{align*}
Consequently, with the natural operator norms for the Jacobian and the
componentwise Hessian,
\[
    \norm{F(\vh)}
    +
    \norm{\nabla F(\vh)}_{\rm op}
    +
    \norm{\nabla^2F(\vh)}_{\rm op}
    \lesssim
    \E\left[
        \norm{q(\vh+\vZ)}(1+\norm{\vZ}^2)
        \ind\{\vh+\vZ\in\mathcal A\}
    \right].
\]
\end{lemma}
\begin{proof}
The key point is that the smoothing comes from the Gaussian density, not from \(q\). Since \(q\)
has polynomial growth, for every fixed \(\vh_0\) there is a neighborhood of \(\vh_0\) on which
\[
    \norm{q(\vy)}
    \left(
        1+\norm{\vy-\vh}^2
    \right)
    \phi(0;\vy-\vh)
\]
is dominated by an integrable function of \(\vy\). Hence dominated convergence allows us to
differentiate under the integral sign. The derivative identities follow from
\[
    \nabla_\vh\phi(0;\vy-\vh)
    =
    (\vy-\vh)\phi(0;\vy-\vh),
\]
and
\[
    \nabla_\vh^2\phi(0;\vy-\vh)
    =
    \left((\vy-\vh)(\vy-\vh)^\top-\mI_d\right)\phi(0;\vy-\vh).
\]
The final bound follows by taking norms in these identities and then changing variables
\(\vy=\vh+\vZ\).
\end{proof}

\section{Local convergence}
\label{appendix: local}

The goal of this section is to prove the local phase under a bounded
trajectory potential.

\paragraph{Notations.}
Recall
\[
    \Delta_{\max}:=
    \max_{i,j\in[m]}\norm{\vmu_i^*-\vmu_j^*},
    \quad
    \Delta:=\min_{i\ne j\in[m]}\norm{\vmu_i^*-\vmu_j^*}.
\]
All assumptions and estimates involving separated true centers are imposed only
when \(m\ge2\); when \(m=1\), the corresponding cross-cluster terms are absent.

We use the same full Voronoi partition as in \cref{appendix:identifiability-new} that
\[
    S_\ell
    :=
    \left\{
        i\in[n]:
        \norm{\vmu_i-\vmu_\ell^*}
        \le
        \norm{\vmu_i-\vmu_j^*}
        \text{ for every }j\in[m]
    \right\}.
\]
For \(\delta>0\), define the \(\delta\)-close subgroup and its total weight by
\[
    S_\ell(\delta):=
    \{i\in S_\ell:\norm{\vmu_i-\vmu_\ell^*}\le\delta\},
    \qquad
    \tldpi_\ell(\delta):=\sum_{i\in S_\ell(\delta)}\pi_i .
\]
When \(\delta\) is clear from context, we write \(\tldpi_\ell\) for
\(\tldpi_\ell(\delta)\).  Finally, define the trajectory potential
\[
    U(\vmu):=
    \sum_{\ell=1}^m\sum_{i\in S_\ell}
    \norm{\vmu_i-\vmu_\ell^*}^2 .
\]
All these quantities are evaluated at the current value of \(\vmu\); along the
trajectory, we suppress this time dependence.

We now state the main result of this section.  It shows that once the loss is
below a sufficiently small threshold \(\eps_0\), the local phase decreases the
loss at the \(1/t^2\) rate.  The initial conditions below are guaranteed by
\cref{thm: global main} with \(\eps_0=\exp(-\Theta(\Delta^2))\) and
\(B\asymp\sqrt{dn}\) when \(m\ge2\).  Together with \cref{assump: delta}, the
separation and small-overlap conditions below are satisfied.

\begin{theorem}[\cref{thm: local main text} restated, main result for local phase]
\label{thm: local main}
    Let \(\Xi_{\rm id}\) be the quantity from
    \cref{thm:local-identifiability-log-separation}.  There exist
    universal constants \(c,C,C_{\rm sep},c_{\rm kkt}>0\) and polynomial
    quantities
    \[
        \sfA_{\rm loc}:=
        \poly\left(
            d,m,n,\frac1{\pimins},
            1+B
        \right),
        \qquad
        \sfA_{\eta}:=
        C\left(
            1+d+B^2+\log\frac1{\pimins}
            +
            \frac1{\pimins}
        \right)
    \]
    such that the following holds.  If \(m\ge2\), suppose
    \[
        \Delta\ge C_{\rm sep}
        \left(
            B+\sqrt{\log\sfA_{\rm loc}}
        \right).
    \]
    When \(m=1\), this separation condition is omitted.

    Suppose that at time \(T_1\),
        $\L(\vpi^{(T_1)},\vmu^{(T_1)})
        \le
        \eps_0$ and 
        $U(\vmu^{(T_1)})\le B^2/2$, where the local threshold \(\eps_0\) is small enough so that
    \[
        \Xi_{\rm id}\sqrt{\eps_0}\le c\pimins,
        \qquad
        (1+B)
        \sqrt{\frac{\Xi_{\rm id}}{\pimins}}\eps_0^{1/4}\le c,
        \qquad
        \sfA_{\rm loc}^{C}\eps_0^c\le c,
        \qquad
        \sqrt{A_{\rm rate}}\eps_0^{1/4}\le cB.
    \]
    Assume that after each weight step with \(\vmu=\vmu^{(t)}\) and
    \(\vpi=\vpi^{(t+1)}\), until the target loss \(L\le\eps\), the weight
    step is loss non-increasing and satisfies dual-gap condition as in \cref{appendix: alg} that
    \[
        G_\pi(\vpi,\vmu)
        \le
        c_{\rm kkt}\L(\vpi,\vmu).
    \]
    If the step size satisfies \(\eta\le c/\sfA_{\eta}\), then for
    every \(t\ge T_1\) until loss \(\L\le\eps\),
    \[
        \L(\vpi^{(t)},\vmu^{(t)})
        \le
        \frac{A_{\rm rate}^2}
        {(\eta(t-T_1))^2+A_{\rm rate}^2/\eps_0},
        \qquad
        U(\vmu^{(t)})\le B^2,
    \]
    where \(A_{\rm rate}\asymp n\Xi_{\rm id}/\pimins\).
\end{theorem}

We prove the theorem by estimating the evolution of the two quantities
\[
    L_t:=\L(\vpi^{(t)},\vmu^{(t)}),
    \qquad
    U_t:=U(\vmu^{(t)}).
\]
The argument relies on two one-step estimates.  First, the loss decreases at order \(L_t^{3/2}\), because the gradient has a positive projection onto the
close-component direction. Second, the mean update on $\vmu$ does not increase the potential \(U\) too much, because the total sum of gradient can be bounded. The proof of the one-step estimates is deferred
to \cref{sec: local descent}.

\begin{proposition}
\label{prop:local-one-step-bootstrap}
    Under the setting of \cref{thm: local main},
    consider a point \((\vpi,\vmu)\) satisfying
    $\eps:=\L(\vpi,\vmu)\le\eps_0$ and $U(\vmu)\le B^2$,
    and suppose that the dual-gap condition in \cref{thm: local main} holds at \((\vpi,\vmu)\).  Let
    $\vmu^+:=\vmu-\eta\nabla_{\vmu}\L(\vpi,\vmu)$.
    If \(\eta\le c/\sfA_{\eta}\), then
    \begin{align}
        \L(\vpi,\vmu^+)
        &\le
        \L(\vpi,\vmu)
        -
        \frac{\eta}{A_{\rm rate}}\L(\vpi,\vmu)^{3/2},
        \label{eq:one-step-loss-decrease}\\
        \eta\norm{\nabla_{\vmu}\L(\vpi,\vmu)}_F
        &\le
        C\sqrt{A_{\rm rate}}
        \frac{
            \L(\vpi,\vmu)-\L(\vpi,\vmu^+)
        }{
            \L(\vpi,\vmu)^{3/4}
        },
        \label{eq:one-step-grad-norm}
    \end{align}
    where \(A_{\rm rate}\asymp n\Xi_{\rm id}/\pimins\).
\end{proposition}

We now show that this one-step statement implies the theorem.
\begin{proof}[Proof of \cref{thm: local main}]
    Let $L_t:=\L(\vpi^{(t)},\vmu^{(t)})$, and $U_t:=U(\vmu^{(t)})$.
    We argue until the stopping time
    $\tau_\eps:=\inf\{t\ge T_1:L_t\le\eps\}$.
    Suppose \(t<\tau_\eps\), \(L_t\le\eps_0\), and
    \(U_t\le B^2\).  Let
    \[
        L_{t+0.5}:=
        \L(\vpi^{(t+1)},\vmu^{(t)})
    \]
    be the loss after the weight step.  The weight step is loss
    non-increasing and does not move the means, so
    \(L_{t+0.5}\le L_t\) and \(U\) is unchanged.  Applying
    \eqref{eq:one-step-loss-decrease} in \cref{prop:local-one-step-bootstrap} at
    \((\vpi^{(t+1)},\vmu^{(t)})\), we get
    \[
        L_{t+1}
        \le
        L_{t+0.5}
        -
        \frac{\eta}{A_{\rm rate}}L_{t+0.5}^{3/2}.
    \]
    We next convert this recurrence into one involving \(L_t\).  If
    \(L_{t+0.5}\ge L_t/2\), then
    $L_{t+1}
        \le
        L_t
        -
        \frac{c\eta}{A_{\rm rate}}L_t^{3/2}$,
    after absorbing a universal constant into \(c\).  If
    \(L_{t+0.5}<L_t/2\), then \(L_{t+1}\le L_{t+0.5}\), and hence $L_t-L_{t+1}
        \ge
        L_t-L_{t+0.5}
        \ge
        \frac12L_t
        \ge
        \frac{c\eta}{A_{\rm rate}}L_t^{3/2}$,
    where the last inequality uses \(L_t\le\eps_0\le1\), \(\eta\le
    c/\sfA_{\eta}\le1\), and \(A_{\rm rate}\ge1\), after decreasing
    \(c\) and enlarging \(A_{\rm rate}\) by a universal constant if necessary.
    Therefore, in all cases,
    \[
        L_{t+1}
        \le
        L_t-
        \frac{c\eta}{A_{\rm rate}}L_t^{3/2}.
    \]
    In particular, \(L_{t+1}\le L_t\le\eps_0\), so the induction on the loss holds. 

    It remains to prove that the trajectory stays inside the larger potential
    ball.  Let
    \[
        \vg_t:=
        \nabla_{\vmu}\L(\vpi^{(t+1)},\vmu^{(t)}).
    \]
    Since \(U^{1/2}\) is the distance to the union of true-center assignments,
    the triangle inequality with the old assignment and \eqref{eq:one-step-grad-norm} in \cref{prop:local-one-step-bootstrap} gives
    \[
        \sqrt{U_{t+1}}
        \le
        \sqrt{U_t}+\eta\norm{\vg_t}_F
        \le
        \sqrt{U_t}
        + C\sqrt{A_{\rm rate}}
        \frac{L_{t+0.5}-L_{t+1}}{L_{t+0.5}^{3/4}}.
    \]
    Because the weight step is loss non-increasing and \(L_{t+1}\le
    L_{t+0.5}\le L_t\), summing over time gives
    \[
    \begin{aligned}
        \sqrt{U_{t+1}}
        - \sqrt{U_{T_1}}
        \le 
        \sum_{s=T_1}^{t}
        \eta\norm{\vg_s}_F
        &\le
        C\sqrt{A_{\rm rate}}
        \sum_{s=T_1}^{t}
        \frac{L_{s+0.5}-L_{s+1}}{L_{s+0.5}^{3/4}}         \le
        C\sqrt{A_{\rm rate}}
        \sum_{s=T_1}^{t}\int_{L_{s+1}}^{L_{s+0.5}} u^{-3/4}\rd u\\
        &\le
        C\sqrt{A_{\rm rate}}
        \int_0^{L_{T_1}}u^{-3/4}\rd u
        \le
        C\sqrt{A_{\rm rate}}\eps_0^{1/4}.
    \end{aligned}
    \]
    Combining \(U_{T_1}\le B^2/2\), we have
    $\sqrt{U_{t+1}}
        \le
        B/\sqrt{2}
        +
        C\sqrt{A_{\rm rate}}\eps_0^{1/4}
        \le
        B$.  This closes the induction.

    Finally, we solve this scalar recurrence of loss.  If \(L_{t+1}=0\), the desired
    bound is immediate.  Otherwise, since \(L_{t+1}\le L_t\),
    \[
    \begin{aligned}
        \frac1{\sqrt{L_{t+1}}}-\frac1{\sqrt{L_t}}
        &=
        \frac{L_t-L_{t+1}}
        {\sqrt{L_tL_{t+1}}(\sqrt{L_t}+\sqrt{L_{t+1}})}
        \ge
        \frac{L_t-L_{t+1}}{2L_t^{3/2}}
        \ge
        \frac{c\eta}{A_{\rm rate}} .
    \end{aligned}
    \]
    Summing from \(T_1\) to \(t-1\), and using
    \(L_{T_1}\le\eps_0\), gives
    \[
        \frac1{\sqrt{L_t}}
        \ge
        \frac1{\sqrt{\eps_0}}
        +
        \frac{c\eta(t-T_1)}{A_{\rm rate}}
        \qquad\Rightarrow\qquad
        L_t
        \le
        \frac{A_{\rm rate}^2}
        {(\eta(t-T_1))^2+A_{\rm rate}^2/\eps_0},
    \]
    where we rearrange and absorb universal constants into \(A_{\rm rate}\).
\end{proof}

\subsection{Descent estimates: Proof of \cref{prop:local-one-step-bootstrap}}\label{sec: local descent}

We now prove the one-step estimates needed for
\cref{prop:local-one-step-bootstrap}.  The close-component descent direction
gives a lower bound on the gradient norm, which implies both the loss decrease
and the step-length bound used to control the trajectory potential.

At the current point, the assumptions of
\cref{prop:local-one-step-bootstrap} imply \(\eps\le\eps_0\) and \(U(\vmu)\le B^2\).  Hence every component in \(S_\ell\) lies within distance \(B\) of \(\vmu_\ell^*\), and the smallness condition on \(\eps_0\) allows us to apply \cref{cor:local-identifiability-consequences}.  Thus we will use
\begin{align}
    \sum_{\ell=1}^m
    \left|
        \sum_{i\in S_\ell}\pi_i-\pi_\ell^*
    \right|
    &\lesssim
    \Xi_{\rm id}\sqrt\eps,
    \label{eq:local-use-weight}\\
    \sum_{\ell=1}^m
    \norm{
        \sum_{i\in S_\ell}
        \pi_i(\vmu_i-\vmu_\ell^*)
    }
    &\lesssim
    \Xi_{\rm id}(1+B^2)\sqrt\eps,
    \label{eq:local-use-first}\\
    \sum_{\ell=1}^m
    \sum_{i\in S_\ell}
    \pi_i\norm{\vmu_i-\vmu_\ell^*}^2
    &\lesssim
    \Xi_{\rm id}(1+B^2)\sqrt\eps,
    \label{eq:local-use-second}\\
    \tldpi_\ell:=
    \sum_{i\in S_\ell(\deltacls)}\pi_i
    &\ge
    \frac12\pi_\ell^*
    \quad\text{with}\quad\deltacls:=
    2\sqrt{\frac{\Xi_{\rm id}}{\pimins}}\eps^{1/4}
    \qquad\text{for every }\ell\in[m].
    \label{eq:local-use-close-weight}
\end{align}

To express the gradient projections, define the second-order remainder
\(\tau_i\) for \(i\in S_\ell\) by
\begin{equation}
    \phi(\vmu_\ell^*;\vx)
    =
    \phi(\vmu_i;\vx)
    \left(
        1-\left\langle \vx-\vmu_i,
        \vmu_i-\vmu_\ell^*\right\rangle
        +\tau_i(\vx)
    \right).
    \label{eq:def-tau-local}
\end{equation}
Equivalently,
\[
    \left\langle
        \nabla_{\vmu_i}\phi(\vmu_i;\vx),
        \vmu_i-\vmu_\ell^*
    \right\rangle
    =
    \phi(\vmu_i;\vx)-\phi(\vmu_\ell^*;\vx)
    +
    \phi(\vmu_i;\vx)\tau_i(\vx).
\]
This identity produces the positive main term \(p-p^*\).  The resulting
second-order terms in the close-component projection are controlled by
\cref{lem:local-linearization-remainders}.

The following estimate gives a positive projection on the close-component direction, which will imply the gradient-norm lower bound.
\begin{lemma}[\cref{lem: local descent dir main text} restated, close-component descent direction]
\label{lem: local descent dir}
    Under the assumptions of \cref{prop:local-one-step-bootstrap}, set
    $\tldpi_\ell:=\sum_{i\in S_\ell(\deltacls)}\pi_i$ with $\deltacls:=
    2\sqrt{\frac{\Xi_{\rm id}}{\pimins}}\eps^{1/4}$.
    Then
    \[
        \sum_{\ell=1}^m
        \sum_{i\in S_\ell(\deltacls)}
        \frac{\pi_\ell^*}{\tldpi_\ell}
        \left\langle
            \nabla_{\vmu_i}\L,\vmu_i-\vmu_\ell^*
        \right\rangle
        \ge \frac12\L(\vpi,\vmu).
    \]
\end{lemma}

\begin{proof}
    Define the close-component projection
    \[
        \mathcal D:=
        \sum_{\ell=1}^m
        \sum_{i\in S_\ell(\deltacls)}
        \frac{\pi_\ell^*}{\tldpi_\ell}
        \left\langle
            \nabla_{\vmu_i}\L,\vmu_i-\vmu_\ell^*
        \right\rangle 
        = \sum_{\ell=1}^m
        \sum_{i\in S_\ell(\deltacls)}
        \frac{\pi_\ell^*}{\tldpi_\ell}\pi_i
        \E_{p^*}
        \left[
            \ell'(p)
            \left\langle
                \nabla_{\vmu_i}\phi(\vmu_i;\vx),
                \vmu_i-\vmu_\ell^*
            \right\rangle
        \right].
    \]
    The goal is to recover the convexity
    term \(\E_{p^*}[\ell'(p)(p-p^*)]\).  Since \(\mathcal D\) only uses the
    close components, we add a KKT correction to complete the model density:
    \[
    \begin{aligned}
        \mathcal K
        &:=
        \sum_{\ell=1}^m
        \sum_{i\in S_\ell(\deltacls)}
        (\nabla_{\pi_i}\L+1)
        \frac{\pi_i}{\tldpi_\ell}
        (\tldpi_\ell-\pi_\ell^*)
        +
        \sum_{\ell=1}^m
        \sum_{i\in S_\ell\setminus S_\ell(\deltacls)}
        (\nabla_{\pi_i}\L+1)\pi_i \\
        &= 
        \sum_{\ell=1}^m
        \sum_{i\in S_\ell(\deltacls)}
        \frac{\pi_i}{\tldpi_\ell}(\tldpi_\ell-\pi_\ell^*)
        \E_{p^*}
        \left[
            \ell'(p)\phi(\vmu_i)
        \right]
        +
        \sum_{\ell=1}^m
        \sum_{i\in S_\ell\setminus S_\ell(\deltacls)}
        \pi_i
        \E_{p^*}
        \left[
            \ell'(p)\phi(\vmu_i)
        \right] .
    \end{aligned}
    \]
    The constants from the \(+1\) terms cancel, because
    \(S_1,\ldots,S_m\) partition \([n]\):
    \[
        \sum_{\ell=1}^m(\tldpi_\ell-\pi_\ell^*)
        +
        \sum_{\ell=1}^m
        \sum_{i\in S_\ell\setminus S_\ell(\deltacls)}
        \pi_i
        =
        \sum_{i=1}^n\pi_i-
        \sum_{\ell=1}^m\pi_\ell^*
        =0 .
    \]
    Using the gradient formula and \eqref{eq:def-tau-local}, we decompose
    \(\mathcal D+\mathcal K\) exactly as
    \[
    \begin{aligned}
        \mathcal D+\mathcal K
        &=
        \E_{p^*}
        \left[
            \ell'(p)(p-p^*)
        \right]
        +
        \sum_{\ell=1}^m
        \sum_{i\in S_\ell(\deltacls)}
        \frac{\pi_\ell^*}{\tldpi_\ell}\pi_i
        \E_{p^*}
        \left[
            \ell'(p)\phi(\vmu_i)\tau_i
        \right]
        \myge{a}
        \L-\frac{\L}{100}
        \ge
        \frac{99}{100}\L .
    \end{aligned}
    \]
    Here (a) uses convexity of \(\ell(z)=-\log z\) for the first term.  For
    the second term, \eqref{eq:local-use-close-weight} gives
    \(\tldpi_\ell\ge\pi_\ell^*/2\), and hence
    $\frac{\pi_\ell^*}{\tldpi_\ell}\pi_i\le 2\pi_i$.
    Thus the close-component case of
    \cref{lem:local-linearization-remainders} lower bounds the second term by
    \(-\L/100\).

    It remains to remove the KKT correction \(\mathcal K\). Since \(\tldpi_\ell\ge \pi_\ell^*/2\), we have $\frac{|\tldpi_\ell-\pi_\ell^*|}{\tilde\pi_\ell}\lesssim 1$.
    By \cref{lem:weight-dual-gap-consequences} and the local dual-gap condition,
    \[
        |\mathcal K|\lesssim
        \sum_{i=1}^n
        \pi_i\abs{\nabla_{\pi_i}\L+1}
        \le
        2G_\pi(\vpi,\vmu)
        \le
        2c_{\rm kkt}\L 
        \le \frac14\L,
    \]
    after choosing \(c_{\rm kkt}\) sufficiently small.
    Combining this with
    \(\mathcal D+\mathcal K\ge 99\L/100\) gives
    $\mathcal D\ge \frac12\L$.
\end{proof}

The close-component descent estimate immediately yields the gradient norm lower bound below.
\begin{lemma}[\cref{lem: grad norm lower bound main text} restated, gradient norm lower bound]
\label{lem: grad norm lower bound}
    Under the assumptions of \cref{prop:local-one-step-bootstrap},
    \[
        \norm{\nabla_{\vmu}\L}_F
        \gtrsim
        \sqrt{\frac{\pimins}{n\Xi_{\rm id}}}
        \L(\vpi,\vmu)^{3/4}.
    \]
\end{lemma}

\begin{proof}
    By \cref{lem: local descent dir},
    \begin{align*}
        \L
        \lesssim
        \sum_{\ell=1}^m
        \sum_{i\in S_\ell(\deltacls)}
        \frac{\pi_\ell^*}{\tldpi_\ell}
        \left\langle
            \nabla_{\vmu_i}\L,\vmu_i-\vmu_\ell^*
        \right\rangle
        &\mylesim{a}
        \left(
        \sum_{\ell=1}^m
        \sum_{i\in S_\ell(\deltacls)}
        \left(\frac{\pi_\ell^*}{\tldpi_\ell}\right)^2
        \norm{\vmu_i-\vmu_\ell^*}^2
        \right)^{1/2}
        \norm{\nabla_{\vmu}\L}_F\\
        &\mylesim{b}
        \sqrt n\deltacls\norm{\nabla_{\vmu}\L}_F,
    \end{align*}
    where (a) uses Cauchy--Schwarz;
    (b) uses \(\tldpi_\ell\ge\pi_\ell^*/2\) and
    \(\norm{\vmu_i-\vmu_\ell^*}\le\deltacls\) on
    \(S_\ell(\deltacls)\).
    
    Substituting
    \(\deltacls=
    2\sqrt{\Xi_{\rm id}/\pimins}\L^{1/4}\)
    yields the claim.
\end{proof}

Now we are ready to give the proof of \cref{prop:local-one-step-bootstrap}.
\begin{proof}[Proof of \cref{prop:local-one-step-bootstrap}]
    Denote $\hat\pi_\ell:=\sum_{i\in S_\ell}\pi_i$. By \eqref{eq:local-use-weight} and
    \(\Xi_{\rm id}\sqrt\eps\le c\pimins\), after decreasing \(c\) if
    necessary, we have
    $\hat\pi_\ell\ge \frac34\pi_\ell^*$ for every $\ell\in[m]$.
    Also \(U(\vmu)\le B^2\) implies
    $\frac1{\hat\pi_\ell}
        \sum_{i\in S_\ell}
        \pi_i\norm{\vmu_i-\vmu_\ell^*}^2
        \le B^2$,
    because every summand is bounded by \(U(\vmu)\). Hence
    \cref{lem:local-smoothness-self-bound} applies with
    \(c_{\rm cov}=3/4\).

    Applying its descent part with
    \(\vdelta=-\eta\nabla_{\vmu}\L(\vpi,\vmu)\), and then using
    \cref{lem: grad norm lower bound}, gives
    \[
        \L(\vpi,\vmu^+)
        \le
        \L(\vpi,\vmu)
        -\frac{\eta}{2}
        \norm{\nabla_{\vmu}\L}_F^2
        \le
        \L(\vpi,\vmu)
        -
        c\eta
        \frac{\pimins}{n\Xi_{\rm id}}
        \L(\vpi,\vmu)^{3/2}.
    \]

    To prove \eqref{eq:one-step-grad-norm}, we have
    \begin{align*}
        \eta\norm{\nabla_{\vmu}\L}_F
        &\myle{a}
        \sqrt{2\eta(\L(\vpi,\vmu)-\L(\vpi,\vmu^+))}
        =
        (\L(\vpi,\vmu)-\L(\vpi,\vmu^+))\sqrt{\frac{2\eta}{\L(\vpi,\vmu)-\L(\vpi,\vmu^+)}}\\
        &\myle{b}
        C\sqrt{A_{\rm rate}}
        \frac{\L(\vpi,\vmu)-\L(\vpi,\vmu^+)}{\L^{3/4}},
    \end{align*}
    where (a) uses $\L(\vpi,\vmu)-\L(\vpi,\vmu^+)\ge \frac{\eta}{2}\norm{\nabla_{\vmu}\L}_F^2$ from \cref{lem:local-smoothness-self-bound}; (b) uses $\L(\vpi,\vmu)-\L(\vpi,\vmu^+)\ge \frac{c\eta}{A_{\rm rate}}\L^{3/2}$ from above.
\end{proof}

\subsection{Technical lemmas}

It remains to prove the linearization estimate used in
\cref{lem: local descent dir}.  Its role is to show that the Taylor remainders
produced by replacing close model components with their assigned true centers
cannot create a negative contribution larger than a small multiple of the loss.

For \(i\in S_\ell\), also define \(\tldtau_i\) by
\begin{equation}
    \phi(\vmu_i;\vx)
    =
    \phi(\vmu_\ell^*;\vx)
    \left(
        1+\left\langle \vx-\vmu_\ell^*,
        \vmu_i-\vmu_\ell^*\right\rangle
        +\tldtau_i(\vx)
    \right).
    \label{eq:def-tldtau-local}
\end{equation}
Here \(\tau_i\) in \eqref{eq:def-tau-local} expands
\(\phi(\vmu_\ell^*)\) around \(\phi(\vmu_i)\), whereas \(\tldtau_i\) expands
\(\phi(\vmu_i)\) around \(\phi(\vmu_\ell^*)\).

\begin{lemma}[Close-component linearization remainders]
\label{lem:local-linearization-remainders}
    Under the assumptions of \cref{prop:local-one-step-bootstrap}, suppose
    \(I_\ell\subseteq S_\ell(\deltacls)\) and
    \(0\le a_{\ell i}\le C_0\pi_i\).  Then
    \[
        \sum_{\ell=1}^m\sum_{i\in I_\ell}
        a_{\ell i}
        \E_{p^*}
        \left[
            \ell'(p)\phi(\vmu_i)\tau_i
        \right]
        \ge
        -\frac1{100}\L .
    \]
\end{lemma}

\begin{proof}
    We will use $\eps=\L$. Define
    \[
        Q(\vx):=
        \sum_{\ell=1}^m\sum_{i\in I_\ell}
        a_{\ell i}\phi(\vmu_i;\vx)\tau_i(\vx).
    \]
    \paragraph{Step 1: preliminary estimates.}
    Fix \(\theta\in(0,1/2]\) to be chosen below, and set
    \(p_\theta=\theta p+(1-\theta)p^*\). We first record the several estimates.  For every fixed \(q<\infty\) and every fixed
    \(\kappa\in(0,1]\),
    \begin{align}
        \sum_{\ell=1}^m\sum_{i\in I_\ell}
        a_{\ell i}\norm{\vmu_i-\vmu_\ell^*}^2
        &\lesssim
        \Xi_{\rm id}(1+B^2)\sqrt\eps,
        \label{eq:Q-square-bound-local}\\
        \left\|\frac{Q}{p_\theta}\right\|_{q,p_\theta}
        +
        \left\|\frac{Q}{p}\right\|_{q,p_\theta}
        +
        \left\|\frac{Q}{p^*}\right\|_{q,p_\theta}
        &\le
        \sfA_{\rm loc}^{C}\sqrt\eps,
        \label{eq:Q-norm-bound}\\
        \left\|\frac{R}{p_\theta}\right\|_{2+\kappa,p_\theta}
        &\le
        \theta^{-C}\eps^{1/(2+\kappa)}.
        \label{eq:R-interpolation-local}
    \end{align}
    In \eqref{eq:Q-square-bound-local}, we use \(a_{\ell i}\le C_0\pi_i\),
    \(I_\ell\subseteq S_\ell\), and \eqref{eq:local-use-second}.  In
    \eqref{eq:Q-norm-bound}, we use \cref{lem:aggregate-remainder-bound}.  
    The last one follows from \cref{lem: local chi square}:
    \[
    \begin{aligned}
        \left\|\frac{R}{p_\theta}\right\|_{2+\kappa,p_\theta}^{2+\kappa}
        &=
        \E_{p_\theta}
        \left[
            \left|\frac{R}{p_\theta}\right|^2
            \left|\frac{R}{p_\theta}\right|^\kappa
        \right]
        \le
        \left\|\frac{R}{p_\theta}\right\|_\infty^\kappa
        \left\|\frac{R}{p_\theta}\right\|_{2,p_\theta}^2
        \lesssim
        \left(\frac2\theta\right)^\kappa
        \frac{\eps}{\theta}
        \le
        \theta^{-C}\eps .
    \end{aligned}
    \]
    
    \paragraph{Step 2: homotopy expansion.}
    We now look at the target term to bound. Choose $\theta:=c_\theta\sfA_{\rm loc}^{-C_\theta}$ with \(c_\theta>0\) small and \(C_\theta>0\) large enough.  Taylor expansion between
    \(p_\theta\) and \(p=p_\theta+(1-\theta)R\) gives,
    \[
    \begin{aligned}
        \sum_{\ell=1}^m\sum_{i\in I_\ell}
        a_{\ell i}
        \E_{p^*}
        \left[
            \ell'(p)\phi(\vmu_i)\tau_i
        \right]
        =
        \E_{p^*}[\ell'(p)Q]
        &=
        \E_{p^*}[\ell'(p_\theta)Q]
        +
        (1-\theta)\E_{p^*}[\ell''(p_\theta)RQ]
        +\mathcal R_{\ge3},
    \end{aligned}
    \]
    where
    \[
        \mathcal R_{\ge3}
        =
        (1-\theta)^2
        \E_{p^*}
        \left[
            Q R^2
            \int_0^1
            (1-u)\ell'''
            \left((1-u)p_\theta+up\right)
            \rd u
        \right].
    \]
    
    \paragraph{Step 2.1: high-order term.}
    For $\mathcal R_{\ge3}$, since \(\ell'''(z)=-2z^{-3}\), the elementary bound
    $b a^2\int_0^1
        \frac{1-u}{((1-u)a+ub)^3}\rd u
        \lesssim 1$
        $(a,b>0)$
    applied pointwise with \(a=p_\theta(\vx)\) and \(b=p(\vx)\), together
    with \(p^*\le(1-\theta)^{-1}p_\theta\), gives
    \[
    \begin{aligned}
        |\mathcal R_{\ge3}|
        &\mylesim{a}
        \E_{p_\theta}
        \left[
            \left|\frac{R}{p_\theta}\right|^2
            \left|\frac{Q}{p}\right|
        \right]
        \mylesim{b}
        \left\|\frac{R}{p_\theta}\right\|_{2+\kappa,p_\theta}^2
        \left\|\frac{Q}{p}\right\|_{(2+\kappa)/\kappa,p_\theta} 
        \mylesim{c}
        \sfA_{\rm loc}^{C}\theta^{-C}
        \eps^{2/(2+\kappa)}\sqrt\eps
        \le
        \frac{\eps}{400}.
    \end{aligned}
    \]
    Here (a) uses the preceding elementary bound and
    \(p^*\le(1-\theta)^{-1}p_\theta\); (b) is Hölder's inequality; and (c)
    uses \eqref{eq:R-interpolation-local} and \eqref{eq:Q-norm-bound}, with
    \(q=(2+\kappa)/\kappa\). The last inequality follows from
    \(\eps\le\eps_0\) and the smallness condition on \(\eps_0\).

    \paragraph{Step 2.2: first-order term.}
    This homotopy term is small because each summand in \(Q\) has
    zero integral. To see this, using \(p^*/p_\theta=1-\theta R/p_\theta\),
    \[
    \begin{aligned}
        \E_{p^*}[\ell'(p_\theta)Q]
        &=-\int \frac{p^*}{p_\theta}Q
        =-\int Q+\theta\int\frac{R}{p_\theta}Q
        =
        \theta
        \E_{p_\theta}
        \left[
            \frac{R}{p_\theta}\frac{Q}{p_\theta}
        \right],
    \end{aligned}
    \]
    where the last equality uses \eqref{eq:tau-zero-mass}.  Therefore,
    by \cref{lem: local chi square} and \eqref{eq:Q-norm-bound},
    \[
        \abs{\E_{p^*}[\ell'(p_\theta)Q]}
        \le
        \sfA_{\rm loc}^{C}\sqrt\theta \eps .
    \]
    We choose \(\theta\le c\sfA_{\rm loc}^{-2}\), so this term is at
    most \(\eps/400\).

    \paragraph{Step 2.3: second-order term.}
    We next replace the leading term by its \(p^*\)-version.  Since
    \(\ell''(z)=z^{-2}\) and \(p_\theta=p^*+\theta R\), we have
    \[
    \begin{aligned}
        \left|
        \E_{p^*}[\ell''(p_\theta)RQ]
        -
        \int\frac{R(\vx)Q(\vx)}{p^*(\vx)}\rd\vx
        \right|
        &=\left|\theta
        \int
        \frac{
            R(\vx)^2Q(\vx)(p^*(\vx)+p_\theta(\vx))
        }{
            p_\theta(\vx)^2p^*(\vx)
        }
        \rd\vx 
        \right|\\
        &\mylesim{a}
        \theta
        \E_{p_\theta}
        \left[
            \left|\frac{R}{p_\theta}\right|^2
            \left|\frac{Q}{p^*}\right|
        \right] \\
        &\mylesim{b}
        \theta
        \left\|\frac{R}{p_\theta}\right\|_{2+\kappa,p_\theta}^2
        \left\|\frac{Q}{p^*}\right\|_{(2+\kappa)/\kappa,p_\theta} \\
        &\mylesim{c}
        \sfA_{\rm loc}^{C}\theta^{-C}\eps^{1+c}
        \le
        \frac{\eps}{400}.
    \end{aligned}
    \]
    Here (a) uses \(p_\theta\ge p^*/2\), which implies
    \((p^*+p_\theta)/p_\theta\lesssim1\); (b) is Hölder's inequality; and
    (c) uses \eqref{eq:R-interpolation-local} and \eqref{eq:Q-norm-bound},
    with \(q=(2+\kappa)/\kappa\).  The last inequality follows from the
    choice of \(\eps_0\).
    
    It remains to lower bound
    \[
        J:=\int\frac{R(\vx)Q(\vx)}{p^*(\vx)}\rd\vx .
    \]

    We now expand \(R=p-p^*\) over the full Voronoi groups \(S_\ell\).  For this
    proof only, write
    \[
        \Gamma^S_{j,0}:=\sum_{k\in S_j}\pi_k-\pi_j^*,
        \qquad
        \Gamma^S_{j,1}:=
        \sum_{k\in S_j}\pi_k(\vmu_k-\vmu_j^*).
    \]
    By the expansion in \eqref{eq:def-tldtau-local},
    \[
    \begin{aligned}
        R(\vx)
        &=
        \sum_{j=1}^m
        \Gamma^S_{j,0}\phi(\vmu_j^*;\vx)
        +
        \sum_{j=1}^m
        \phi(\vmu_j^*;\vx)
        \left\langle \vx-\vmu_j^*,\Gamma^S_{j,1}\right\rangle
        +
        \sum_{j=1}^m
        \phi(\vmu_j^*;\vx)
        \sum_{k\in S_j}\pi_k\tldtau_k(\vx).
    \end{aligned}
    \]
    Accordingly, write \(J=J_0+J_1+J_2\), with \(J_0,J_1,J_2\)
    corresponding to the three displayed terms.

    \paragraph{Step 2.3.1.} For \(J_0\), add and subtract the
    same-cluster leading term.  By \eqref{eq:tau-zero-mass},
    \[
    \begin{aligned}
        J_0
        &=
        \sum_{\ell=1}^m
        \sum_{i\in I_\ell}
        \frac{\Gamma^S_{\ell,0}a_{\ell i}}{\pi_\ell^*}
        \int
            \phi(\vmu_i;\vx)\tau_i(\vx)\rd\vx
        +\mathcal R_0
        =
        \mathcal R_0,
    \end{aligned}
    \]
    where
    \[
    \begin{aligned}
        \mathcal R_0
        &:=
        \sum_{\ell=1}^m
        \sum_{i\in I_\ell}
        \Gamma^S_{\ell,0}a_{\ell i}
        \int
        \left(
            \frac{\phi(\vmu_\ell^*;\vx)}{p^*(\vx)}
            -
            \frac1{\pi_\ell^*}
        \right)
        \phi(\vmu_i;\vx)\tau_i(\vx)\rd\vx\\
        &\quad+
        \sum_{\substack{j,\ell\in[m]\\j\ne\ell}}
        \sum_{i\in I_\ell}
        \Gamma^S_{j,0}a_{\ell i}
        \int
        \frac{
            \phi(\vmu_j^*;\vx)\phi(\vmu_i;\vx)\tau_i(\vx)
        }{
            p^*(\vx)
        }
        \rd\vx .
    \end{aligned}
    \]
    Since
    $\frac{\phi(\vmu_\ell^*;\vx)}{p^*(\vx)}
        -
        \frac1{\pi_\ell^*}
        =
        -\frac1{\pi_\ell^*}
        \sum_{s\ne\ell}
        \pi_s^*
        \frac{\phi(\vmu_s^*;\vx)}{p^*(\vx)}$,
    \eqref{eq:separated-overlap-tau}, \eqref{eq:local-use-weight}, and \eqref{eq:Q-square-bound-local} give
    \[
        |J_0|
        = |\mathcal R_0|
        \le
        \sfA_{\rm loc}^{C}e^{-c(\Delta-CB)^2}
        \left(
            \sum_{j=1}^m\abs{\Gamma^S_{j,0}}
        \right)
        \left(
            \sum_{\ell=1}^m\sum_{i\in I_\ell}
            a_{\ell i}\norm{\vmu_i-\vmu_\ell^*}^2
        \right)
        \le
        \sfA_{\rm loc}^{C}e^{-c(\Delta-CB)^2}\eps .
    \]

    \paragraph{Step 2.3.2.} Similarly, for \(J_1\), the same-cluster leading term cancels by
    \eqref{eq:tau-zero-first}:
    \[
    \begin{aligned}
        J_1
        &=
        \sum_{\ell=1}^m
        \sum_{i\in I_\ell}
        \frac{a_{\ell i}}{\pi_\ell^*}
        \int
            \phi(\vmu_i;\vx)\tau_i(\vx)
            \left\langle
                \vx-\vmu_\ell^*,\Gamma^S_{\ell,1}
            \right\rangle
        \rd\vx
        +\mathcal R_1
        =
        \mathcal R_1,
    \end{aligned}
    \]
    where
    \[
    \begin{aligned}
        \mathcal R_1
        &:=
        \sum_{\ell=1}^m
        \sum_{i\in I_\ell}
        a_{\ell i}
        \int
        \left(
            \frac{\phi(\vmu_\ell^*;\vx)}{p^*(\vx)}
            -
            \frac1{\pi_\ell^*}
        \right)
        \phi(\vmu_i;\vx)\tau_i(\vx)
        \left\langle
            \vx-\vmu_\ell^*,\Gamma^S_{\ell,1}
        \right\rangle
        \rd\vx\\
        &\quad+
        \sum_{\substack{j,\ell\in[m]\\j\ne\ell}}
        \sum_{i\in I_\ell}
        a_{\ell i}
        \int
        \frac{
            \phi(\vmu_j^*;\vx)\phi(\vmu_i;\vx)\tau_i(\vx)
        }{
            p^*(\vx)
        }
        \left\langle
            \vx-\vmu_j^*,\Gamma^S_{j,1}
        \right\rangle
        \rd\vx .
    \end{aligned}
    \]
    Therefore, by \eqref{eq:separated-overlap-linear},
    \eqref{eq:local-use-first}, and \eqref{eq:Q-square-bound-local} ,
    \[
        |\mathcal R_1|
        \le
        \sfA_{\rm loc}^{C}e^{-c(\Delta-CB)^2}
        \left(
            \sum_{j=1}^m\norm{\Gamma^S_{j,1}}
        \right)
        \left(
            \sum_{\ell=1}^m\sum_{i\in I_\ell}
            a_{\ell i}\norm{\vmu_i-\vmu_\ell^*}^2
        \right)
        \le
        \sfA_{\rm loc}^{C}e^{-c(\Delta-CB)^2}\eps .
    \]

    \paragraph{Step 2.3.3.}
    It remains to bound \(J_2\) from below.  Split it into the leading
    same-cluster term and an error:
    \[
    \begin{aligned}
        J_2
        &=
        \sum_{\ell=1}^m
        \sum_{i\in I_\ell}
        \sum_{k\in S_\ell}
        \frac{a_{\ell i}\pi_k}{\pi_\ell^*}
        \int
            \phi(\vmu_i;\vx)\tau_i(\vx)\tldtau_k(\vx)
        \rd\vx
        +\mathcal R_2
        \myge{a}
        \mathcal R_2,
    \end{aligned}
    \]
    where (a) uses \eqref{eq:same-cluster-positive}, and
    \[
    \begin{aligned}
        \mathcal R_2
        &:=
        \sum_{\ell=1}^m
        \sum_{i\in I_\ell}
        \sum_{k\in S_\ell}
        a_{\ell i}\pi_k
        \int
        \left(
            \frac{\phi(\vmu_\ell^*;\vx)}{p^*(\vx)}
            -
            \frac1{\pi_\ell^*}
        \right)
        \phi(\vmu_i;\vx)\tau_i(\vx)\tldtau_k(\vx)
        \rd\vx\\
        &\quad+
        \sum_{\substack{j,\ell\in[m]\\j\ne\ell}}
        \sum_{k\in S_j}
        \sum_{i\in I_\ell}
        \pi_k a_{\ell i}
        \int
        \frac{
            \phi(\vmu_j^*;\vx)\phi(\vmu_i;\vx)
            \tau_i(\vx)\tldtau_k(\vx)
        }{
            p^*(\vx)
        }
        \rd\vx .
    \end{aligned}
    \]
    Using the identity for
    \(\phi(\vmu_\ell^*)/p^*-1/\pi_\ell^*\) above and then applying
    \eqref{eq:separated-overlap-second}, we obtain
    \[
    \begin{aligned}
        |\mathcal R_2|
        &\le
        \sfA_{\rm loc}^{C}e^{-c(\Delta-CB)^2}
        \left(
            \sum_{\ell=1}^m\sum_{i\in I_\ell}
            a_{\ell i}\norm{\vmu_i-\vmu_\ell^*}^2
        \right)
        \left(
            \sum_{j=1}^m\sum_{k\in S_j}
            \pi_k\norm{\vmu_k-\vmu_j^*}^2
        \right)
        \le
        \sfA_{\rm loc}^{C}e^{-c(\Delta-CB)^2}\eps,
    \end{aligned}
    \]
    where the last line uses \eqref{eq:Q-square-bound-local}  and
    \eqref{eq:local-use-second}. 

    Combining the bounds for \(J_0\), \(J_1\), and \(J_2\), we obtain
    \[
        J
        =
        \int\frac{R(\vx)Q(\vx)}{p^*(\vx)}\rd\vx
        \ge
        -\sfA_{\rm loc}^{C}e^{-c(\Delta-CB)^2}\eps
        \ge - \frac{\eps}{400}
        \quad\Rightarrow\quad
        (1-\theta)\E_{p^*}[\ell''(p_\theta)RQ]
        \ge
        -\frac{\eps}{200},
    \]
    after adjusting the universal constants.

    \paragraph{Summary.}
    Combining the Taylor expansion with the bounds from Steps 2.1, 2.2, and
    2.3 gives
    \[
    \begin{aligned}
        \E_{p^*}[\ell'(p)Q]
        &\ge
        -\frac{\eps}{400}
        -
        \frac{\eps}{400}
        -
        \frac{\eps}{200}
        =-\frac{\eps}{100}.
    \end{aligned}
    \]

\end{proof}

\subsection{Auxiliary analytic estimates}

We finish by proving the auxiliary estimates used in the linearization and
one-step arguments.

\begin{lemma}[Local cell Gaussian ratio moment bound]
\label{lem:close-cell-gaussian-ratio-moment}
    Under the assumptions of \cref{prop:local-one-step-bootstrap}, the
    conclusions of \cref{lem:cell-gaussian-ratio-moment} hold with
    \(\sfA_{\rm cell}^{C}\) replaced by \(\sfA_{\rm loc}^{C}\).
    More explicitly, let \(\Omega_1,\ldots,\Omega_m\) denote the true Voronoi
    cells.  Fix constants \(1\le q<\infty\), \(0\le K<\infty\), and
    \(C_0<\infty\).  Let \(0\le\rho\le B\), and 
    \(\vh,\vh',\vw\in\mathbb R^d\) satisfy \(\norm{\vh},\norm{\vh'}\le\rho\).  Suppose
    \(\vnu\) is either a true center \(\vmu_r^*\) or a model center
    \(\vmu_j\) with \(j\in S_r\).  Then, for every \(\ell\in[m]\),
    \begin{align}
        &\sum_{s=1}^m
        \int_{\Omega_s}
        \phi(\vnu;\vx)
        \left(
            \frac{\phi(\vmu_\ell^*;\vx)}
            {\phi(\vmu_s^*;\vx)}
        \right)^q
        \left|
            \left\langle
                \vx-\vmu_\ell^*,\vw
            \right\rangle
        \right|^K
        \exp\left(
            C_0
            \left|
                \left\langle
                    \vx-\vmu_\ell^*,\vh
                \right\rangle
            \right|
            +
            C_0|\ip{\vx-\vmu_s^*,\vh'}|
            +
            C_0\rho^2
        \right)
        \rd\vx
        \notag\\
        &\hspace{55mm}\le
        e^{CB\rho+C\rho^2}\sfA_{\rm loc}^{C}\norm{\vw}^K .
        \label{eq:cell-gaussian-ratio-homogeneous-local}
    \end{align}
    For \(K=0\), the factor \(\norm{\vw}^K\) is interpreted as one.  The
    analogous estimate with
    \(1+\abs{\ip{\vx-\vmu_\ell^*,\vw}}^K\) on the left has right-hand side
    \[
        e^{CB\rho+C\rho^2}\sfA_{\rm loc}^{C}
        \left(1+\norm{\vw}^K\right).
    \]
    In particular, the right-hand side is
    \(e^{CB^2}\sfA_{\rm loc}^{C}(1+\norm{\vw}^K)\) when
    \(\rho=B\).  If \(\rho=\deltacls\) and \((1+B)\deltacls\le c\), then it is
    \(\sfA_{\rm loc}^{C}(1+\norm{\vw}^K)\).
\end{lemma}

\begin{proof}
    Under \cref{prop:local-one-step-bootstrap}, any admissible \(\vnu\) is
    either a true center or a model center \(\vmu_j\in S_r\), and hence
    \(\norm{\vnu-\vmu_r^*}\le B\).  Also
    \(\sfA_{\rm cell}\le \sfA_{\rm loc}^{C}\), after increasing the
    universal constant \(C\).  The local separation condition implies the
    separation condition in \cref{lem:cell-gaussian-ratio-moment}, since
    \(\rho\le B\).  Applying
    \cref{lem:cell-gaussian-ratio-moment} gives
    \[
        e^{CB\rho+C\rho^2}
        \sfA_{\rm loc}^{C}.
    \]
    When $\rho=\deltacls$ The terms \(C B\rho+C\rho^2\) are absorbed into
    \(\sfA_{\rm loc}^{C}\), because
    the local smallness condition gives
    \((1+B)\deltacls\le c\).  This proves the claim.
\end{proof}

\begin{lemma}[Close-displacement ratio bound]
\label{lem:close-displacement-ratio-bound}
    Suppose \(i\in S_\ell(\deltacls)\), \(j\in S_r\), and
    \((1+B)\deltacls\le c\).  Then, for every fixed \(q<\infty\),
    \[
        \left\|
            \frac{\phi(\vmu_i)\tau_i}{p^*}
        \right\|_{q,\phi(\vmu_j)}
        \le
        \sfA_{\rm loc}^{C}
        \norm{\vmu_i-\vmu_\ell^*}^2,
        \quad
        \left\|
            \frac{\phi(\vmu_i)\tau_i}{p}
        \right\|_{q,p^*}
        \le
        \sfA_{\rm loc}^{C}
        \norm{\vmu_i-\vmu_\ell^*}^2 .
    \]
    Also, for every \(s\in[m]\),
    \[
        \left\|
            \frac{\phi(\vmu_i)\tau_i}{p^*}
        \right\|_{q,\phi(\vmu_s^*)}
        \le
        \sfA_{\rm loc}^{C}
        \norm{\vmu_i-\vmu_\ell^*}^2 .
    \]
\end{lemma}

\begin{proof}
    Write \(\vh:=\vmu_i-\vmu_\ell^*\).  If \(\vh=0\), then
    \(\tau_i=0\), so the claim is immediate.  Otherwise set
    \(\bar\vh:=\vh/\norm{\vh}\).  Taylor's theorem and the Gaussian ratio
    identity give
    \begin{align}
        \frac{\phi(\vmu_i;\vx)}{\phi(\vmu_\ell^*;\vx)}
        &=
        \exp\left(
            \left\langle
                \vx-\vmu_\ell^*,\vh
            \right\rangle
            -\frac12\norm{\vh}^2
        \right),
        \label{eq:close-density-ratio-unified}\\
        \abs{\tau_i(\vx)}
        &\le
        C\norm{\vh}^2
        \left(
            1+
            \left|
                \left\langle
                    \vx-\vmu_\ell^*,\bar\vh
                \right\rangle
            \right|^2
        \right)
        \exp\left(
            C\norm{\vh}
            \left|
                \left\langle
                    \vx-\vmu_\ell^*,\bar\vh
                \right\rangle
            \right|
            +C\norm{\vh}^2
        \right).
        \label{eq:close-tau-pointwise-bound-unified}
    \end{align}
    Since \(\norm{\vh}\le\deltacls\), the product
    \(\phi(\vmu_i)\tau_i\) is bounded by
    \(\phi(\vmu_\ell^*)\norm{\vh}^2\) times the right-hand side in
    \cref{lem:close-cell-gaussian-ratio-moment} with
    \(\rho=\deltacls\), \(\vh\) as the exponential vector, and
    \(\vw=\bar\vh\).

    We first prove the \(p^*\)-denominator estimates.  Split the integral over
    \(\Omega_s\).  On \(\Omega_s\),
    \(p^*(\vx)\ge \pimins\phi(\vmu_s^*;\vx)\).  Therefore
    \eqref{eq:close-density-ratio-unified}--\eqref{eq:close-tau-pointwise-bound-unified}
    and \cref{lem:close-cell-gaussian-ratio-moment}, with
    \(\rho=\deltacls\), give
    \[
        \left\|
            \frac{\phi(\vmu_i)\tau_i}{p^*}
        \right\|_{q,\phi(\vmu_j)}
        +
        \left\|
            \frac{\phi(\vmu_i)\tau_i}{p^*}
        \right\|_{q,\phi(\vmu_s^*)}
        \le
        \sfA_{\rm loc}^{C}\norm{\vh}^2,
    \]
    where the factor \(e^{CB\deltacls+C\deltacls^2}\) is universal because
    \((1+B)\deltacls\le c\).

    It remains to handle the \(p\)-denominator under the ambient measure
    \(p^*\).  On \(\Omega_s\), the close-weight lower bound
    \eqref{eq:local-use-close-weight} gives
    \[
        p(\vx)
        \ge
        c\pi_s^*\phi(\vmu_s^*;\vx)
        \exp\left(
            -C\ip{\vx-\vmu_s^*,\vh'}
            -C\deltacls^2
        \right).
        \quad\text{with some $\norm{\vh'}\le \deltacls$}
    \]
    Also \(p^*(\vx)\le\phi(\vmu_s^*;\vx)\) on \(\Omega_s\).  Thus the
    corresponding \(q\)-th moment is bounded by the same cell integral as
    above, with the additional factor
    \(\exp(Cq|\ip{\vx-\vmu_s^*,\vh'}|+Cq\deltacls^2)\).  This is included
    in \cref{lem:close-cell-gaussian-ratio-moment} with
    \(\rho=\deltacls\).  Hence
    \[
        \left\|
            \frac{\phi(\vmu_i)\tau_i}{p}
        \right\|_{q,p^*}
        \le
        \sfA_{\rm loc}^{C}\norm{\vh}^2 .
    \]
\end{proof}

\begin{lemma}
\label{lem:aggregate-remainder-bound}
    Under the assumptions of \cref{prop:local-one-step-bootstrap}, suppose
    \(I_\ell\subseteq S_\ell(\deltacls)\) and
    \(0\le a_{\ell i}\le C_0\pi_i\).  Consider
    \[
        Q_\tau(\vx):=
        \sum_{\ell=1}^m\sum_{i\in I_\ell}
        a_{\ell i}\phi(\vmu_i;\vx)\tau_i(\vx).
    \]
    Then, for every fixed \(1\le q<\infty\) and every
    \(\theta\in(0,1/2]\),
    \[
        \left\|\frac{Q_\tau}{p_\theta}\right\|_{q,p_\theta}
        +
        \left\|\frac{Q_\tau}{p}\right\|_{q,p_\theta}
        +
        \left\|\frac{Q_\tau}{p^*}\right\|_{q,p_\theta}
        \le
        \sfA_{\rm loc}^{C}\sqrt\eps .
    \]
\end{lemma}

\begin{proof}
    Since \(a_{\ell i}\le C_0\pi_i\), \(I_\ell\subseteq S_\ell\), and
    \eqref{eq:local-use-second} holds,
    \begin{equation}
        \sum_{\ell=1}^m\sum_{i\in I_\ell}
        a_{\ell i}\norm{\vmu_i-\vmu_\ell^*}^2
        \le \sfA_{\rm loc}^{C}\sqrt\eps .
        \label{eq:Qtau-EQ-bound}
    \end{equation}
    Moreover, \(I_\ell\subseteq S_\ell(\deltacls)\) implies
    \(\norm{\vmu_i-\vmu_\ell^*}^2\le\deltacls^2\).  The Taylor bound in
    \eqref{eq:close-tau-pointwise-bound-unified} gives, for every fixed
    \(1\le q<\infty\),
    \begin{equation}
        \int \phi(\vmu_i;\vx)\abs{\tau_i(\vx)}^q\rd\vx
        \le
        \sfA_{\rm loc}^{C}
        \norm{\vmu_i-\vmu_\ell^*}^{2q}.
        \label{eq:Qtau-single-tau-moment-revised}
    \end{equation}

    We prove the four base estimates
    \begin{equation}
    \begin{aligned}
        \left\|\frac{Q_\tau}{p}\right\|_{q,p}
        +
        \left\|\frac{Q_\tau}{p^*}\right\|_{q,p^*}
        +
        \left\|\frac{Q_\tau}{p}\right\|_{q,p^*}
        +
        \left\|\frac{Q_\tau}{p^*}\right\|_{q,p}
        \le
        \sfA_{\rm loc}^{C}\sqrt\eps .
    \end{aligned}
    \label{eq:Qtau-four-base-bounds-revised}
    \end{equation}
    These estimates imply the claimed \(p_\theta\)-bounds because
    \(p_\theta=\theta p+(1-\theta)p^*\) and
    \(p_\theta\ge(1-\theta)p^*\ge p^*/2\).

    For the first base estimate, Jensen's inequality and
    \(a_{\ell i}\le C_0\pi_i\) give
    \[
    \begin{aligned}
        \left\|\frac{Q_\tau}{p}\right\|_{q,p}^{q}
        &\le
        \sfA_{\rm loc}^{C}
        \sum_{\ell=1}^m\sum_{i\in I_\ell}
        a_{\ell i}
        \int \phi(\vmu_i;\vx)\abs{\tau_i(\vx)}^q\rd\vx 
        \le
        \sfA_{\rm loc}^{C}
        \sum_{\ell=1}^m\sum_{i\in I_\ell}
        a_{\ell i}\norm{\vmu_i-\vmu_\ell^*}^{2q}\\
        &\le
        \sfA_{\rm loc}^{C}\deltacls^{2(q-1)}
        \sum_{\ell=1}^m\sum_{i\in I_\ell}
        a_{\ell i}\norm{\vmu_i-\vmu_\ell^*}^2
        \le
        \sfA_{\rm loc}^{C}\eps^{q/2}.
    \end{aligned}
    \]
    Thus \(\norm{Q_\tau/p}_{q,p}\le\sfA_{\rm loc}^{C}\sqrt\eps\).

    The remaining three base estimates follow from
    \cref{lem:close-displacement-ratio-bound}.  Namely,
    \[
    \begin{aligned}
        \left\|\frac{Q_\tau}{p^*}\right\|_{q,p^*}
        &\le
        \sum_{\ell=1}^m\sum_{i\in I_\ell}
        a_{\ell i}
        \left\|
            \frac{\phi(\vmu_i)\tau_i}{p^*}
        \right\|_{q,p^*}
        \le
        \sfA_{\rm loc}^{C}\sum_{\ell=1}^m\sum_{i\in I_\ell}
        a_{\ell i}\norm{\vmu_i-\vmu_\ell^*}^2
        \le \sfA_{\rm loc}^{C} \sqrt{\eps},\\
        \left\|\frac{Q_\tau}{p}\right\|_{q,p^*}
        &\le
        \sum_{\ell=1}^m\sum_{i\in I_\ell}
        a_{\ell i}
        \left\|
            \frac{\phi(\vmu_i)\tau_i}{p}
        \right\|_{q,p^*}
        \le
        \sfA_{\rm loc}^{C}\sum_{\ell=1}^m\sum_{i\in I_\ell}
        a_{\ell i}\norm{\vmu_i-\vmu_\ell^*}^2
        \le \sfA_{\rm loc}^{C} \sqrt{\eps}.
    \end{aligned}
    \]
    Finally, expanding \(p=\sum_{r=1}^m\sum_{j\in S_r}\pi_j\phi(\vmu_j)\) and
    applying Minkowski's inequality,
    \[
    \begin{aligned}
        \left\|\frac{Q_\tau}{p^*}\right\|_{q,p}
        &\le
        \sum_{\ell=1}^m\sum_{i\in I_\ell}
        a_{\ell i}
        \left(
            \sum_{r=1}^m\sum_{j\in S_r}
            \pi_j
            \left\|
                \frac{\phi(\vmu_i)\tau_i}{p^*}
            \right\|_{q,\phi(\vmu_j)}^q
        \right)^{1/q}
        \le
        \sfA_{\rm loc}^{C}\sum_{\ell=1}^m\sum_{i\in I_\ell}
        a_{\ell i}\norm{\vmu_i-\vmu_\ell^*}^2
        \le \sfA_{\rm loc}^{C} \sqrt{\eps}.
    \end{aligned}
    \]
    Combining these bounds with \eqref{eq:Qtau-EQ-bound} proves
    \eqref{eq:Qtau-four-base-bounds-revised}, and hence the lemma.
\end{proof}

\begin{lemma}
\label{lem: gaussian cancel positive}
    Recall the definitions of \(\tau_i\) and \(\tldtau_i\) in
    \eqref{eq:def-tau-local} and \eqref{eq:def-tldtau-local}.  For
    \(i\in S_\ell\),
    \begin{align}
        \int \phi(\vmu_i;\vx)\tau_i(\vx)\rd\vx&=0,
        \qquad
        \int \phi(\vmu_\ell^*;\vx)\tldtau_i(\vx)\rd\vx=0,
        \label{eq:tau-zero-mass}\\
        \int
        (\vx-\vmu_\ell^*)
        \phi(\vmu_i;\vx)\tau_i(\vx)\rd\vx&=0,
        \qquad
        \int
        (\vx-\vmu_\ell^*)\phi(\vmu_\ell^*;\vx)
        \tldtau_i(\vx)\rd\vx=0.
        \label{eq:tau-zero-first}
    \end{align}
    Moreover, for \(i,k\in S_\ell\),
    \begin{align}
        \int
        \phi(\vmu_i;\vx)\tau_i(\vx)\tldtau_k(\vx)\rd\vx
        &\ge0,
        \label{eq:same-cluster-positive}\\
        \left|
        \int
        \phi(\vmu_\ell^*;\vx)
        \tldtau_i(\vx)\tldtau_k(\vx)\rd\vx
        \right|
        &\le
        e^{CB^2}\sfA_{\rm loc}^{C}
        \norm{\vmu_i-\vmu_\ell^*}^2
        \norm{\vmu_k-\vmu_\ell^*}^2 .
        \label{eq:tldtau-second-moment-bound}
    \end{align}
\end{lemma}

\begin{proof}
    The four identities are exact moment cancellations.  Indeed,
    \[
    \begin{aligned}
        \int \phi(\vmu_i)\tau_i
        &=
        \int \phi(\vmu_\ell^*)
        -
        \int \phi(\vmu_i)
        +
        \int
        \phi(\vmu_i)
        \left\langle
            \vx-\vmu_i,
            \vmu_i-\vmu_\ell^*
        \right\rangle
        =
        0,\\
        \int
        (\vx-\vmu_\ell^*)\phi(\vmu_i)\tau_i
        &=
        \int
        (\vx-\vmu_\ell^*)\phi(\vmu_\ell^*)
        -
        \int
        (\vx-\vmu_\ell^*)\phi(\vmu_i)
        +
        \int
        (\vx-\vmu_\ell^*)\phi(\vmu_i)
        \left\langle
            \vx-\vmu_i,
            \vmu_i-\vmu_\ell^*
        \right\rangle\\
        &=
        0
        -
        (\vmu_i-\vmu_\ell^*)
        +
        (\vmu_i-\vmu_\ell^*)
        =
        0.
    \end{aligned}
    \]
    The two identities involving \(\tldtau_i\) follow similarly from
    \eqref{eq:def-tldtau-local}, first by integrating directly and then by
    multiplying by \(\vx-\vmu_\ell^*\) before integrating.

    For positivity, a direct calculation under \(X\sim N(\vmu_i,I)\), or
    equivalently the Hermite expansion of the two second-order remainders,
    gives
    \[
    \begin{aligned}
        \int \phi(\vmu_i)\tau_i\tldtau_k
        &=
        \exp\left(
            \left\langle
                \vmu_i-\vmu_\ell^*,
                \vmu_k-\vmu_\ell^*
            \right\rangle
        \right)
        \left[
            \exp\left(
                -
                \left\langle
                    \vmu_i-\vmu_\ell^*,
                    \vmu_k-\vmu_\ell^*
                \right\rangle
            \right)
            -
            1
            +
            \left\langle
                \vmu_i-\vmu_\ell^*,
                \vmu_k-\vmu_\ell^*
            \right\rangle
        \right]
        \ge 0,
    \end{aligned}
    \]
    since \(e^u\ge 1+u\) for all \(u\).
    
    Finally, write \(X=\vmu_\ell^*+Z\), where \(Z\sim N(0,I)\).  Then
    \[
        \tldtau_i(X)
        =
        \exp\left(
            \left\langle
                Z,
                \vmu_i-\vmu_\ell^*
            \right\rangle
            -
            \frac12\norm{\vmu_i-\vmu_\ell^*}^2
        \right)
        -
        1
        -
        \left\langle
            Z,
            \vmu_i-\vmu_\ell^*
        \right\rangle .
    \]
    The elementary inequality \(|e^u-1-u|\lesssim u^2e^{|u|}\) implies
    \[
    \begin{aligned}
        |\tldtau_i(X)|
        \lesssim
        \left(
            \norm{\vmu_i-\vmu_\ell^*}^2
            +
            \left|
                \left\langle
                    Z,
                    \vmu_i-\vmu_\ell^*
                \right\rangle
            \right|^2
        \right)
        \exp\left(
            \left|
                \left\langle
                    Z,
                    \vmu_i-\vmu_\ell^*
                \right\rangle
            \right|
            +
            \norm{\vmu_i-\vmu_\ell^*}^2
        \right).
    \end{aligned}
    \]
    Applying Cauchy--Schwarz and using the fixed Gaussian exponential-moment
    bound for
    \(\norm{\vmu_i-\vmu_\ell^*},\norm{\vmu_k-\vmu_\ell^*}\le B\), we obtain
    \[
    \begin{aligned}
        \left|
        \E\left[\tldtau_i(X)\tldtau_k(X)\right]
        \right|
        &\le
        e^{CB^2}\sfA_{\rm loc}^{C}
        \norm{\vmu_i-\vmu_\ell^*}^2
        \norm{\vmu_k-\vmu_\ell^*}^2 .
    \end{aligned}
    \]
    This proves \eqref{eq:tldtau-second-moment-bound}.
\end{proof}

\begin{lemma}[Separated overlap estimates]
\label{lem: separated overlap}
    Work under the assumptions of \cref{prop:local-one-step-bootstrap}.  Let
    \(i\in S_\ell\), let \(s\ne\ell\), and let \(\vv,\vw\in\mathbb R^d\).  Then,
    for every \(r,t\in\{\ell,s\}\),
    \begin{align}
        \left|
        \E_{p^*}
        \left[
        \frac{\phi(\vmu_s^*)}{p^*}
        \frac{\phi(\vmu_i)}{p^*}
        \tau_i
        \right]
        \right|
        &\le
        \sfA_{\rm loc}^{C}e^{-c(\Delta-CB)^2}
        \norm{\vmu_i-\vmu_\ell^*}^2,
        \label{eq:separated-overlap-tau}\\
        \left|
        \E_{p^*}
        \left[
        \frac{\phi(\vmu_s^*)}{p^*}
        \frac{\phi(\vmu_i)}{p^*}
        \tau_i
        \left\langle \vx-\vmu_r^*,\vv\right\rangle
        \right]
        \right|
        &\le
        \sfA_{\rm loc}^{C}e^{-c(\Delta-CB)^2}
        \norm{\vmu_i-\vmu_\ell^*}^2\norm{\vv} .
        \label{eq:separated-overlap-linear}
    \end{align}
    Moreover, if \(k\in S_\rho\) with \(\rho\in\{\ell,s\}\), then
    \begin{equation}
        \left|
        \E_{p^*}
        \left[
        \frac{\phi(\vmu_s^*)}{p^*}
        \frac{\phi(\vmu_i)}{p^*}
        \tau_i\tldtau_k
        \right]
        \right|
        \le
        \sfA_{\rm loc}^{C}e^{-c(\Delta-CB)^2}
        \norm{\vmu_i-\vmu_\ell^*}^2
        \norm{\vmu_k-\vmu_\rho^*}^2 .
        \label{eq:separated-overlap-second}
    \end{equation}

    The following true-center estimates also hold.  For every \(s\ne\ell\)
    and every \(r,t\in\{\ell,s\}\),
    \begin{align}
        \left|
        \E_{p^*}
        \left[
        \frac{\phi(\vmu_s^*)}{p^*}
        \frac{\phi(\vmu_\ell^*)}{p^*}
        \left\langle \vx-\vmu_r^*,\vv\right\rangle
        \right]
        \right|
        &\le
        \sfA_{\rm loc}^{C}e^{-c(\Delta-CB)^2}\norm{\vv},
        \label{eq:separated-overlap-true-linear}\\
        \left|
        \E_{p^*}
        \left[
        \frac{\phi(\vmu_s^*)}{p^*}
        \frac{\phi(\vmu_\ell^*)}{p^*}
        \left\langle \vx-\vmu_r^*,\vv\right\rangle
        \left\langle \vx-\vmu_t^*,\vw\right\rangle
        \right]
        \right|
        &\le
        \sfA_{\rm loc}^{C}e^{-c(\Delta-CB)^2}
        \norm{\vv} \norm{\vw}.
        \label{eq:separated-overlap-true-bilinear}
    \end{align}
    If \(k\in S_\rho\) and \(h\in S_\eta\), with
    \(\rho,\eta\in\{\ell,s\}\), then
    \begin{align}
        \left|
        \E_{p^*}
        \left[
        \frac{\phi(\vmu_s^*)}{p^*}
        \frac{\phi(\vmu_\ell^*)}{p^*}
        \tldtau_k
        \right]
        \right|
        &\le
        \sfA_{\rm loc}^{C}e^{-c(\Delta-CB)^2}
        \norm{\vmu_k-\vmu_\rho^*}^2,
        \label{eq:separated-overlap-true-tldtau}\\
        \left|
        \E_{p^*}
        \left[
        \frac{\phi(\vmu_s^*)}{p^*}
        \frac{\phi(\vmu_\ell^*)}{p^*}
        \left\langle \vx-\vmu_r^*,\vv\right\rangle
        \tldtau_k
        \right]
        \right|
        &\le
        \sfA_{\rm loc}^{C}e^{-c(\Delta-CB)^2}
        \norm{\vv}\norm{\vmu_k-\vmu_\rho^*}^2,
        \label{eq:separated-overlap-true-linear-tldtau}\\
        \left|
        \E_{p^*}
        \left[
        \frac{\phi(\vmu_s^*)}{p^*}
        \frac{\phi(\vmu_\ell^*)}{p^*}
        \tldtau_k\tldtau_h
        \right]
        \right|
        &\le
        \sfA_{\rm loc}^{C}e^{-c(\Delta-CB)^2}
        \norm{\vmu_k-\vmu_\rho^*}^2
        \norm{\vmu_h-\vmu_\eta^*}^2 .
        \label{eq:separated-overlap-true-second}
    \end{align}
\end{lemma}
\begin{proof}
    We use the product-form off-diagonal estimate
    \eqref{eq:cell-gaussian-ratio-product-offdiag} from
    \cref{lem:cell-gaussian-ratio-moment}.  Under the assumptions of
    \cref{prop:local-one-step-bootstrap}, any model center \(\vmu_j\in S_r\)
    satisfies \(\norm{\vmu_j-\vmu_r^*}\le B\), so the hypotheses of
    \cref{lem:cell-gaussian-ratio-moment} apply after replacing
    \(\sfA_{\rm cell}^{C}\) by \(\sfA_{\rm loc}^{C}\).  We will
    freely weaken \(e^{-c\Delta^2}\) to \(e^{-c(\Delta-CB)^2}\).

    In the applications below, the two relevant true centers are always
    \(\vmu_\ell^*\) and \(\vmu_s^*\).  These are exactly the ambient true
    center and the ratio numerator in the product-form estimate.  Hence the
    polynomial factors and the Taylor-remainder exponentials in \(H\), after
    applying the Taylor bounds below, are covered by
    \eqref{eq:cell-gaussian-ratio-product-offdiag}.

    Let \(\Omega_a\) be the Voronoi cell of \(\vmu_a^*\) among the true centers.  We prove all estimates by decomposing over the true cells. 
    For any measurable factor \(H\), decomposing over the true cells and using
    \(p^*(\vx)\ge \pi_a^*\phi(\vmu_a^*;\vx)\) on \(\Omega_a\), we have
    \[
    \begin{aligned}
        &\E_{p^*}
        \left[
        \frac{\phi(\vmu_s^*)}{p^*}
        \frac{\phi(\vmu_i)}{p^*}
        |H|
        \right]
        \le
        \frac{1}{\pimins}
        \sum_{a=1}^m
        \int_{\Omega_a}
        \phi(\vmu_i;\vx)
        \frac{\phi(\vmu_s^*;\vx)}
             {\phi(\vmu_a^*;\vx)}
        |H(\vx)|\rd\vx .
    \end{aligned}
    \]
    Since \(i\in S_\ell\), we have
    \(\norm{\vmu_i-\vmu_\ell^*}\le B\).  For each cell \(\Omega_a\), apply
    \eqref{eq:cell-gaussian-ratio-product-offdiag} with \(q=1\).  Since the ambient true group is \(\ell\ne s\), every cell contribution is off-diagonal and carries the factor \(e^{-c\Delta^2}\), which we weaken to \(e^{-c(\Delta-CB)^2}\).  The finite sum over cells and the factor \(1/\pimins\) are absorbed into
    \(\sfA_{\rm loc}^{C}\).

    The explicit powers on the right-hand sides come from the Taylor bounds for \(i\in S_\ell\) and \(k\in S_\rho\)
    \[
    \begin{aligned}
    |\tau_i(\vx)|
    &\le
    C\left(
        \left|
            \left\langle
                \vx-\vmu_\ell^*,
                \vmu_i-\vmu_\ell^*
            \right\rangle
        \right|^2
        +
        \norm{\vmu_i-\vmu_\ell^*}^2
    \right)
    \exp\left(
        C\left|
            \left\langle
                \vx-\vmu_\ell^*,
                \vmu_i-\vmu_\ell^*
            \right\rangle
        \right|
        +
        C\norm{\vmu_i-\vmu_\ell^*}^2
    \right),\\
    |\tldtau_k(\vx)|
    &\le
    C\left(
        \left|
            \left\langle
                \vx-\vmu_\rho^*,
                \vmu_k-\vmu_\rho^*
            \right\rangle
        \right|^2
        +
        \norm{\vmu_k-\vmu_\rho^*}^2
    \right)
    \exp\left(
        C\left|
            \left\langle
                \vx-\vmu_\rho^*,
                \vmu_k-\vmu_\rho^*
            \right\rangle
        \right|
        +
        C\norm{\vmu_k-\vmu_\rho^*}^2
    \right).
    \end{aligned}
    \]

    Expanding the products in these Taylor bounds reduces each integrand to a finite sum of terms covered by
    \eqref{eq:cell-gaussian-ratio-product-offdiag}.  Indeed, every polynomial
    factor is a fixed product of linear factors centered at either
    \(\vmu_\ell^*\) or \(\vmu_s^*\), and every exponential linear factor is
    also centered at either \(\vmu_\ell^*\) or \(\vmu_s^*\).  The constant
    terms in the Taylor bounds already carry the corresponding squared
    displacement factor.  Thus each occurrence of \(\tau_i\) contributes a
    factor \(\norm{\vmu_i-\vmu_\ell^*}^2\), and each occurrence of
    \(\tldtau_k\) contributes a factor
    \(\norm{\vmu_k-\vmu_\rho^*}^2\), after integration.
    Taking
    \(H=\tau_i\),
    \(H=\tau_i\left\langle \vx-\vmu_r^*,\vv\right\rangle\), and
    \(H=\tau_i\tldtau_k\), respectively, proves
    \eqref{eq:separated-overlap-tau},
    \eqref{eq:separated-overlap-linear}, and
    \eqref{eq:separated-overlap-second}, with the exponential factor
    \(e^{-c(\Delta-CB)^2}\).

    For the true-center estimates, we use the same way.  For any measurable factor \(H\),
    \[
    \begin{aligned}
        &\E_{p^*}
        \left[
        \frac{\phi(\vmu_s^*)}{p^*}
        \frac{\phi(\vmu_\ell^*)}{p^*}
        |H|
        \right]
        \le
        \frac{1}{\pimins}
        \sum_{a=1}^m
        \int_{\Omega_a}
        \phi(\vmu_\ell^*;\vx)
        \frac{\phi(\vmu_s^*;\vx)}
             {\phi(\vmu_a^*;\vx)}
        |H(\vx)|\rd\vx .
    \end{aligned}
    \]
    For each cell \(\Omega_a\), apply
    \eqref{eq:cell-gaussian-ratio-product-offdiag} with ambient center
    \(\vnu=\vmu_\ell^*\), ambient true group \(\ell\), ratio numerator \(s\),
    cell index \(a\), and \(q=1\).  Again the diagonal case is impossible
    because \(s\ne\ell\).  Hence every cell carries the factor
    \(e^{-c(\Delta-CB)^2}\), after absorbing the finite cell sum and
    \(1/\pimins\) into \(\sfA_{\rm loc}^{C}\).

    The linear and bilinear factors are covered by the product-form polynomial
    part of \eqref{eq:cell-gaussian-ratio-product-offdiag}, giving the factors
    \(\norm{\vv}\) and \(\norm{\vw}\).  The Taylor bound for \(\tldtau\) gives
    one factor \(\norm{\vmu_k-\vmu_\rho^*}^2\) for each occurrence of
    \(\tldtau_k\), exactly as above.

    Taking \(H\) equal to
    $\left\langle \vx-\vmu_r^*,\vv\right\rangle$,
    $\left\langle \vx-\vmu_r^*,\vv\right\rangle
    \left\langle \vx-\vmu_t^*,\vw\right\rangle$, $\tldtau_k,$
    and then to
    $\left\langle \vx-\vmu_r^*,\vv\right\rangle\tldtau_k$,
    $\tldtau_k\tldtau_h$,
    proves
    \eqref{eq:separated-overlap-true-linear}--\eqref{eq:separated-overlap-true-second}.
\end{proof}

\begin{lemma}[Homotopy \(\chi^2\) control]
\label{lem: local chi square}
    Let \(R:=p-p^*\) and \(p_\theta:=\theta p+(1-\theta)p^*\).  For every
    \(\theta\in(0,1/2]\),
    \begin{equation}
        \E_{p_\theta}
        \left[
            \frac{R^2}{p_\theta^2}
        \right]
        \lesssim \frac{\L}{\theta},
        \qquad
        \left|\frac{R}{p_\theta}\right|
        \le \frac2\theta .
        \label{eq:homotopy-chi-square}
    \end{equation}
\end{lemma}

\begin{proof}
    The first estimate is known; see, for example, Proposition 1(c) in
    \cite{nishiyama2020relations}.  We give a self-contained proof below.
    We first record the scalar inequality
    \[
        z-1-\log z
        \ge
        c\theta
        \frac{(z-1)^2}{(1-\theta)+\theta z},
        \qquad
        z>0,\quad \theta\in(0,1/2],
    \]
    for a universal constant \(c>0\).  Indeed, if \(z\ge1\), then
    \(z-1-\log z\gtrsim (z-1)^2/z\), while
    $\frac{\theta}{(1-\theta)+\theta z}
        \le
        \frac1z$.
    Hence
    $\theta
        \frac{(z-1)^2}{(1-\theta)+\theta z}
        \le
        \frac{(z-1)^2}{z}
        \lesssim
        z-1-\log z$.
    If \(0<z\le1\), then
    \(z-1-\log z\gtrsim (z-1)^2\), while
    $\frac{\theta}{(1-\theta)+\theta z}
        \le
        \frac{\theta}{1-\theta}
        \le
        1$.
    Hence the same scalar inequality follows.

    Applying this inequality with \(z=p/p^*\), and using
    \(\int(p-p^*)=0\), gives
    \[
    \begin{aligned}
        \L
        &=
        \int p^*\log\frac{p^*}{p}\rd\vx
        =
        \int p^*
        \left(
            \frac{p}{p^*}
            -
            1
            -
            \log\frac{p}{p^*}
        \right)
        \rd\vx
        \gtrsim
        \theta
        \int
        p^*
        \frac{
            \left(\frac{p}{p^*}-1\right)^2
        }{
            (1-\theta)+\theta\frac{p}{p^*}
        }
        \rd\vx
        =
        \theta
        \int
        \frac{(p-p^*)^2}{p_\theta}
        \rd\vx\\
        &=
        \theta
        \E_{p_\theta}
        \left[
            \frac{R^2}{p_\theta^2}
        \right].
    \end{aligned}
    \]
    This proves the first estimate.

    The pointwise bound follows from
    \(p_\theta\ge\theta p\), \(p_\theta\ge(1-\theta)p^*\) and \(\theta\le1/2\):
    \[
        \left|\frac{R}{p_\theta}\right|
        \le
        \frac{p}{p_\theta}
        +
        \frac{p^*}{p_\theta}
        \le
        \frac1\theta
        +
        \frac1{1-\theta}
        \le
        \frac2\theta.
    \]
\end{proof}

\section{Global convergence}
\label{appendix: global}

The goal of this section is to prove that the population iterates reach the
local threshold \(\eps_0\) from random initialization.

Throughout this section, \(S_\ell\) denotes the full Voronoi group of model
components assigned to \(\vmu_\ell^*\), and
\[
    \hat\pi_\ell:=\sum_{i\in S_\ell}\pi_i,
    \qquad
    U(\vmu):=
    \sum_{\ell=1}^m\sum_{i\in S_\ell}
    \norm{\vmu_i-\vmu_\ell^*}^2 .
\]
All these quantities are evaluated at the current value of \(\vmu\).  The
groups are used only for analysis and are recomputed at the current mean vector.
All estimates involving separated true centers are imposed only when \(m\ge2\);
when \(m=1\), the corresponding cross-cluster errors are zero.

\subsection{Initialization}
\leminit*

\begin{proof}
    We write the initialization equivalently as first drawing
    \(J_i\in[m]\) with probabilities \((\pi_1^*,\ldots,\pi_m^*)\), and then
    setting
    \[
        \vmu_i^{(0)}=\vmu_{J_i}^*+\vZ_i,
        \qquad
        \vZ_i\sim N(0,I_d).
    \]

    \paragraph{Step 1.}
    The proof is by standard coupon collector argument and Gaussian concentration. Choose a universal constant \(C\) large enough so that, for
    \(\vZ\sim N(0,I_d)\), we have $p:=\Pr\{\norm{\vZ}^2\le Cd\}\ge c$
    for a universal constant \(c>0\). For a fixed \(\ell\), the probability that no initializer is both drawn from cluster \(\ell\) and
    within distance \(\sqrt{Cd}\) of \(\vmu_\ell^*\) is at most
    \[
        (1-\pi_\ell^*p)^n
        \le \exp(-c\pimins n).
    \]
    Hence, by a union bound and the lower bound on \(n\), with probability at
    least \(1-\delta/2\), for every \(\ell\in[m]\) there is an index \(i_\ell\)
    such that \(J_{i_\ell}=\ell\) and
    \[
        \norm{\vmu_{i_\ell}^{(0)}-\vmu_\ell^*}^2
        =
        \norm{\vZ_{i_\ell}}^2
        \le Cd.
    \]
    This immediately implies the result by using $\Delta\gtrsim \sqrt{d}$.

    \paragraph{Step 2.}
    We next control the initial trajectory potential.  With probability at
    least \(1-\delta/2\),
    \[
        U(\vmu^{(0)})
        =
        \sum_{i=1}^n \min_{\ell\in[m]}
        \norm{\vmu_i^{(0)}-\vmu_\ell^*}^2
        \le
        \sum_{i=1}^n\norm{\vZ_i}^2
        \lesssim nd+\log(1/\delta)
        \lesssim nd.
    \]
    Here we used the standard chi-square concentration for
    \(\sum_{i=1}^n\norm{\vZ_i}^2\sim\chi^2_{nd}\), together with
    \(\delta>e^{-\sqrt d}\).

    \paragraph{Step 3.}
    Finally, we bound the initial loss.  On the event from Step 1, use the representatives \(i_\ell\). Since
    \(\pi_i^{(0)}=1/n\), for \(\vx\sim N(\vmu_\ell^*,I_d)\),
    $p^{(0)}(\vx)
        \ge
        \frac1n\phi(\vmu_{i_\ell}^{(0)};\vx)$ and $p^*(\vx)\le (2\pi)^{-d/2}$.
    Therefore,
    \[
        \log\frac{p^*(\vx)}{p^{(0)}(\vx)}
        \le
        \log n+\frac12\norm{\vx-\vmu_{i_\ell}^{(0)}}^2 .
    \]
    Taking expectation over \(\vx\sim N(\vmu_\ell^*,I_d)\) and summing over
    \(\ell\) yields
    \[
    \begin{aligned}
        \L^{(0)}
        &\le
        \sum_{\ell=1}^m
        \pi_\ell^*
        \left(
            \log n
            +
            \frac12
            \E_{x\sim N(\vmu_\ell^*,I_d)}
            \norm{x-\vmu_{i_\ell}^{(0)}}^2
        \right)
        =
        \log n
        +
        \frac12
        \sum_{\ell=1}^m
        \pi_\ell^*
        \left(
            d+
            \norm{\vmu_{i_\ell}^{(0)}-\vmu_\ell^*}^2
        \right)
        \lesssim
        \log n+d .
    \end{aligned}
    \]
    Combining the two high-probability events proves the lemma.
\end{proof}

\subsection{Main global-phase theorem}

\begin{theorem}[\cref{thm: global main text} restated, main result for global phase]
\label{thm: global main}
    Consider \cref{alg} with random initialization event in \cref{lem: init}. Let \(L_0\asymp d+\log n\) and \(B^2\asymp dn\) be chosen large enough so that
    \[
        \L(\vpi^{(0)},\vmu^{(0)})\le L_0,
        \qquad
        U(\vmu^{(0)})\le \frac{B^2}{2}.
    \]
    There are universal constants \(c,C>0\) and polynomial quantities
    \[
        \sfA_{\rm g}:=
        \poly\left(
            d,m,n,\frac1{\pimins},
            1+B
        \right),
        \qquad
        \sfA_{\eta}:=
        C\left(
            1+d+B^2+\log\frac1{\pimins}
            +
            \frac1{\pimins}
        \right)
    \]
    such that the following holds. Let \(\eps_0\) be the local threshold, with \(\eps_0\le c\pimins\). When \(m\ge2\), suppose
    \begin{equation}
        \Delta
        \ge
        C_{\rm sep}
        \left(
            B+\sqrt{\log \sfA_{\rm g}}
            +\sqrt{\frac{L_0+\log(1/\pimins)}{\pimins}}
            + \sqrt{\log \frac{1}{\eps_0}}
        \right).
        \label{eq:global-cell-separation}
    \end{equation}
    When \(m=1\), this condition is omitted.

    Assume that each weight update is loss non-increasing and that, after each
    weight update, the post-weight point
    \((\vpi,\vmu)=(\vpi^{(t+1)},\vmu^{(t)})\) satisfies the dual-gap condition in \cref{appendix: alg}
    \begin{equation}
        G_\pi(\vpi,\vmu)\le \eps_\pi,
        \qquad
        \eps_\pi
        \le
        c\,\sfA_{\rm g}^{-1}\eps_0 .
        \label{eq:global-kkt}
    \end{equation}
    If \(\eta\le c/\sfA_{\eta}\), then for any
    \(\eps\ge \eps_0\), the global phase reaches \(\L\le \eps\) within
    \[
        T_1
        \le
        \frac{C B^2}{\eta\eps}
    \]
    iterations.  Moreover, before the global phase stops,
    \[
        U(\vmu^{(t)})\le \frac{B^2}{2}.
    \]
\end{theorem}

We first state a one-step proposition and then derive the theorem from it.  The
proof of \cref{prop:global-one-step-bootstrap} is deferred to
\cref{sec:global-one-step-proof}.

\begin{proposition}[One global step]
\label{prop:global-one-step-bootstrap}
    Under the assumptions and parameter choices of \cref{thm: global main},
    consider a post-weight point \((\vpi,\vmu)\) satisfying
    \[
        \eps_0\le L:=\L(\vpi,\vmu)\le L_0,
        \qquad
        U(\vmu)\le \frac{B^2}{2},
    \]
    together with \eqref{eq:global-kkt}.  Let
    $\vmu^+:=\vmu-\eta\nabla_{\vmu}\L(\vpi,\vmu)$.
    If \(\eta\le c/\sfA_{\eta}\), then
    \[
        U(\vmu^+)\le U(\vmu),
        \qquad
        \L(\vpi,\vmu^+)
        \le
        \L(\vpi,\vmu)
        -
        \frac{c\eta}{B^2}\L(\vpi,\vmu)^2 .
    \]
\end{proposition}

\begin{proof}[Proof of \cref{thm: global main}]
    Fix a target \(\eps\ge\eps_0\), and let
    \[
        L_t:=\L(\vpi^{(t)},\vmu^{(t)}),
        \qquad
        U_t:=U(\vmu^{(t)}),
        \qquad
        L_{t+0.5}:=
        \L(\vpi^{(t+1)},\vmu^{(t)}).
    \]
    Let \(\tau_\eps\) be the first time during the global phase at which the
    loss is at most \(\eps\).  We prove by induction that, before this stopping
    time,
    \[
        L_t\le L_0,
        \qquad
        U_t\le \frac{B^2}{2}.
    \]
    These bounds hold at \(t=0\) on the initialization event.  Suppose they
    hold at time \(t\), and assume the global phase has not yet stopped, so
    \(L_t>\eps\).  The weight update is loss non-increasing and does not move
    the means, hence \(L_{t+0.5}\le L_t\) and \(U\) is unchanged.  If
    \(L_{t+0.5}\le\eps\), then the target has been reached.  Otherwise,
    $\eps_0\le \eps < L_{t+0.5}\le L_0$,
    and \cref{prop:global-one-step-bootstrap} applies at
    \((\vpi^{(t+1)},\vmu^{(t)})\).  Therefore
    \[
        U_{t+1}\le U_t\le \frac{B^2}{2},
        \qquad
        L_{t+1}
        \le
        L_{t+0.5}
        -
        \frac{c\eta}{B^2}L_{t+0.5}^2.
    \]
    In particular, \(L_{t+1}\le L_t\le L_0\), so the induction holds.

    We next convert the decrease from \(L_{t+0.5}\) to \(L_{t+1}\) into a
    recurrence in \(L_t\).  If \(L_{t+0.5}\ge L_t/2\), then $L_{t+1}
        \le
        L_t-\frac{c\eta}{B^2}L_t^2$
    after decreasing \(c\).  If \(L_{t+0.5}<L_t/2\), then
    \(L_{t+1}\le L_{t+0.5}<L_t/2\), which implies the same bound after
    decreasing \(c\), since \(L_t\le L_0\lesssim B^2\) and
    \(\eta\le1\).  Thus, before the global phase reaches the target,
    \begin{align*}
        L_{t+1}
        \le
        L_t-\frac{c\eta}{B^2}L_t^2 .
    \end{align*}

    It remains to solve this scalar recurrence.  If \(L_{t+1}=0\), the target
    has already been reached.  Otherwise, since \(L_{t+1}\le L_t\),
    \[
        \frac1{L_{t+1}}-\frac1{L_t}
        =
        \frac{L_t-L_{t+1}}{L_tL_{t+1}}
        \ge
        \frac{L_t-L_{t+1}}{L_t^2}
        \ge
        \frac{c\eta}{B^2}.
    \]
    Summing over all steps before the stopping time gives
    \[
        \frac1{L_t}
        \ge
        \frac1{L_0}
        +
        \frac{c\eta t}{B^2}.
    \]
    Hence \(L_t\le\eps\) once
    \(t\ge CB^2/(\eta\eps)\).  Therefore the target loss \(\L\le\eps\) is
    reached within \(CB^2/(\eta\eps)\) iterations.  Taking
    \(\eps=\eps_0\) gives the stated global phase to the local threshold.
\end{proof}

\subsection{Proof of \cref{prop:global-one-step-bootstrap}}
\label{sec:global-one-step-proof}

The one-step proof uses a full-potential projection estimate.  This estimate
is the global analogue of the full-potential descent estimate in the local
section, but its proof does not Taylor-expand all components around their
assigned true centers.

\begin{lemma}[\cref{lem:global-full-potential-linear-main-text} restated]
\label{lem:global-full-potential-linear}
    Under the assumptions and parameter choices of \cref{thm: global main}, consider a post-weight point \((\vpi,\vmu)\) satisfying \eqref{eq:global-kkt}.
    Suppose
    $U(\vmu)\le \frac{B^2}{2}$ and
    $\eps_0\le \L(\vpi,\vmu)\le L_0.$
    Then
    \[
        \sum_{\ell=1}^m\sum_{i\in S_\ell}
        \left\langle
            \nabla_{\vmu_i}\L(\vpi,\vmu),
            \vmu_i-\vmu_\ell^*
        \right\rangle
        \ge
        c\L(\vpi,\vmu).
    \]
\end{lemma}

This immediately gives the proof of \cref{prop:global-one-step-bootstrap}.
\begin{proof}[Proof of \cref{prop:global-one-step-bootstrap}]
    Denote
    \[
        \vg:= \nabla_{\vmu}\L(\vpi,\vmu),
        \qquad
        \mathcal P:=
        \sum_{\ell=1}^m\sum_{i\in S_\ell}
        \left\langle
            \nabla_{\vmu_i}\L(\vpi,\vmu),
            \vmu_i-\vmu_\ell^*
        \right\rangle .
    \]
    By \cref{lem:global-group-weight-control},
    \(\hat\pi_\ell\ge\pi_\ell^*/2\) for every \(\ell\).  Moreover,
    \(U(\vmu)\le B^2/2\) implies
    \(\norm{\vmu_i-\vmu_\ell^*}\le B\) for every \(i\in S_\ell\), and hence
    $\frac1{\hat\pi_\ell}
        \sum_{i\in S_\ell}
        \pi_i\norm{\vmu_i-\vmu_\ell^*}^2
        \le B^2$.
    Therefore \cref{lem:local-smoothness-self-bound} applies with
    \(c_{\rm cov}=1/2\) and with the same \(\sfA_{\eta}\), up to
    universal constants.  In particular, we have
    \[
        \norm{\vg}_F^2\le 2\sfA_{\eta}L,
        \qquad
        \L(\vpi,\vmu-\eta \vg)
        \le
        \L(\vpi,\vmu)
        -
        \frac{\eta}{2}\norm{\vg}_F^2 .
    \]
    We first prove that the potential is non-increasing.  Evaluating the updated centers with the old assignment and using
    \cref{lem:global-full-potential-linear}, we get
    \[
    \begin{aligned}
        U(\vmu^+)
        &\le
        \sum_{\ell=1}^m\sum_{i\in S_\ell}
        \norm{
            \vmu_i-\eta \vg_i-\vmu_\ell^*
        }^2
        =
        U(\vmu)-2\eta\mathcal P+\eta^2\norm{\vg}_F^2
        \le
        U(\vmu)-c\eta L+2\eta^2\sfA_{\eta}L
        \le U(\vmu),
    \end{aligned}
    \]
    where the last step uses \(\eta\le c/\sfA_{\eta}\).

    We next prove the loss decrease.  By Cauchy--Schwarz,
    \cref{lem:global-full-potential-linear}, and \(U(\vmu)\le B^2/2\),
    \[
        \norm{\vg}_F
        \ge
        \frac{\mathcal P}{\sqrt{U(\vmu)}}
        \ge
        \frac{cL}{B}.
    \]
    Combining this bound with the descent estimate above gives
    \[
        \L(\vpi,\vmu^+)
        \le
        \L(\vpi,\vmu)
        -
        \frac{c\eta}{B^2}\L(\vpi,\vmu)^2 .
    \]
\end{proof}

\subsection{Proof of \cref{lem:global-full-potential-linear}}
\label{sec:global-full-projection}

We now prove \cref{lem:global-full-potential-linear}.  The proof has three steps.  First, we reduce the full-potential projection to a sum of same-group terms.  Second, we lower bound each same-group term by a same-group KL divergence, up to a within-group weight-entropy error.  Third, we compare the sum of same-group KL divergences to the global loss.

\begin{proof}[Proof of \cref{lem:global-full-potential-linear}]
    Since
    \(U(\vmu)\le B^2/2\), we have \(\norm{\vmu_i-\vmu_\ell^*}\le B\) for every \(i\in S_\ell\).  By \cref{lem:global-group-weight-control}, we have $\hat\pi_\ell\ge\frac12\pi_\ell^*$ for every $\ell\in[m]$.
    
    Define the normalized within-group mixture and posterior by
    \[
        q_\ell(\vx):=
        \frac1{\hat\pi_\ell}
        \sum_{i\in S_\ell}\pi_i\phi(\vmu_i;\vx),
        \qquad
        \alpha_i^{(\ell)}:=\frac{\pi_i}{\hat\pi_\ell}
        \qquad
        \psi_i^{(\ell)}(\vx)
        :=
        \frac{\alpha_i^{(\ell)}\phi(\vmu_i;\vx)}{q_\ell(\vx)}
        \quad (i\in S_\ell).
    \]
    The associated within-group barycenter is
    \[
        \tldvpsi_\ell(\vx):=
        \sum_{i\in S_\ell}
        \psi_i^{(\ell)}(\vx)(\vmu_i-\vmu_\ell^*) .
    \]

    \paragraph{Step 1: reduction to same-group terms.}
    Let the target term be
    \[
        \mathcal P:=
        \sum_{\ell=1}^m\sum_{i\in S_\ell}
        \left\langle
            \nabla_{\vmu_i}\L,
            \vmu_i-\vmu_\ell^*
        \right\rangle .
    \]
    Using the gradient representation \eqref{eq: gradient form}, we expand
    \[
    \begin{aligned}
        \mathcal P
        &=
        \sum_{\ell=1}^m
        \sum_{i\in S_\ell}
        \sum_{r=1}^m
        \pi_r^*
        \E_{\vx\sim r}
        \left[
            \psi_i(\vx)
            \left\langle
                \vmu_i-\vmu_\ell^*,
                \sum_{s=1}^m\sum_{k\in S_s}
                \psi_k(\vx)(\vmu_k-\vmu_r^*)
            \right\rangle
        \right]                                      \\
        &=
        \sum_{\ell=1}^m
        \pi_\ell^*
        \E_{\vx\sim \ell}
        \left[
            \left\langle
                \sum_{i\in S_\ell}\psi_i(\vx)(\vmu_i-\vmu_\ell^*),
                \sum_{s=1}^m\sum_{k\in S_s}
                \psi_k(\vx)(\vmu_k-\vmu_\ell^*)
            \right\rangle
        \right]                                      \\
        &\quad+
        \sum_{\ell=1}^m
        \sum_{\substack{r=1\\ r\ne \ell}}^m
        \pi_r^*
        \E_{\vx\sim r}
        \left[
            \left\langle
                \sum_{i\in S_\ell}\psi_i(\vx)(\vmu_i-\vmu_\ell^*),
                \sum_{s=1}^m\sum_{k\in S_s}
                \psi_k(\vx)(\vmu_k-\vmu_r^*)
            \right\rangle
        \right] .
    \end{aligned}
    \]
    For the first term, we separate the correct group \(s=\ell\) from the wrong
    groups \(s\ne\ell\). This leads to the exact decomposition
    \begin{align*}
        \mathcal P
        &=
        \sum_{\ell=1}^m
        \pi_\ell^*
        \E_{\vx\sim \ell}
        \norm{\sum_{i\in S_\ell}\psi_i(\vx)(\vmu_i-\vmu_\ell^*)}^2
        +
        \mathcal E_{\rm sep},
    \end{align*}
    where
    \[
    \begin{aligned}
        \mathcal E_{\rm sep}
        &:=
        \sum_{\ell=1}^m
        \pi_\ell^*
        \E_{\vx\sim \ell}
        \left[
            \left\langle
                \sum_{i\in S_\ell}\psi_i(\vx)(\vmu_i-\vmu_\ell^*),
                \sum_{\substack{s=1\\s\ne\ell}}^m
                \sum_{k\in S_s}
                \psi_k(\vx)(\vmu_k-\vmu_\ell^*)
            \right\rangle
        \right]                                      \\
        &\quad+
        \sum_{\ell=1}^m
        \sum_{\substack{r=1\\ r\ne \ell}}^m
        \pi_r^*
        \E_{\vx\sim r}
        \left[
            \left\langle
                \sum_{i\in S_\ell}\psi_i(\vx)(\vmu_i-\vmu_\ell^*),
                \sum_{s=1}^m\sum_{k\in S_s}
                \psi_k(\vx)(\vmu_k-\vmu_r^*)
            \right\rangle
        \right] .
    \end{aligned}
    \]
        All terms in \(\mathcal E_{\rm sep}\) contain either data from a wrong true
    cluster or a posterior factor from a wrong Voronoi group.  We use
    \cref{lem:global-separated-overlap} in the following slightly stronger
    form, which follows from the same proof: whenever \(a\ne b\),
    \[
        \E_{\vx\sim a}
        \sum_{i\in S_b}
        \psi_i(\vx)
        \left(1+\norm{\vmu_i-\vmu_a^*}\right)
        \le
        \sfA_{\rm g}e^{-c\Delta^2}.
    \]
    Indeed, in the proof of \cref{lem:global-separated-overlap}, the pairwise
    contribution from group \(b\) against data group \(a\) carries the sharper
    factor
    \(e^{-c\norm{\vmu_b^*-\vmu_a^*}^2}\).  Since
    $\norm{\vmu_i-\vmu_a^*}
        \le
        \norm{\vmu_b^*-\vmu_a^*}+B$
    for every $i\in S_b$,
    this extra displacement factor is absorbed by the same Gaussian tail,
    without introducing any dependence on \(\Delta_{\max}\).
    
    Hence
    \begin{equation*}
        \mathcal P
        \ge
        \sum_{\ell=1}^m
        \pi_\ell^*
        \E_{\vx\sim \ell}
        \norm{\sum_{i\in S_\ell}\psi_i(\vx)(\vmu_i-\vmu_\ell^*)}^2
        -
        \sfA_{\rm g}e^{-c\Delta^2}.
    \end{equation*}
    It remains to replace the full posterior $\psi_i$ by the
    normalized same-group posterior \(\psi_i^{(\ell)}\).  For \(\vx\sim\ell\), denote
    \[
        \sum_{i\in S_\ell}\psi_i(\vx)(\vmu_i-\vmu_\ell^*)
        =
        \omega_\ell(\vx)\tldvpsi_\ell(\vx),
        \qquad
        \omega_\ell(\vx):=
        \frac{\hat\pi_\ell q_\ell(\vx)}{p(\vx)}.
    \]
    Since \(0\le\omega_\ell\le1\) and
    \(\norm{\tldvpsi_\ell(\vx)}\le B\), we have
    \[
    \begin{aligned}
        \E_{\vx\sim\ell}\norm{\sum_{i\in S_\ell}\psi_i(\vx)(\vmu_i-\vmu_\ell^*)}^2
        &=
        \E_{\vx\sim\ell}
        \left[
            \omega_\ell(\vx)^2
            \norm{\tldvpsi_\ell(\vx)}^2
        \right]\\
        &=
        \E_{\vx\sim\ell}\norm{\tldvpsi_\ell(\vx)}^2
        -
        \E_{\vx\sim\ell}
        \left[
            \left(1-\omega_\ell(\vx)^2\right)
            \norm{\tldvpsi_\ell(\vx)}^2
        \right]                                      \\
        &\ge
        \E_{\vx\sim\ell}\norm{\tldvpsi_\ell(\vx)}^2
        -
        2B^2
        \E_{\vx\sim\ell}
        \left[
            1-\omega_\ell(\vx)
        \right].
    \end{aligned}
    \]
    By \cref{lem:global-separated-overlap},
    $\sum_{\ell=1}^m
        \pi_\ell^*
        \E_{\vx\sim\ell}
        \left[
            1-\omega_\ell(\vx)
        \right]
        \le
        \sfA_{\rm g}e^{-c\Delta^2}$.
    Absorbing the polynomial factor \(B^2\) into \(\sfA_{\rm g}\), we get
    \begin{align*}
        \mathcal P
        \ge
        \sum_{\ell=1}^m
        \pi_\ell^*
        \E_{\vx\sim\ell}\norm{\sum_{i\in S_\ell}\psi_i(\vx)(\vmu_i-\vmu_\ell^*)}^2
        -
        \sfA_{\rm g}e^{-c\Delta^2}
        \ge
        c\sum_{\ell=1}^m
        \pi_\ell^*
        \E_{\vx\sim\ell}\norm{\tldvpsi_\ell(\vx)}^2
        -
        \sfA_{\rm g}e^{-c\Delta^2}.
    \end{align*}

    \paragraph{Step 2: one-cluster entropy production.}
    For each \(\ell\), translate by \(\vmu_\ell^*\) and apply
    \cref{lem:one-cluster-entropy-production} to \(q_\ell\).  If
    \[
        \beta_i^{(\ell)}
        :=
        \E_{\vx\sim\ell}\psi_i^{(\ell)}(\vx),
        \qquad i\in S_\ell,
    \]
    then
    \[
        \E_{\vx\sim\ell}\norm{\tldvpsi_\ell(\vx)}^2
        \ge
        2\kl(\phi(\vmu_\ell^*)\|q_\ell)
        -
        2\kl(\beta^{(\ell)}\|\alpha^{(\ell)}).
    \]
    Summing over \(\ell\) and using
    \cref{lem:within-group-weight-entropy}, we obtain
    \begin{align*}
        \sum_{\ell=1}^m
        \pi_\ell^*
        \E_{\vx\sim\ell}\norm{\tldvpsi_\ell(\vx)}^2
        &\ge
        2\sum_{\ell=1}^m
        \pi_\ell^*
        \kl(\phi(\vmu_\ell^*)\|q_\ell)
        -
        C\left(
            \eps_\pi
            +
            \sfA_{\rm g}e^{-c\Delta^2}
        \right).
    \end{align*}

    \paragraph{Step 3: comparison to the global loss.}
    By \cref{lem:global-loss-comparison},
    \begin{equation*}
        \L(\vpi,\vmu)
        \le
        C\sum_{\ell=1}^m
        \pi_\ell^*
        \kl(\phi(\vmu_\ell^*)\|q_\ell)
        +
        C\left(
            \eps_\pi
            +
            \sfA_{\rm g}e^{-c\Delta^2}
        \right).
    \end{equation*}
    Combining above and Step 1,2 gives
    \[
        \mathcal P
        \ge
        c\L(\vpi,\vmu)
        -
        C\left(
            \eps_\pi
            +
            \sfA_{\rm g}e^{-c\Delta^2}
        \right).
    \]
    The error term is at most \(c\L(\vpi,\vmu)/2\), because
    \(\eps_\pi\le c\sfA_{\rm g}^{-1}\eps_0\),
    \(\sfA_{\rm g}e^{-c\Delta^2}\le c\eps_0\), and
    \(\L(\vpi,\vmu)\ge\eps_0\).  This proves the lemma.
\end{proof}

\subsection{Technical lemmas for the global phase}
\label{sec:global-technical-lemmas}
We collect the technical estimates used in the proof of
\cref{lem:global-full-potential-linear}.  The main input is the shared cell
Gaussian ratio estimate \cref{lem:cell-gaussian-ratio-moment}, together with
a cross-group overlap consequence for posterior denominators $\psi_i$. The
only trajectory input is the bound
\(\norm{\vmu_i-\vmu_\ell^*}\le B\) for \(i\in S_\ell\), which follows from
\(U(\vmu)\le B^2/2\).

\begin{lemma}[Global cross-group estimates]
\label{lem:global-separated-overlap}
    Under the assumptions and parameter choices of \cref{thm: global main},
    suppose \(U(\vmu)\le B^2/2\) and \(\L(\vpi,\vmu)\le L_0\).  Then the
    following estimates hold.

    \begin{enumerate}
        \item Every group has a loose positive mass lower bound:
        \begin{equation}
            \hpi_\ell
            \ge
            \pi_\ell^*
            \exp\left(
                -C\frac{L_0+\log(1/\pimins)}{\pimins}
            \right)
            \qquad\text{for every }\ell\in[m].
            \label{eq:global-loose-group-mass}
        \end{equation}
        In particular, \(q_\ell\) below is well-defined, where
        \[
            q_\ell(x):=
            \frac1{\hat\pi_\ell}
            \sum_{i\in S_\ell}\pi_i\phi(\vmu_i;x).
        \]

        \item For \(i\in S_\ell\) and \(j\ne\ell\), the cross-posterior term is
        bounded as
        \begin{equation}
            \E_{\vx\sim j}\psi_i(\vx)
            \le
            \sfA_{\rm g} e^{-c\Delta^2}.
            \label{eq:global-posterior-cross}
        \end{equation}
        Consequently, for every \(\ell\in[m]\),
        \begin{equation}
            \E_{\vx\sim\ell}
            \sum_{i\notin S_\ell}\psi_i(\vx)
            +
            \sum_{j\ne\ell}
            \E_{\vx\sim j}
            \sum_{i\in S_\ell}\psi_i(\vx)
            \le
            \sfA_{\rm g} e^{-c\Delta^2}.
            \label{eq:global-posterior-group-cross}
        \end{equation}

        \item Denote
        \[
            p_{-\ell}(x):=
            \sum_{r\ne\ell}\sum_{i\in S_r}\pi_i\phi(\vmu_i;x),
            \qquad
            \rho_\ell(x):=\frac{p_{-\ell}(x)}{\hat\pi_\ell q_\ell(x)},
            \qquad
            \omega_\ell(x):=\frac1{1+\rho_\ell(x)}.
        \]
        Then
        \begin{equation}
            \E_{\vx\sim\ell}\log(1+\rho_\ell(\vx))
            +
            \E_{\vx\sim\ell}\bigl[1-\omega_\ell(\vx)\bigr]
            \le
            \sfA_{\rm g} e^{-c\Delta^2}.
            \label{eq:global-model-cross-ratio}
        \end{equation}

        \item The true-mixture analogue of point 3 also holds:
        \[
            \rho_\ell^*(x):=
            \frac{\sum_{r\ne\ell}\pi_r^*\phi(\vmu_r^*;x)}
            {\pi_\ell^*\phi(\vmu_\ell^*;x)}
        \]
        satisfies
        \begin{equation}
            \E_{\vx\sim\ell}\log(1+\rho_\ell^*(\vx))
            \le
            \sfA_{\rm g} e^{-c\Delta^2}.
            \label{eq:global-true-cross-ratio}
        \end{equation}
    \end{enumerate}
\end{lemma}

\begin{proof}
    If \(m=1\), all cross-cluster terms vanish.  Hence assume \(m\ge2\).
    Since \(U(\vmu)\le B^2/2\), every \(i\in S_\ell\) satisfies $\norm{\vmu_i-\vmu_\ell^*}\le B$.
    We use \cref{lem:cell-gaussian-ratio-moment}, with
    \(\sfA_{\rm cell}^{C}\) replaced by \(\sfA_{\rm g}^{C}\).
    By \eqref{eq:global-cell-separation},  every raw bound of the form
    \[
        \sfA_{\rm g}^{C}
        \exp\left(
            C\frac{L_0+\log(1/\pimins)}{\pimins}
        \right)
        e^{-c(\Delta-CB)^2}
    \]
    is bounded by \(\sfA_{\rm g} e^{-c\Delta^2}\).

    We also use the following elementary Gaussian overlap bound.  Let
    \(a\ne b\), \(u\in S_a\), \(v\in S_b\), and \(A\ge0\) satisfy
    \[
        \log(1+A)
        \le
        C\left(
            \log\sfA_{\rm g}
            +
            \frac{L_0+\log(1/\pimins)}{\pimins}
        \right).
    \]
    Then
    \begin{equation}
        \E_{\vx\sim a}
        \sqrt{
            A\frac{\phi(\vmu_v;\vx)}{\phi(\vmu_u;\vx)}
        }
        \le
        \sfA_{\rm g} e^{-c\Delta^2}.
        \label{eq:global-hellinger-overlap}
    \end{equation}
    Indeed, write $\vmu_u=\vmu_a^*+\vh_u$,
    $\vmu_v=\vmu_b^*+\vh_v$, and
    $\norm{\vh_u},\norm{\vh_v}\le B$,
    and set \(\vx=\vmu_a^*+\vZ\).  Then, using the Gaussian ratio identity,
    \[
    \begin{aligned}
        \E_{\vx\sim a}
        \sqrt{
            A\frac{\phi(\vmu_v;\vx)}{\phi(\vmu_u;\vx)}
        }
        &=
        A^{1/2}
        \E_{\vZ\sim N(0,I)}
        \exp\left(
            \frac14\norm{\vZ-\vh_u}^2
            -
            \frac14\norm{\vZ-(\vmu_b^*-\vmu_a^*)-\vh_v}^2
        \right)                                                   \\
        &\quad\le
        \exp\left(
            C\log\sfA_{\rm g}
            +
            C\frac{L_0+\log(1/\pimins)}{\pimins}
        \right)
        \exp\left(-c(\Delta-CB)^2\right)
        \le
        \sfA_{\rm g} e^{-c\Delta^2},
    \end{aligned}
    \]
    where the last step uses \eqref{eq:global-cell-separation}.  The same
    bound holds when \(\vmu_u,\vmu_v\) are replaced by the corresponding true
    centers.

    \paragraph{Loose group mass.}
    Jensen's inequality gives
    \begin{equation}
    \begin{aligned}
        \L(\vpi,\vmu)
        =
        \E_{p^*}\log\frac{p^*}{p}
        \ge
        -\sum_{r=1}^m
        \pi_r^*
        \log
        \E_{\vx\sim r}\frac{p(\vx)}{p^*(\vx)} .
    \end{aligned}
    \label{eq:global-mass-jensen}
    \end{equation}
    For \(i\in S_\ell\), using
    \(p^*(\vx)\ge\pi_\ell^*\phi(\vmu_\ell^*;\vx)\), we get
    \[
        \E_{\vx\sim\ell}
        \frac{\phi(\vmu_i;\vx)}{p^*(\vx)}
        \le
        \frac1{\pi_\ell^*}.
    \]
    For \(i\in S_r\) with \(r\ne\ell\), decompose over true cells and use
    \(p^*(\vx)\ge\pi_a^*\phi(\vmu_a^*;\vx)\) on \(\Omega_a\):
    \[
    \begin{aligned}
        \E_{\vx\sim\ell}
        \frac{\phi(\vmu_i;\vx)}{p^*(\vx)}
        &\le
        \frac1{\pimins}
        \sum_{a=1}^m
        \int_{\Omega_a}
        \phi(\vmu_\ell^*;\vx)
        \frac{\phi(\vmu_i;\vx)}
             {\phi(\vmu_a^*;\vx)}
        \rd\vx                                                   \\
        &=
        \frac1{\pimins}
        \sum_{a=1}^m
        \int_{\Omega_a}
        \phi(\vmu_\ell^*;\vx)
        \frac{\phi(\vmu_r^*;\vx)}
             {\phi(\vmu_a^*;\vx)}
        \exp\left(
            \ip{\vx-\vmu_r^*,\vmu_i-\vmu_r^*}
            -
            \frac12\norm{\vmu_i-\vmu_r^*}^2
        \right)
        \rd\vx                                                   \\
        &\le
        \sfA_{\rm g} e^{-c\Delta^2}.
    \end{aligned}
    \]
    In the last step we applied
    \eqref{eq:cell-gaussian-ratio-product-offdiag} with ambient true group
    \(\ell\), ratio numerator \(r\), cell index \(a\), and \(q=1\).  Since
    \(r\ne\ell\), every cell contribution is off-diagonal.

    Therefore
    \begin{equation}
        \E_{\vx\sim\ell}\frac{p(\vx)}{p^*(\vx)}
        =\sum_{i\in[n]}\E_{\vx\sim\ell}\frac{\pi_i\phi(\vmu_i;\vx)}{p^*(\vx)}
        \le
        \frac{\hat\pi_\ell}{\pi_\ell^*}
        +
        \sfA_{\rm g}e^{-c\Delta^2}.
        \label{eq:global-group-mass-upper}
    \end{equation}
    Also, the same bound and \(\hat\pi_r\le1\) imply
    $\E_{\vx\sim r}\frac{p(\vx)}{p^*(\vx)}
        \le
        \frac{C}{\pimins}$
     for every $r$.
    Isolating the \(\ell\)-th term in
    \eqref{eq:global-mass-jensen}, and using
    \(\L(\vpi,\vmu)\le L_0\), gives
    \[
    \begin{aligned}
        \E_{\vx\sim\ell}\frac{p(\vx)}{p^*(\vx)}
        &\ge
        \exp\left(
            -C\frac{L_0+\log(1/\pimins)}{\pimins}
        \right).
    \end{aligned}
    \]
    Combining this lower bound with
    \eqref{eq:global-group-mass-upper}, we obtain
    \[
        \hat\pi_\ell
        \ge
        \pi_\ell^*
        \exp\left(
            -C\frac{L_0+\log(1/\pimins)}{\pimins}
        \right)
        -
        \pi_\ell^*\sfA_{\rm g}e^{-c\Delta^2}.
    \]
    By \eqref{eq:global-cell-separation}, after increasing the constant \(C\)
    in the exponential, the second term is at most half of the first.  This
    proves \eqref{eq:global-loose-group-mass}.

    \paragraph{Posterior cross terms.}
    Fix \(i\in S_\ell\) and \(j\ne\ell\).  By
    \eqref{eq:global-loose-group-mass}, choose \(k\in S_j\) such that
    $\pi_k\ge \frac{\hat\pi_j}{n}>0$.
    If \(\pi_i=0\), then \(\psi_i\equiv0\).  Otherwise,
    using \(p\ge \pi_i\phi(\vmu_i)+\pi_k\phi(\vmu_k)\), we have
    \[
    \begin{aligned}
        \psi_i(\vx)
        &=
        \frac{\pi_i\phi(\vmu_i;\vx)}{p(\vx)}
        \le
        \frac{
            \frac{\pi_i}{\pi_k}
            \frac{\phi(\vmu_i;\vx)}{\phi(\vmu_k;\vx)}
        }{
            1+
            \frac{\pi_i}{\pi_k}
            \frac{\phi(\vmu_i;\vx)}{\phi(\vmu_k;\vx)}
        }
        \le
        \sqrt{
            \frac{\pi_i}{\pi_k}
            \frac{\phi(\vmu_i;\vx)}{\phi(\vmu_k;\vx)}
        } .
    \end{aligned}
    \]
    Moreover, by \eqref{eq:global-loose-group-mass},
    \[
        \log\left(1+\frac{\pi_i}{\pi_k}\right)
        \le
        C\left(
            \log\sfA_{\rm g}
            +
            \frac{L_0+\log(1/\pimins)}{\pimins}
        \right).
    \]
    Applying \eqref{eq:global-hellinger-overlap} with
    \(a=j\), \(b=\ell\), \(u=k\), and \(v=i\), gives
    \[
        \E_{\vx\sim j}\psi_i(\vx)
        \le
        \sfA_{\rm g}e^{-c\Delta^2}.
    \]
    This proves \eqref{eq:global-posterior-cross}.  Summing over \(i\) and
    \(j\), and absorbing polynomial factors into \(\sfA_{\rm g}\), gives
    \eqref{eq:global-posterior-group-cross}.

    \paragraph{Model cross ratio.}
    Fix \(\ell\).  Again by \eqref{eq:global-loose-group-mass}, choose
    \(k\in S_\ell\) such that
    $\pi_k\ge \frac{\hat\pi_\ell}{n}>0$.
    Since
    $\hat\pi_\ell q_\ell(\vx)
        =
        \sum_{h\in S_\ell}\pi_h\phi(\vmu_h;\vx)
        \ge
        \pi_k\phi(\vmu_k;\vx),$
    we have
    \[
        \rho_\ell(\vx)
        =
        \frac{p_{-\ell}(\vx)}
             {\hat\pi_\ell q_\ell(\vx)}
        \le
        \sum_{i\notin S_\ell}
        \frac{\pi_i\phi(\vmu_i;\vx)}
             {\pi_k\phi(\vmu_k;\vx)} .
    \]
    For \(y_i(\vx):=\pi_i\phi(\vmu_i;\vx)/(\pi_k\phi(\vmu_k;\vx))\), the
    elementary inequalities
    \[
        \log\left(1+\sum_i y_i\right)
        \le
        \sum_i\log(1+y_i)
        \le
        C\sum_i\sqrt{y_i},
        \qquad
        \frac{\sum_i y_i}{1+\sum_i y_i}
        \le
        \sum_i\frac{y_i}{1+y_i}
        \le
        \sum_i\sqrt{y_i}
    \]
    give
    \[
    \begin{aligned}
        &\E_{\vx\sim\ell}\log(1+\rho_\ell(\vx))
        +
        \E_{\vx\sim\ell}[1-\omega_\ell(\vx)]              \le
        C\sum_{r\ne\ell}\sum_{i\in S_r}
        \E_{\vx\sim\ell}
        \sqrt{
            \frac{\pi_i}{\pi_k}
            \frac{\phi(\vmu_i;\vx)}
                 {\phi(\vmu_k;\vx)}
        } .
    \end{aligned}
    \]
    For each \(i\in S_r\), \(r\ne\ell\), the coefficient satisfies
    \[
        \log\left(1+\frac{\pi_i}{\pi_k}\right)
        \le
        C\left(
            \log\sfA_{\rm g}
            +
            \frac{L_0+\log(1/\pimins)}{\pimins}
        \right).
    \]
    Applying \eqref{eq:global-hellinger-overlap} with
    \(a=\ell\), \(b=r\), \(u=k\), and \(v=i\), then summing over
    \(i\notin S_\ell\), proves
    \eqref{eq:global-model-cross-ratio}.

    \paragraph{True-mixture cross ratio.}
    For \(\vx\sim N(\vmu_\ell^*,I)\), write
    $\rho_\ell^*(\vx)
        =
        \sum_{r\ne\ell}
        \frac{\pi_r^*}{\pi_\ell^*}
        \frac{\phi(\vmu_r^*;\vx)}
             {\phi(\vmu_\ell^*;\vx)} .$
    Since \(\log(1+\sum_r y_r)\le C\sum_r\sqrt{y_r}\) for \(y_r\ge0\),
    \[
    \begin{aligned}
        \E_{\vx\sim\ell}\log(1+\rho_\ell^*(\vx))
        &\le
        C\sum_{r\ne\ell}
        \E_{\vx\sim\ell}
        \sqrt{
            \frac{\pi_r^*}{\pi_\ell^*}
            \frac{\phi(\vmu_r^*;\vx)}
                 {\phi(\vmu_\ell^*;\vx)}
        }
        \le
        \sfA_{\rm g}e^{-c\Delta^2},
    \end{aligned}
    \]
    where the last step is the true-center version of
    \eqref{eq:global-hellinger-overlap}, and the finite sum over \(r\) is
    absorbed into \(\sfA_{\rm g}\).  This proves
    \eqref{eq:global-true-cross-ratio}.
\end{proof}

\begin{lemma}[Group-weight control]
\label{lem:global-group-weight-control}
    Suppose \(U(\vmu)\le B^2/2\), \(\L(\vpi,\vmu)\le L_0\), and
    \eqref{eq:global-kkt} holds.  Then
    \[
        \sum_{\ell=1}^m
        \abs{\hat\pi_\ell-\pi_\ell^*}
        \le
        C\eps_\pi
        +
        \sfA_{\rm g}e^{-c\Delta^2}.
    \]
    In particular, under 
    \eqref{eq:global-cell-separation}, and the threshold smallness
    \(\eps_0\le c\pimins\),
    \[
        \hat\pi_\ell\ge \frac12\pi_\ell^*
        \qquad\text{for every }\ell\in[m].
    \]
\end{lemma}
\begin{proof}
    By \cref{lem:weight-dual-gap-consequences}, \eqref{eq:global-kkt} gives
    $\sum_i
        \abs{\pi_i-\E_{p^*}\psi_i}
        \le
        2\eps_\pi$.
    For \(\Psi_\ell:=\sum_{i\in S_\ell}\psi_i\), this implies
    \[
        \hat\pi_\ell
        =
        \E_{p^*}\Psi_\ell+r_\ell,
        \qquad
        \sum_{\ell=1}^m\abs{r_\ell}\le 2\eps_\pi.
    \]
    By \eqref{eq:global-posterior-group-cross},
    \[
    \begin{aligned}
        \E_{p^*}\Psi_\ell
        &=
        \pi_\ell^*
        \E_{\vx\sim\ell}\Psi_\ell
        +
        \sum_{j\ne\ell}
        \pi_j^*
        \E_{\vx\sim j}\Psi_\ell                           =
        \pi_\ell^*
        \left(
            1-\E_{\vx\sim\ell}
            \sum_{i\notin S_\ell}\psi_i(\vx)
        \right)
        +
        O\left(\sfA_{\rm g}e^{-c\Delta^2}\right).
    \end{aligned}
    \]
    Summing the resulting errors over \(\ell\) and using
    \eqref{eq:global-posterior-group-cross} again gives
    \[
        \sum_{\ell=1}^m
        \abs{\E_{p^*}\Psi_\ell-\pi_\ell^*}
        \le
        \sfA_{\rm g}e^{-c\Delta^2}.
    \]
    Combining above proves the first
    claim. The lower bound \(\hat\pi_\ell\ge\pi_\ell^*/2\) follows from
    $\Delta$ is large and $\eps_{kkt}$ is small.
\end{proof}

\begin{lemma}
\label{lem:one-cluster-entropy-production}
    Let
    \[
        q(\vx):=\sum_{i=1}^N\alpha_i\phi(\vh_i;\vx),
        \qquad
        \sum_i\alpha_i=1,
        \qquad
        \alpha_i\ge0,
    \]
    and let \(\gamma=\phi(\vzero;\cdot)\).  Define
    \[
        \psi_i(\vx):=\frac{\alpha_i\phi(\vh_i;\vx)}{q(\vx)},
        \qquad
        \bar \vh(\vx):=\sum_i\psi_i(\vx)\vh_i,
        \qquad
        \beta_i:=\E_\gamma\psi_i .
    \]
    Then
    \[
        \E_\gamma\norm{\bar \vh(\vx)}^2
        \ge
        2\kl(\gamma\|q)
        -
        2\kl(\beta\|\alpha),
    \]
    where indices with \(\alpha_i=0\) or \(\beta_i=0\) are omitted.
\end{lemma}

\begin{proof}
We omit indices with \(\alpha_i=0\).  If \(\beta_i=0\), then \(\psi_i=0\) \(\gamma\)-almost surely, so that index contributes nothing
and may also be omitted.

The proof compares two ways of measuring the same posterior fluctuation:
the barycenter \(\bar \vh\) and the posterior variance
\(V(\vx):=\sum_i\psi_i(\vx)\norm{\vh_i-\bar \vh(\vx)}^2\).  Since
\[
    \frac{q(\vx)}{\gamma(\vx)}
    =
    \sum_i\alpha_i
    \exp\left(
        \ip{\vh_i,\vx}-\frac12\norm{\vh_i}^2
    \right),
\]
differentiating gives
\[
    \nabla\log\frac{q}{\gamma}(\vx)=\bar \vh(\vx),
    \qquad
    \nabla\psi_i(\vx)=\psi_i(\vx)(\vh_i-\bar \vh(\vx)).
\]
Therefore the trace of the Jacobian of \(\bar \vh\) is
\[
    \operatorname{Tr}(\nabla\bar \vh(\vx))
    =
    \sum_i\ip{\nabla\psi_i(\vx),\vh_i}
    =
    \sum_i\psi_i(\vx)\ip{\vh_i-\bar \vh(\vx),\vh_i}
    =
    V(\vx),
\]
where the last equality uses
\(\sum_i\psi_i(\vh_i-\bar \vh)=0\).

We next derive a basic identity.  By the definition of \(\psi_i\), for
every index with \(\psi_i(\vx)>0\),
\[
    \psi_i(\vx)
    =
    \frac{\alpha_i\phi(\vh_i;\vx)}{q(\vx)}
    =
    \frac{\alpha_i\gamma(\vx)}{q(\vx)}
    \exp\left(
        \ip{\vh_i,\vx}-\frac12\norm{\vh_i}^2
    \right).
\]
Rearranging and taking logarithms gives
\[
    \log\frac{q(\vx)}{\gamma(\vx)}
    =
    \log\frac{\alpha_i}{\psi_i(\vx)}
    +
    \ip{\vh_i,\vx}
    -
    \frac12\norm{\vh_i}^2 .
\]
Averaging this identity with weights \(\psi_i(\vx)\), whose sum is one,
yields the posterior identity
\[
    \log\frac{q(\vx)}{\gamma(\vx)}
    =
    \sum_i\psi_i(\vx)
    \left[
        \log\frac{\alpha_i}{\psi_i(\vx)}
        +
        \ip{\vh_i,\vx}
        -
        \frac12\norm{\vh_i}^2
    \right].
\]
Hence, using Stein's identity
$\E_\gamma\ip{\vx,\sum_i \psi_i(\vx) \vh_i}
    =
    \E_\gamma\ip{\vx,\bar \vh(\vx)}
    =
    \E_\gamma\operatorname{Tr}(\nabla\bar \vh(\vx))
    =
    \E_\gamma V(\vx)$,
and the pointwise variance decomposition
\(\sum_i\psi_i\norm{\vh_i}^2=\norm{\bar \vh}^2+V\), we obtain
\begin{equation}
    \kl(\gamma\|q)
    =
    \frac12\E_\gamma\norm{\bar \vh}^2
    +
    \E_\gamma\sum_i\psi_i\log\frac{\psi_i}{\alpha_i}
    -
    \frac12\E_\gamma V .
    \label{eq:one-cluster-kl-identity}
\end{equation}

It remains to control the middle term in
\eqref{eq:one-cluster-kl-identity}.  Gaussian log-Sobolev applied to
\(\psi_i\) gives
\[
    \E_\gamma\psi_i\log\frac{\psi_i}{\beta_i}
    \le
    2\E_\gamma\norm{\nabla\sqrt{\psi_i}}^2
    =
    \frac12\E_\gamma\psi_i\norm{\vh_i-\bar \vh}^2.
\]
Therefore
\[
    \E_\gamma\sum_i\psi_i\log\frac{\psi_i}{\alpha_i}
    =
    \E_\gamma\sum_i\psi_i\log\frac{\psi_i}{\beta_i}
    +
    \sum_i\beta_i\log\frac{\beta_i}{\alpha_i}
    \le \frac12\E_\gamma V + \kl(\beta\|\alpha).
\]
Substituting this into \eqref{eq:one-cluster-kl-identity} yields
\[
    \kl(\gamma\|q)
    \le
    \frac12\E_\gamma\norm{\bar \vh}^2
    +
    \kl(\beta\|\alpha).
\]
Rearranging proves the claim.

\end{proof}

\begin{lemma}
\label{lem:within-group-weight-entropy}
    Suppose \(U(\vmu)\le B^2/2\), \(\L(\vpi,\vmu)\le L_0\), and
    \eqref{eq:global-kkt} holds.  For \(i\in S_\ell\), define
    \[
        \alpha_i^{(\ell)}:=\frac{\pi_i}{\hat\pi_\ell},
        \qquad
        q_\ell(x):=
        \sum_{i\in S_\ell}
        \alpha_i^{(\ell)}\phi(\vmu_i;x),
        \qquad
        \beta_i^{(\ell)}
        :=
        \E_{\vx\sim\ell}
        \frac{\alpha_i^{(\ell)}\phi(\vmu_i;\vx)}
        {q_\ell(\vx)} .
    \]
    Then
    \[
        \sum_{\ell=1}^m
        \pi_\ell^*
        \kl(\beta^{(\ell)}\|\alpha^{(\ell)})
        \le
        \eps_\pi
        +
        \sfA_{\rm g}e^{-c\Delta^2}.
    \]
\end{lemma}

\begin{proof}
    The group weights are positive by
    \eqref{eq:global-loose-group-mass}, so \(q_\ell\) is well-defined.  Let
    \[
        \bar q_\ell(\vx):=
        \sum_{i\in S_\ell}\beta_i^{(\ell)}\phi(\vmu_i;\vx),
        \qquad
        \bar p(\vx):=
        \sum_{\ell=1}^m\hat\pi_\ell\bar q_\ell(\vx).
    \]
    Thus \(\bar p\) is obtained from \(p\) by keeping each group mass
    \(\hat\pi_\ell\) fixed and replacing only the normalized weights inside
    each group. Also denote the corresponding weight vector \(\bar\vpi\) by
    $\bar\pi_i
        :=
        \hat\pi_\ell\beta_i^{(\ell)}$
    for $i\in S_\ell$.

    Recall $p_{-\ell}(\vx):=
            \sum_{r\ne\ell}\sum_{i\in S_r}\pi_i\phi(\vmu_i;\vx)$ and $
            \rho_\ell(\vx):=\frac{p_{-\ell}(\vx)}{\hat\pi_\ell q_\ell(\vx)}$ in \cref{lem:global-separated-overlap}.
    For \(x\sim N(\vmu_\ell^*,I)\), the lower bound
    \(\bar p(\vx)\ge \hat\pi_\ell\bar q_\ell(\vx)\) and the identity
    \(p(\vx)=\hat\pi_\ell q_\ell(\vx)(1+\rho_\ell(\vx))\) give
    \[
    \begin{aligned}
        \log\frac{\bar p(\vx)}{p(\vx)}
        &\ge
        \log\frac{\bar q_\ell(\vx)}{q_\ell(\vx)}
        -
        \log(1+\rho_\ell(\vx))
        =
        \log\frac{\bar q_\ell(\vx)}{q_\ell(\vx)}
        -
        \log(1+\rho_\ell(\vx))\\
        &=
        \log
        \sum_{i\in S_\ell}
        \frac{\alpha_i^{(\ell)}\phi(\vmu_i;\vx)}{q_\ell(\vx)}
        \frac{\beta_i^{(\ell)}}{\alpha_i^{(\ell)}}
        -
        \log(1+\rho_\ell(\vx))
        \ge
        \sum_{i\in S_\ell}
        \frac{\alpha_i^{(\ell)}\phi(\vmu_i;\vx)}{q_\ell(\vx)}
        \log\frac{\beta_i^{(\ell)}}{\alpha_i^{(\ell)}}
        -
        \log(1+\rho_\ell(\vx)).
    \end{aligned}
    \]
    where we use Jensen inequality at the end.
    
    Taking expectation over \(x\sim N(\vmu_\ell^*,I)\), we get
    \[
        \E_{\vx\sim\ell}
        \log\frac{\bar p(\vx)}{p(\vx)}
        \ge
        \kl(\beta^{(\ell)}\|\alpha^{(\ell)})
        -
        \E_{\vx\sim\ell}\log(1+\rho_\ell(\vx)).
    \]
    Multiplying by \(\pi_\ell^*\), summing over \(\ell\), and applying
    \eqref{eq:global-model-cross-ratio}, we obtain
    \[
        \E_{p^*}\log\frac{\bar p}{p}
        \ge
        \sum_{\ell=1}^m
        \pi_\ell^*
        \kl(\beta^{(\ell)}\|\alpha^{(\ell)})
        -
        \sfA_{\rm g}e^{-c\Delta^2}.
    \]
    On the other hand, by \cref{lem:weight-dual-gap-consequences}, the dual-gap condition
    \eqref{eq:global-kkt} implies the objective-gap bound
    $\L(\vpi,\vmu)
        \le
        \L(\bar\vpi,\vmu)+\eps_\pi$.
    Thus    
    \[
        \E_{p^*}\log\frac{\bar p}{p}
        =
        \L(\vpi,\vmu)-\L(\bar\vpi,\vmu)
        \le
        \eps_\pi.
    \]
    Combining the above two proves the claim.
\end{proof}

\begin{lemma}
\label{lem:global-loss-comparison}
    Suppose \(U(\vmu)\le B^2/2\), \(\L(\vpi,\vmu)\le L_0\), and
    \eqref{eq:global-kkt} holds.  With $\alpha_i^{(\ell)}:=\frac{\pi_i}{\hat\pi_\ell}$ and
    $q_\ell(x):=
        \sum_{i\in S_\ell}
        \alpha_i^{(\ell)}\phi(\vmu_i;x)$ as above,
    \[
        \L(\vpi,\vmu)
        \le
        C\sum_{\ell=1}^m
        \pi_\ell^*
        \kl(\phi(\vmu_\ell^*)\|q_\ell)
        +
        C\left(
            \eps_\pi
            +
            \sfA_{\rm g}e^{-c\Delta^2}
        \right).
    \]
\end{lemma}

\begin{proof}
    By \cref{lem:global-group-weight-control},
    \[
        \sum_{\ell=1}^m
        \abs{\hat\pi_\ell-\pi_\ell^*}
        \le
        \eps_\pi
        +
        \sfA_{\rm g}e^{-c\Delta^2},
        \qquad
        \hat\pi_\ell\ge\frac12\pi_\ell^* .
    \]
    Hence
    \begin{equation}
        \sum_{\ell=1}^m
        \pi_\ell^*
        \log\frac{\pi_\ell^*}{\hat\pi_\ell}
        \le
        C\left(
            \eps_\pi
            +
            \sfA_{\rm g}e^{-c\Delta^2}
        \right).
        \label{eq:global-group-weight-log-error}
    \end{equation}
    For \(\vx\sim N(\vmu_\ell^*,\mI)\), recall $\rho_\ell^*(x):=
            \frac{\sum_{r\ne\ell}\pi_r^*\phi(\vmu_r^*;x)}
            {\pi_\ell^*\phi(\vmu_\ell^*;x)}$, so
    \[
        p^*(\vx)
        =
        \pi_\ell^*\phi(\vmu_\ell^*;\vx)(1+\rho_\ell^*(\vx)),
        \qquad
        p(\vx)\ge \hat\pi_\ell q_\ell(\vx).
    \]
    Therefore
    \[
    \begin{aligned}
        \L(\vpi,\vmu)
        &=
        \sum_{\ell=1}^m
        \pi_\ell^*
        \E_{\vx\sim\ell}
        \log\frac{p^*(\vx)}{p(\vx)}
        \le
        \sum_{\ell=1}^m
        \pi_\ell^*
        \kl(\phi(\vmu_\ell^*)\|q_\ell)
        +
        \sum_{\ell=1}^m
        \pi_\ell^*
        \log\frac{\pi_\ell^*}{\hat\pi_\ell}
        +
        \sum_{\ell=1}^m
        \pi_\ell^*
        \E_{\vx\sim\ell}\log(1+\rho_\ell^*(\vx)).
    \end{aligned}
    \]
    The second term is bounded by
    \eqref{eq:global-group-weight-log-error}, and the third term is bounded by
    \eqref{eq:global-true-cross-ratio}.  This proves the claim.
\end{proof}

\section{Regularity condition}\label{appendix: reg condition}
In this section, we prove a local smoothness bound for \(\L\) and a Gaussian ratio estimate used in both the local and global phases.

\begin{lemma}[Local smoothness]
\label{lem:local-smoothness-self-bound}
    Fix \(\vpi\). Suppose for every \(\ell\in[m]\), we have
    $\hat\pi_\ell:=\sum_{i\in S_\ell}\pi_i
        \ge c_{\rm cov}\pi_\ell^*$ and $
        \frac1{\hat\pi_\ell}
        \sum_{i\in S_\ell}
        \pi_i\norm{\vmu_i-\vmu_\ell^*}^2
        \le B^2$.
    Let \(\vdelta=(\vdelta_1,\ldots,\vdelta_n)\) be any direction satisfying,
    for every \(\ell\in[m]\),
    $\frac1{\hat\pi_\ell}
        \sum_{i\in S_\ell}
        \pi_i\norm{\vdelta_i}^2
        \le D^2$.
    Then, for $\sfA_{\eta}(D)
        :=
        C\left(
            d+B^2+D^2+\log\frac1{c_{\rm cov}\pimins}
        \right)$
    with \(C\) sufficiently large,
    \[
        \L(\vpi,\vmu+\vdelta)
        \le
        \L(\vpi,\vmu)
        +
        \left\langle
            \nabla_{\vmu}\L(\vpi,\vmu),
            \vdelta
        \right\rangle
        +
        \frac{\sfA_{\eta}(D)}{2}
        \norm{\vdelta}_F^2 .
    \]
    Moreover, with \(\vg:=\nabla_{\vmu}\L(\vpi,\vmu)\), if
    $\sfA_{\eta}
        :=
        C\left(
            d+B^2+\log\frac1{c_{\rm cov}\pimins}
            +
            \frac1{c_{\rm cov}\pimins}
        \right)$,
    then
    \begin{equation}
        \norm{\vg}_F^2\le \sfA_{\eta}.
        \label{eq:local-gradient-size-bound}
    \end{equation}
    In addition, for any direction \(\vdelta\) satisfying
    $\norm{\vdelta+\eta \vg}_F
        \le
        \frac{\eta}{4}\norm{\vg}_F$ and 
        $\eta\le \frac{c}{\sfA_{\eta}}$,
    we have
    \[
        \L(\vpi,\vmu+\vdelta)
        \le
        \L(\vpi,\vmu)
        -
        \frac{\eta}{2}\norm{\vg}_F^2 .
    \]
    Finally,
    \begin{equation}
        \norm{\nabla_{\vmu}\L(\vpi,\vmu)}_F^2
        =
        \norm{\vg}_F^2
        \le
        2\sfA_{\eta}\L(\vpi,\vmu).
        \label{eq:gradient-self-bound}
    \end{equation}
\end{lemma}

\begin{proof}
    We first prove the smoothness bound for an arbitrary bounded direction
    \(\vdelta\).  For \(u\in[0,1]\), write
    \[
        \vmu(u):=\vmu+u\vdelta .
    \]
    The following bounds are with respect to the fixed groups \(S_\ell\) from
    the base point \(\vmu\).  They are used only to lower bound the model
    density along the segment, and no reassignment of the groups is needed inside
    this lemma.
    
    By the assumptions on \(\vmu\) and \(\vdelta\),
    \[
    \begin{aligned}
        \frac1{\hat\pi_\ell}
        \sum_{i\in S_\ell}
        \pi_i\norm{\vmu_i(u)-\vmu_\ell^*}^2
        &\le
        \frac2{\hat\pi_\ell}
        \sum_{i\in S_\ell}
        \pi_i\norm{\vmu_i-\vmu_\ell^*}^2
        +
        \frac{2u^2}{\hat\pi_\ell}
        \sum_{i\in S_\ell}
        \pi_i\norm{\vdelta_i}^2
        \le
        2B^2+2D^2 .
    \end{aligned}
    \]
    Thus the same local coverage and second-moment bounds hold along the
    segment, with \(B^2\) replaced by \(2B^2+2D^2\).

    For one sample \(\vx\), define
    \[
        f_\vx(\vmu(u))
        :=
        -\log\sum_{i=1}^n\pi_i\phi(\vmu_i(u);\vx),
        \qquad
        \psi_i^u(\vx)
        :=
        \frac{\pi_i\phi(\vmu_i(u);\vx)}
        {\sum_{j=1}^n\pi_j\phi(\vmu_j(u);\vx)} .
    \]
    Then
    \[
        \nabla_{\vmu_i} f_\vx(\vmu(u))
        =
        \psi_i^u(\vx)(\vmu_i(u)-\vx).
    \]
    For a direction \(\vv=(\vv_1,\ldots,\vv_n)\), set
    \(F(t):=f_\vx(\vmu(u)+t\vv)\).  Direct differentiation gives
    \[
    \begin{aligned}
        F''(0)
        &=
        \sum_{i=1}^n
        \psi_i^u(\vx)\norm{\vv_i}^2
        -
        \left[
            \sum_{i=1}^n
            \psi_i^u(\vx)
            \left\langle
                \vx-\vmu_i(u),\vv_i
            \right\rangle^2
            -
            \left(
                \sum_{i=1}^n
                \psi_i^u(\vx)
                \left\langle
                    \vx-\vmu_i(u),\vv_i
                \right\rangle
            \right)^2
        \right],\\
        |F''(0)|
        &\le
        \sum_{i=1}^n
        \psi_i^u(\vx)\norm{\vv_i}^2
        +
        \sum_{i=1}^n
        \psi_i^u(\vx)
        \left\langle
            \vx-\vmu_i(u),\vv_i
        \right\rangle^2
        \le
        \left(
            1+
            \sum_{i=1}^n
            \psi_i^u(\vx)\norm{\vx-\vmu_i(u)}^2
        \right)
        \sum_{i=1}^n\norm{\vv_i}^2 .
    \end{aligned}
    \]
    Therefore
    \[
        \norm{\nabla_{\vmu}^2 f_\vx(\vmu(u))}_{\op}
        \le
        1+
        \sum_{i=1}^n
        \psi_i^u(\vx)\norm{\vx-\vmu_i(u)}^2 .
    \]
    It remains to bound the right hand side.  Fix a true component
    \(\ell\), and consider \(\vx\) under \(\phi(\vmu_\ell^*)\). We use the Gibbs variational formula
    \[
        -\log\sum_{i=1}^n \pi_i e^{-a_i}
        =
        \min_{q\in\Delta_n}
        \left\{
            \sum_{i=1}^n q_i a_i
            +
            \sum_{i=1}^n q_i\log\frac{q_i}{\pi_i}
        \right\}.
    \]
    Applying this with
    \(a_i=\frac12\norm{\vx-\vmu_i(u)}^2\), the minimizer is
    \(q_i=\psi_i^u(\vx)\).  Therefore, for any distribution \(q\in\Delta_n\),
    \[
    \begin{aligned}
        \frac12
        \sum_{i=1}^n
        \psi_i^u(\vx)\norm{\vx-\vmu_i(u)}^2
        +
        \sum_{i=1}^n
        \psi_i^u(\vx)
        \log\frac{\psi_i^u(\vx)}{\pi_i}
        \le
        \frac12
        \sum_{i=1}^n
        q_i\norm{\vx-\vmu_i(u)}^2
        +
        \sum_{i=1}^n
        q_i\log\frac{q_i}{\pi_i}.
    \end{aligned}
    \]
    Taking \(q_i=\pi_i/\hat\pi_\ell\) for \(i\in S_\ell\), and \(q_i=0\)
    otherwise, and dropping the nonnegative entropy term on the left, gives 
    \[
    \begin{aligned}
        \sum_{i=1}^n
        \psi_i^u(\vx)\norm{\vx-\vmu_i(u)}^2
        &\le
        \frac1{\hat\pi_\ell}
        \sum_{i\in S_\ell}
        \pi_i\norm{\vx-\vmu_i(u)}^2
        +
        2\log\frac1{\hat\pi_\ell}\\
        &\le
        2\norm{\vx-\vmu_\ell^*}^2
        +
        \frac2{\hat\pi_\ell}
        \sum_{i\in S_\ell}
        \pi_i\norm{\vmu_i(u)-\vmu_\ell^*}^2
        +
        2\log\frac1{\hat\pi_\ell}\\
        &\le
        2\norm{\vx-\vmu_\ell^*}^2
        +
        4B^2+4D^2
        +
        2\log\frac1{c_{\rm cov}\pimins}.
    \end{aligned}
    \]
    Averaging over
    \(p^*=\sum_{\ell=1}^m\pi_\ell^*\phi(\vmu_\ell^*)\) gives
    \[
        \norm{\nabla_{\vmu}^2\L(\vpi,\vmu(u))}_{\op}
        \le
        \sfA_{\eta}(D)
        \qquad
        \text{for every }u\in[0,1].
    \]
    The Taylor upper bound along the segment from \(\vmu\) to
    \(\vmu+\vdelta\) follows:
    \[
        \L(\vpi,\vmu+\vdelta)
        \le
        \L(\vpi,\vmu)
        +
        \left\langle
            \nabla_{\vmu}\L(\vpi,\vmu),
            \vdelta
        \right\rangle
        +
        \frac{\sfA_{\eta}(D)}{2}
        \norm{\vdelta}_F^2 .
    \]
    
    We next prove the gradient size bound \eqref{eq:local-gradient-size-bound}.  At \(u=0\), by Cauchy--Schwarz and
    \(\psi_i^0(\vx)\le1\),
    \[
    \begin{aligned}
        \norm{\vg}_F^2
        &=
        \sum_{i=1}^n
        \left\|
        \E_{p^*}
        \left[
            \psi_i^0(\vx)(\vmu_i-\vx)
        \right]
        \right\|^2
        \le
        \E_{p^*}
        \sum_{i=1}^n
        \psi_i^0(\vx)\norm{\vmu_i-\vx}^2
        \lesssim
            d+B^2+\log\frac1{c_{\rm cov}\pimins}
        \le
        \sfA_{\eta}.
    \end{aligned}
    \]

    Now assume
    \(\norm{\vdelta+\eta \vg}_F\le \frac \eta4\norm{\vg}_F\).  Then
    $\norm{\vdelta}_F
        \le
        \frac54\eta\norm{\vg}_F$.
    Hence, for every \(\ell\),
    \[
    \begin{aligned}
        \frac1{\hat\pi_\ell}
        \sum_{i\in S_\ell}
        \pi_i\norm{\vdelta_i}^2
        &\le
        \frac{\norm{\vdelta}_F^2}{c_{\rm cov}\pimins}
        \le
        \frac{C\eta^2\norm{\vg}_F^2}{c_{\rm cov}\pimins}
        \le
        1,
    \end{aligned}
    \]
    after choosing \(c>0\) sufficiently small and using the definition of
    \(\sfA_{\eta}\) and
    \(\eta\le c/\sfA_{\eta}\).  Thus the arbitrary-direction smoothness bound
    applies with \(D=1\).  Also,
    \[
        \left\langle \vg,\vdelta\right\rangle
        =
        -\eta\norm{\vg}_F^2
        +
        \left\langle
            \vg,\vdelta+\eta \vg
        \right\rangle
        \le
        -\frac34\eta\norm{\vg}_F^2,
        \quad\text{and}\quad
        \frac{\sfA_{\eta}}2\norm{\vdelta}_F^2
        \le
        C\sfA_{\eta}\eta^2\norm{\vg}_F^2
        \le
        \frac14\eta\norm{\vg}_F^2 .
    \]
    Therefore
    \[
        \L(\vpi,\vmu+\vdelta)
        \le
        \L(\vpi,\vmu)
        -
        \frac{\eta}{2}\norm{\vg}_F^2 .
    \]

    Finally, apply the arbitrary-direction bound with
    \(\vdelta=-\sfA_{\eta}^{-1}\vg\).  By the gradient size bound and the
    definition of \(\sfA_{\eta}\), this direction satisfies the required
    \(D=1\) bound.  Hence
    \[
    \begin{aligned}
        \L\left(
            \vpi,
            \vmu-\sfA_{\eta}^{-1}\vg
        \right)
        &\le
        \L(\vpi,\vmu)
        -
        \frac1{\sfA_{\eta}}\norm{\vg}_F^2
        +
        \frac1{2\sfA_{\eta}}\norm{\vg}_F^2
        =
        \L(\vpi,\vmu)
        -
        \frac1{2\sfA_{\eta}}\norm{\vg}_F^2 .
    \end{aligned}
    \]
    Since the KL loss is nonnegative, the left-hand side is at least \(0\).
    Therefore, we get \eqref{eq:gradient-self-bound}
    \[
        \norm{\vg}_F^2
        \le
        2\sfA_{\eta}\L(\vpi,\vmu).
    \]
\end{proof}

\begin{lemma}[Cell Gaussian ratio moment bound]
\label{lem:cell-gaussian-ratio-moment}
    Let \(\Omega_1,\ldots,\Omega_m\) denote the true Voronoi cells.  Fix
    constants \(1\le q<\infty\), \(0\le K<\infty\), and \(C_0<\infty\).
    Let \(0\le \rho\le B\), \(0\le\lambda\le C_0B\), and let
    \(\vh,\vh',\vw\in\mathbb R^d\) satisfy \(\norm{\vh},\norm{\vh'}\le \rho\). Suppose
    \(\vnu\) satisfies $\norm{\vnu-\vmu_r^*}\le B$
    for some \(r\in[m]\).  Let \(\sfA_{\rm cell}:=\poly(d,m,1+B)\).  If \(m\ge2\), assume
    \[
        \Delta
        \ge
        C_{\rm sep}
        \left(
            B+\rho+\lambda+\sqrt{\log \sfA_{\rm cell}}
        \right).
    \]
    When \(m=1\), this separation condition is omitted.

    For \(s,\ell\in[m]\), define
    \[
    \begin{aligned}
        I_s(\vnu,\ell)
        &:=
        \int_{\Omega_s}
        \phi(\vnu;\vx)
        \left(
            \frac{\phi(\vmu_\ell^*;\vx)}
            {\phi(\vmu_s^*;\vx)}
        \right)^q
        \left|
            \left\langle
                \vx-\vmu_\ell^*,\vw
            \right\rangle
        \right|^K
        \exp\left(
            C_0
            \left|
                \left\langle
                    \vx-\vmu_\ell^*,\vh
                \right\rangle
            \right|
            +
            \lambda|\ip{\vx-\vmu_s^*,\vh'}|
            +
            C_0\rho^2
        \right)
        \rd\vx .
    \end{aligned}
    \]
    Then
    \begin{equation}
        \sum_{s=1}^m I_s(\vnu,\ell)
        \le
        \exp\left(CB\rho+C\rho^2+C\lambda B+C\lambda^2\right)
        \sfA_{\rm cell}^{C}\norm{\vw}^K ,
        \label{eq:cell-gaussian-ratio-homogeneous}
    \end{equation}
    For \(K=0\), the factor \(\norm{\vw}^K\) is interpreted as one. Moreover,
    each off-diagonal cell satisfies
    \begin{equation}
        I_s(\vnu,\ell)
        \le
        e^{-c\Delta^2}\norm{\vw}^K,
        \qquad
        (r,s,\ell)\ne(\ell,\ell,\ell).
        \label{eq:cell-gaussian-ratio-offdiag}
    \end{equation}

    The analogous estimates with
    \(1+\abs{\ip{\vx-\vmu_\ell^*,\vw}}^K\) in place of
    \(\abs{\ip{\vx-\vmu_\ell^*,\vw}}^K\) hold with
    \(\norm{\vw}^K\) replaced by \(1+\norm{\vw}^K\).

    We will also use the following product-form off-diagonal version.  Fix nonnegative integers \(M_1,M_2\).  Let
    \(\alpha_a,\beta_b\in\{r,\ell\}\), let
    \(\vw_a,\vh_b\in\mathbb R^d\), and suppose
    \(\norm{\vh_b}\le B\).  Define
    \[
        \mathcal F(\vx)
        :=
        \prod_{a=1}^{M_1}
        \abs{\ip{\vx-\vmu_{\alpha_a}^*,\vw_a}}
        \cdot
        \exp\left(
            C_0\sum_{b=1}^{M_2}
            \abs{\ip{\vx-\vmu_{\beta_b}^*,\vh_b}}
            +
            C_0\sum_{b=1}^{M_2}\norm{\vh_b}^2
        \right).
    \]
    If \((r,s,\ell)\ne(\ell,\ell,\ell)\), then
    \begin{equation}
        \int_{\Omega_s}
        \phi(\vnu;\vx)
        \left(
            \frac{\phi(\vmu_\ell^*;\vx)}
                 {\phi(\vmu_s^*;\vx)}
        \right)^q
        \mathcal F(\vx)
        \rd\vx
        \le
        e^{-c\Delta^2}
        \prod_{a=1}^{M_1}\norm{\vw_a}.
        \label{eq:cell-gaussian-ratio-product-offdiag}
    \end{equation}
\end{lemma}

\begin{proof}
    Write \(\vnu=\vmu_r^*+\va\), where \(\norm{\va}\le B\).  We prove the
    homogeneous estimate \eqref{eq:cell-gaussian-ratio-homogeneous} and the
    off-diagonal estimate \eqref{eq:cell-gaussian-ratio-offdiag}.  The
    inhomogeneous estimates follow by adding the case \(K=0\).  If
    \(\vw=0\) and \(K>0\), the claim is trivial, so assume \(\vw\ne0\).  All
    constants below may depend on the fixed constants \(q,K,C_0\).

    We use the following elementary one-dimensional estimates.  Let
    \(Z\sim N(0,1)\), \(D\ge0\), \(\abs b\le B\), \(0\le\rho\le B\), and
    \(0\le\lambda\le CB\).  For fixed constants \(a\ge1\) and \(K<\infty\),
    \begin{align}
        &\E\left[
            \mathbf 1\left\{Z+b\le \frac D2\right\}
            (1+D+B+\abs Z)^K
            \exp\left(
                aD(Z+b)-\frac{aD^2}{2}
                +C\rho(D+B+\abs Z)+C\lambda(D+B+\abs Z)
            \right)
        \right]\notag\\
        &\qquad\le
        \sfA_{\rm cell}^{C}
        e^{CB\rho+C\rho^2+C\lambda B+C\lambda^2}
        e^{-c(D-CB)_+^2},
        \label{eq:cell-ratio-one-dim-ratio}
    \end{align}
    Here \((u)_+:=\max\{u,0\}\).  We prove
    \eqref{eq:cell-ratio-one-dim-ratio}. 
    Up to changing the constant
    \(\sfA_{\rm cell}^{C}\), the fixed-degree polynomial can be absorbed by
    replacing the exponential weight \(C(\rho+\lambda)\abs Z\) with a slightly
    larger one.  Using
    \(e^{C(\rho+\lambda)\abs Z}\le e^{C(\rho+\lambda)Z}+e^{-C(\rho+\lambda)Z}\),
    it is enough to bound, for \(\sigma\in\{-1,1\}\),
    \[
    \begin{aligned}
        &\exp\left(
            -\frac{aD^2}{2}+aDb+C(\rho+\lambda)(D+B)
        \right)
        \E\left[
            \mathbf 1\{Z\le D/2-b\}
            \exp\left((aD+\sigma C(\rho+\lambda))Z\right)
        \right].
    \end{aligned}
    \]
    By the identity
    $\E[e^{tZ}\mathbf 1\{Z\le u\}]
        =
        e^{t^2/2}\Phi(u-t),$
    this equals
    \[
    \begin{aligned}
        \exp\left(
            -\frac{aD^2}{2}+aDb+C(\rho+\lambda)(D+B)
            +\frac12(aD+\sigma C(\rho+\lambda))^2
        \right)
        \Phi\left(
            \left(\frac12-a\right)D-b-\sigma C(\rho+\lambda)
        \right).
    \end{aligned}
    \]
    Since \(a\ge1\), the Gaussian tail in the last factor cancels the positive
    part of the completing-square prefactor.  Keeping the remaining tail gives
    the factor \(e^{-c(D-CB)_+^2}\), while the non-tail terms are bounded by
    \[
        \exp(CB\rho+C\rho^2+C\lambda B+C\lambda^2).
    \]
    This proves \eqref{eq:cell-ratio-one-dim-ratio}.

    We now bound the contribution of each cell \(\Omega_s\).  First suppose
    \(r=s\).  If \(s=\ell\), then the density ratio is one.  Under
    \(\phi(\vnu)\), the mean of \(\vx-\vmu_\ell^*\) has norm at most \(B\), and
    a standard Gaussian moment estimate gives
    \[
        I_\ell(\vnu,\ell)
        \le
        e^{CB\rho+C\rho^2+C\lambda B+C\lambda^2}
        \sfA_{\rm cell}^{C}\norm{\vw}^K .
    \]
    This is the only diagonal contribution.

    If \(r=s\) but \(s\ne\ell\), let
    \(D=\norm{\vmu_\ell^*-\vmu_s^*}\) and
    \(\vu=(\vmu_\ell^*-\vmu_s^*)/D\).  On \(\Omega_s\),
    $\langle \vx-\vmu_s^*,\vu\rangle\le D/2,$
    and
    \[
        \frac{\phi(\vmu_\ell^*;\vx)}{\phi(\vmu_s^*;\vx)}
        =
        \exp\left(
            D\left\langle \vx-\vmu_s^*,\vu\right\rangle
            -\frac{D^2}{2}
        \right).
    \]
    Writing \(\vx=\vmu_s^*+\va+\vZ\), the preceding display is exactly of the
    form \eqref{eq:cell-ratio-one-dim-ratio}, with the remaining Gaussian
    coordinates contributing only fixed-degree moments.  Hence
    \[
        I_s(\vnu,\ell)
        \le
        e^{CB\rho+C\rho^2+C\lambda B+C\lambda^2}
        \sfA_{\rm cell}^{C}
        e^{-c(\Delta-CB)_+^2}
        \norm{\vw}^K .
    \]

    It remains to consider \(r\ne s\).  By Cauchy--Schwarz, the cell integral is
    bounded by the product of two factors.  The first factor is the same cell
    integral with ambient density \(\phi(\vmu_s^*)\), with exponent \(2q\) and
    doubled constants; it is controlled by the case \(r=s\) just proved.  The
    second factor is
    \[
        \left(
        \int_{\Omega_s}
        \phi(\vmu_s^*;\vx)
        \left(
            \frac{\phi(\vmu_r^*;\vx)}{\phi(\vmu_s^*;\vx)}
        \right)^2
        \left(
            \frac{\phi(\vnu;\vx)}{\phi(\vmu_r^*;\vx)}
        \right)^2
        \rd\vx
        \right)^{1/2}.
    \]
    If \(\vnu=\vmu_r^*\), the last ratio is one.  Otherwise
    \(\norm{\vnu-\vmu_r^*}\le B\), so the last ratio contributes only a linear
    Gaussian exponential generated by a displacement of size \(B\).  Applying
    \eqref{eq:cell-ratio-one-dim-ratio}, with \(a=2\), to the bisector between
    \(\vmu_r^*\) and \(\vmu_s^*\), and using
    \(\norm{\vmu_r^*-\vmu_s^*}\ge\Delta\), gives
    \[
        \left(
        \int_{\Omega_s}
        \phi(\vmu_s^*;\vx)
        \left(
            \frac{\phi(\vmu_r^*;\vx)}{\phi(\vmu_s^*;\vx)}
        \right)^2
        \left(
            \frac{\phi(\vnu;\vx)}{\phi(\vmu_r^*;\vx)}
        \right)^2
        \rd\vx
        \right)^{1/2}
        \le
        \sfA_{\rm cell}^{C}
        e^{CB^2+CB\rho+C\rho^2+C\lambda B+C\lambda^2}
        e^{-c(\Delta-CB)_+^2}.
    \]
    Here we absorbed the square root into the constants.   Thus the \(r\ne s\) contribution satisfies the raw off-diagonal bound
    \[
        I_s(\vnu,\ell)
        \le
        \sfA_{\rm cell}^{C}
        e^{CB^2+CB\rho+C\rho^2+C\lambda B+C\lambda^2}
        e^{-c(\Delta-CB)_+^2}
        \norm{\vw}^K .
    \]

    Combining the two off-diagonal cases, namely \(r=s\ne\ell\) and
    \(r\ne s\), we have the raw bound
    \[
        I_s(\vnu,\ell)
        \le
        \sfA_{\rm cell}^{C}
        e^{CB^2+CB\rho+C\rho^2+C\lambda B+C\lambda^2}
        e^{-c(\Delta-CB)_+^2}
        \norm{\vw}^K
    \]
    whenever \((r,s,\ell)\ne(\ell,\ell,\ell)\).  Under the stated separation
    condition, the prefactor is absorbed into the Gaussian tail, so after
    increasing \(C_{\rm sep}\) and decreasing \(c\),
    \[
        I_s(\vnu,\ell)\le e^{-c\Delta^2}\norm{\vw}^K .
    \]
    This proves \eqref{eq:cell-gaussian-ratio-offdiag}.

    Summing over the finitely many cells proves
    \eqref{eq:cell-gaussian-ratio-homogeneous}: the only diagonal contribution
    is \(r=s=\ell\), which is bounded by the same-cell Gaussian moment estimate,
    while all off-diagonal contributions are exponentially small and hence
    absorbed into \(\sfA_{\rm cell}^{C}\).  The inhomogeneous version \(1+\abs{\ip{\vx-\vmu_\ell^*,\vw}}^K\)
    follows by applying the same argument to the \(1\) term and to the
    homogeneous term separately.

    The product-form off-diagonal estimate
    \eqref{eq:cell-gaussian-ratio-product-offdiag} follows from the same
    proof as \eqref{eq:cell-gaussian-ratio-offdiag}.  The extra polynomial
    factors are handled by the fixed Gaussian moment bound.  The finitely many
    exponential linear factors are centered only at the ambient true center
    \(\vmu_r^*\) or the ratio numerator \(\vmu_\ell^*\).  Applying the same
    one-dimensional completing-square estimates above, and using that
    \((r,s,\ell)\ne(\ell,\ell,\ell)\), gives a raw off-diagonal bound of the
    form
    \[
        \sfA_{\rm cell}^{C}
        \exp(CB^2+CB\rho+C\rho^2+C\lambda B+C\lambda^2)
        e^{-c(\Delta-CB)_+^2}
        \prod_{a=1}^{M_1}\norm{\vw_a}.
    \]
    The stated separation condition absorbs the prefactor into the Gaussian
    tail, proving \eqref{eq:cell-gaussian-ratio-product-offdiag}.
\end{proof}

\section{Proof of main result Theorem~\ref{thm: main}}
We give the proof of main result below, which is a combination of global convergence, local convergence, and random initialization.

\thmmain*
\begin{proof}
    Let \(\mathcal E_{\rm init}\) be the initialization event in
    \cref{lem: init}. Set \(B^2\asymp dn\) and \(L_0\asymp d+\log n\), with constants large
    enough so that, by the assumed lower bound on \(n\),
    with probability at least \(1-\delta\)
    \[
        \L(\vpi^{(0)},\vmu^{(0)})\le L_0,
        \qquad
        U(\vmu^{(0)})\le B^2/2.
    \]
    
    We choose the parameter here so that condition in global and local phase theorems \cref{thm: global main} and \cref{thm: local main} are satisfied.
    Choose \(\eps_0=1/\poly(d,m,n,1/\pimins)\) small enough so that all
    local-threshold conditions in \cref{thm: local main} hold with this
    value of \(B\) and \(A_{\eta}\) and \(A_{\pi}\) large enough to dominate the corresponding
    polynomial quantities in \cref{thm: global main} and \cref{thm: local main}. Then set $\eps_\pi$ accordingly to satisfy the weight subproblem requirements. Finally, choosing the separation constant in \cref{assump: delta} be large enough, so that the separation hypotheses in both \cref{thm: global main} and \cref{thm: local main}
    are also satisfied.
    
    First suppose \(\eps\ge \eps_0\). Applying \cref{thm: global main} with
    target \(\eps\) gives
    \[
        \L(\vpi^{(T)},\vmu^{(T)})\le \eps
        \qquad
        \text{for some }
        T\le \frac{CB^2}{\eta\eps}
        =
        O\!\left(\frac{dn}{\eta\eps}\right).
    \]

    Now suppose \(\eps<\eps_0\). Applying \cref{thm: global main} with
    target \(\eps_0\) gives a time \(T_1\) such that
    \[
        \L(\vpi^{(T_1)},\vmu^{(T_1)})\le \eps_0,
        \qquad
        U(\vmu^{(T_1)})\le \frac{B^2}{2},
        \qquad
        T_1\le \frac{CB^2}{\eta\eps_0}
        =
        O\!\left(\frac{dn}{\eta\eps_0}\right).
    \]
    These are the initial conditions for \cref{thm: local main}. Therefore \cref{thm: local main} yields
    \[
        \L(\vpi^{(t)},\vmu^{(t)})
        \le
        \frac{A_{\rm rate}^2}
        {\eta^2(t-T_1)^2+A_{\rm rate}^2/\eps_0}
        \qquad
        \text{for all } t\ge T_1
    \]
    until the target loss is reached. The right-hand side is at most
    \(\eps\) once
    \[
        t-T_1
        \ge
        \frac{A_{\rm rate}}{\eta}
        \sqrt{\frac1\eps-\frac1{\eps_0}} .
    \]
    This proves the claimed bound for \(\eps<\eps_0\).

    It remains to justify the recovery statement. Since theorems keep
    \(U(\vmu)\le B^2\), so every convergent subsequence has a bounded
    limit. Applying the identifiability theorem
    \cref{cor:local-identifiability-consequences} gives
    for every \(S_\ell\),
    \[
        \sum_{i\in S_\ell}
        \pi_i\norm{\vmu_i-\vmu_\ell^*}^2
        \to 0 .
    \]
    Hence any component with positive limiting weight must converge to one
    of the true centers. All other limiting components have zero weight.
    This proves the recovery property and completes the proof.
\end{proof}
\section{Sample complexity}
\label{appendix: sample complexity}

In this section, we prove a finite-sample version of the population theorem using \cref{alg: fixed sample}. We first prove a uniform concentration result on a bounded parameter class, so that the empirical losses and gradients uniformly approximate their population
counterparts. We then repeat the global and local phase arguments with the
empirical loss.

\begin{algorithm}[t]
\caption{Fixed-sample Gradient-EM with near-optimal weight updates}
\label{alg: fixed sample}
\begin{algorithmic}[1]
\STATE \textbf{Input:} Stepsize \(\eta\), iterations \(T\), target error
\(\eps\), weight subproblem error \(\eps_\pi\), sample size \(N\).
\STATE \textbf{Random initialization:}
\(\vmu_i^{(0)}\sim p^*\) independently for every \(i\in[n]\), and
\(\pi_i^{(0)}=1/n\).
\STATE Draw fixed samples \(\vx_1,\ldots,\vx_N\sim p^*\), and form the
empirical loss \(\hL\) in \eqref{eq:empirical-loss}.
\FOR{\(t=0,\ldots,T-1\)}
    \STATE Let \(\vpi^{(t+1)}\) be an \(\eps_\pi\)-optimal solution of
    \[
        \min_{\vpi\in\gP}\hL(\vpi,\vmu^{(t)}).
    \]
    \STATE Set
    \[
        \vmu^{(t+1)}
        =
        \vmu^{(t)}
        -
        \eta\nabla_\vmu\hL(\vpi^{(t+1)},\vmu^{(t)}).
    \]
\ENDFOR
\STATE \textbf{Output:} \((\vpi^{(T)},\vmu^{(T)})\).
\end{algorithmic}
\end{algorithm}

The empirical loss is
\begin{align}
    \hL(\vpi,\vmu)
    =
    \frac1N\sum_{a=1}^N
    \log\frac{p^*(\vx_a)}{p(\vx_a)}
    =
    \frac1N\sum_{a=1}^N
    \ell(p(\vx_a))-\ell(p^*(\vx_a)),
    \qquad
    \ell(u)=-\log u .
    \label{eq:empirical-loss}
\end{align}
For fixed \((\vpi,\vmu)\), the empirical loss and gradients are unbiased
estimators of their population counterparts.

The \(\eps_\pi\)-optimality condition in \cref{alg: fixed sample} means that
the empirical weight step is loss non-increasing and has small empirical
simplex dual gap:
\begin{align}
    \widehat G_\pi(\vpi^{(t+1)},\vmu^{(t)})
    :=
    \max_{\bar\vpi\in\gP}
    \left\langle
        \nabla_{\vpi}\hL(\vpi^{(t+1)},\vmu^{(t)}),
        \vpi^{(t+1)}-\bar\vpi
    \right\rangle
    \le \eps_\pi,
    \qquad
    \hL(\vpi^{(t+1)},\vmu^{(t)})
    \le
    \hL(\vpi^{(t)},\vmu^{(t)}).
    \label{eq:empirical-weight-certificate}
\end{align}

We now state the finite-sample theorem.  The result is stated for the same
target accuracy as the population theorem.

\thmsamplecomplexity*

Let
\[
    \Theta_{2B}
    :=
    \left\{
        (\vpi,\vmu):\vpi\in\gP,\ U(\vmu)\le 4B^2
    \right\}.
\]
The enlarged radius \(2B\) is used only for concentration and smoothness.

\begin{lemma}[Uniform empirical event]
\label{lem:uniform-empirical-event}
    Under \cref{assump: delta}, with probability at least \(1-\delta\), the
    following estimates hold simultaneously for all
    \((\vpi,\vmu)\in\Theta_{2B}\), provided
    \[
        N
        \ge
        C\sfA_{\rm samp}
        \left(
            \gamma_L^{-2}
            +
            \gamma_\mu^{-2}
            +
            \gamma_\pi^{-2}
            +
            1
        \right)
        \log
        \frac{C\sfA_{\rm samp}}
        {\delta\min\{\gamma_L,\gamma_\mu,\gamma_\pi,1\}} .
    \]
    First,
    \begin{align}
        \abs{\hL(\vpi,\vmu)-\L(\vpi,\vmu)}
        &\le \gamma_L,
        &
        \norm{
            \nabla_\vmu\hL(\vpi,\vmu)
            -
            \nabla_\vmu\L(\vpi,\vmu)
        }_F
        &\le \gamma_\mu .
        \label{eq:uniform-loss-gradient}
    \end{align}
    Second,
    \begin{align}
        \sum_{i=1}^n
        \pi_i
        \abs{
            \nabla_{\pi_i}\hL(\vpi,\vmu)
            -
            \nabla_{\pi_i}\L(\vpi,\vmu)
        }
        \le \gamma_\pi .
        \label{eq:uniform-pi-gradient}
    \end{align}
    Finally, whenever the line segment $\{(\vpi,\vmu+s\vdelta):s\in[0,1]\}$ is contained in \(\Theta_{2B}\),
    \begin{align}
        \hL(\vpi,\vmu+\vdelta)
        \le
        \hL(\vpi,\vmu)
        +
        \left\langle
            \nabla_\vmu\hL(\vpi,\vmu),\vdelta
        \right\rangle
        +
        A_\eta\norm{\vdelta}_F^2 .
        \label{eq:uniform-empirical-smoothness}
    \end{align}
\end{lemma}

In the phase proofs we work on the event in
\cref{lem:uniform-empirical-event} with
\begin{align}
    \gamma_L
    =
    \gamma_\pi
    :=
    \frac{c}{A_\pi}\eps_{\rm stat},
    \qquad
    \gamma_\mu
    :=
    c\min\left\{
        \frac{\eps_0}{B},
        A_{\rm rate}^{-1/2}\eps^{3/4}
    \right\}.
    \label{eq:sample-gamma-choices}
\end{align}
After increasing \(\sfA_{\rm samp}\), the sample size
\eqref{eq:fixed-sample-size} implies the sample-size requirement in
\cref{lem:uniform-empirical-event}.

We will use the population global and local projection estimates in the
following consequence form.  In the global phase, the required weight
hypotheses are
\begin{align}
    \sum_i \pi_i
    \abs{\nabla_{\pi_i}\L(\vpi,\vmu)+1}
    &\le
    cA_\pi^{-1}\eps_0,
    \label{eq:robust-global-weight-kkt}\\
    \L(\vpi,\vmu)
    &\le
    \inf_{\bar\vpi\in\gP}\L(\bar\vpi,\vmu)
    +
    cA_\pi^{-1}\eps_0.
    \label{eq:robust-global-weight-gap}
\end{align}
In the local phase, the required weight hypothesis is
\begin{align}
    \sum_i \pi_i
    \abs{\nabla_{\pi_i}\L(\vpi,\vmu)+1}
    \le
    c_{\rm kkt}\L(\vpi,\vmu).
    \label{eq:robust-local-weight-kkt}
\end{align}
These are exactly the consequences of the population dual-gap condition used
in the population proofs through \cref{lem:weight-dual-gap-consequences}.
Thus the finite-sample argument only needs to transfer these weighted conditions.

\subsection{Fixed-sample global phase}

The next theorem is the finite-sample analogue of the global phase. Uniform concentration transfers the
empirical weight certificate and empirical gradient projection back to the population quantities used in \cref{thm: global main}.

\begin{theorem}[Fixed-sample global phase]
\label{thm:fixed-sample-global}
    Under the assumptions of \cref{thm: global main}, suppose
    \(\hL(\vpi^{(0)},\vmu^{(0)})\le L_0+1\) and
    \(U(\vmu^{(0)})\le B^2/2\) and \cref{lem:uniform-empirical-event} holds.  For any \(\eps\ge\eps_0\), \cref{alg: fixed sample} reaches
    \[
        \L(\vpi^{(T)},\vmu^{(T)})\le \eps,
        \qquad
        U(\vmu^{(T)})\le \frac{B^2}{2},
    \]
    within \(T\le CB^2/(\eta\eps)\) iterations.
\end{theorem}

\begin{proof}
    We prove the same scalar recurrence as in the population global phase, but
    for the empirical loss.  Let
    \[
        \hL_t:=\hL(\vpi^{(t)},\vmu^{(t)}),
        \qquad
        \hL_{t+1/2}:=\hL(\vpi^{(t+1)},\vmu^{(t)}).
    \]
    By \eqref{eq:empirical-weight-certificate}, \(\hL_{t+1/2}\le\hL_t\).

    We first transfer the empirical weight certificate to the population conditions \eqref{eq:robust-global-weight-kkt}--\eqref{eq:robust-global-weight-gap}. The empirical version of \cref{lem:weight-dual-gap-consequences}, followed by
    \eqref{eq:uniform-pi-gradient}, gives
    \[
        \sum_i\pi_i^{(t+1)}
        \abs{
            \nabla_{\pi_i}\L(\vpi^{(t+1)},\vmu^{(t)})+1
        }
        \le
        2\widehat G_\pi(\vpi^{(t+1)},\vmu^{(t)})+\gamma_\pi
        \le
        \frac{c}{A_\pi}\eps_0 .
    \]
    Similarly, the objective-gap consequence of the empirical dual gap and
    \eqref{eq:uniform-loss-gradient} imply
    \[
        \L(\vpi^{(t+1)},\vmu^{(t)})
        \le
        \inf_{\vpi\in\gP}\L(\vpi,\vmu^{(t)})
        +
        \eps_\pi+2\gamma_L
        \le
        \inf_{\vpi\in\gP}\L(\vpi,\vmu^{(t)})
        +
        \frac{c}{A_\pi}\eps_0 .
    \]
    Thus the post-weight point satisfies the population weight hypotheses of
    the global phase.

    We now prove the one-step empirical descent while
    \(\hL_{t+1/2}\ge \eps\).  Set
    \(L_{t+1/2}:=\L(\vpi^{(t+1)},\vmu^{(t)})\).  By
    \eqref{eq:uniform-loss-gradient}, \(L_{t+1/2}\asymp\hL_{t+1/2}\).
    Using \cref{lem:global-full-potential-linear}, replacing \(\nabla_\vmu\L\) by \(\nabla_\vmu\hL\) costs at most
    \(\gamma_\mu\sqrt{U(\vmu^{(t)})}\).  Hence, as long as
    \(U(\vmu^{(t)})\le B^2/2\),
    \[
    \begin{aligned}
        \sum_{\ell=1}^m\sum_{i\in S_\ell}
        \left\langle
            \nabla_{\vmu_i}\hL(\vpi^{(t+1)},\vmu^{(t)}),
            \vmu_i^{(t)}-\vmu_\ell^*
        \right\rangle
        &\ge
        cL_{t+1/2}-\gamma_\mu\sqrt{U(\vmu^{(t)})}
        \ge
        c\hL_{t+1/2},
    \end{aligned}
    \]
    where the last step uses \(\gamma_\mu\le c\eps_0/B\) and
    \(\eps\ge\eps_0\).

    Let \(\widehat{\vg}_t:=
    \nabla_\vmu\hL(\vpi^{(t+1)},\vmu^{(t)})\).  Cauchy--Schwarz gives
    \(\norm{\widehat{\vg}_t}_F\gtrsim \hL_{t+1/2}/B\).  Applying
    \eqref{eq:uniform-empirical-smoothness} to
    \(\vdelta=-\eta\widehat{\vg}_t\), and using \(\eta\le c/A_\eta\), yields
    \[
        \hL(\vpi^{(t+1)},\vmu^{(t+1)})
        \le
        \hL_{t+1/2}
        -
        \frac{\eta}{2}\norm{\widehat{\vg}_t}_F^2
        \le
        \hL_{t+1/2}
        -
        \frac{c\eta}{B^2}\hL_{t+1/2}^2 .
    \]

    The potential estimate is the same projection calculation.  Evaluating the
    updated point using the old assignment,
    \[
    \begin{aligned}
        U(\vmu^{(t+1)})
        &\le
        U(\vmu^{(t)})
        -
        2\eta
        \sum_{\ell=1}^m\sum_{i\in S_\ell}
        \left\langle
            \widehat{\vg}_{t,i},
            \vmu_i^{(t)}-\vmu_\ell^*
        \right\rangle
        +
        \eta^2\norm{\widehat{\vg}_t}_F^2
        \le
        U(\vmu^{(t)})
        -
        c\eta\hL_{t+1/2}
        +
        \eta^2\norm{\widehat{\vg}_t}_F^2 .
    \end{aligned}
    \]
    The empirical self-bound
    \(\norm{\widehat{\vg}_t}_F^2\le C A_\eta\hL_{t+1/2}\) follows from the
    population self-bound, \eqref{eq:uniform-loss-gradient}, and
    \(\hL_{t+1/2}\ge \eps\).  Thus \(U(\vmu^{(t+1)})\le U(\vmu^{(t)})\) by \(\eta\le c/A_\eta\).  This finishes the induction.

    Combining the mean-step recurrence with empirical monotonicity of the
    weight step gives
    \[
        \hL_{t+1}
        \le
        \hL_t
        -
        \frac{c\eta}{B^2}\hL_t^2
    \]
    until \(\hL_t\le \eps\), after changing constants in the usual
    two-case way.  Solving this scalar recurrence gives
    \(T\le CB^2/(\eta\eps)\).  At the stopping time,
    \eqref{eq:uniform-loss-gradient} gives
    \(\L(\vpi^{(T)},\vmu^{(T)})\le \eps\), since
    \(\gamma_L\le c\eps_{\rm stat}\le c\eps\).
\end{proof}

\subsection{Fixed-sample local phase}

We next prove the finite-sample analogue of the local phase. The only new work is to show
that the empirical close-component projection remains positive after replacing
\(\nabla_\vmu\L\) by \(\nabla_\vmu\hL\).

\begin{theorem}[Fixed-sample local phase]
\label{thm:fixed-sample-local}
    Under the assumptions of \cref{thm: local main}, suppose that \cref{lem:uniform-empirical-event} holds and at time \(T_1\) and target error $\eps<\eps_0$,
    \[
        \L(\vpi^{(T_1)},\vmu^{(T_1)})\le C\eps_0,
        \qquad
        U(\vmu^{(T_1)})\le \frac{B^2}{2}.
    \]
    Then \cref{alg: fixed sample} reaches an iterate satisfying
    \(\L(\vpi^{(T)},\vmu^{(T)})\le \eps\) after
    \[
        T-T_1
        \le
        \frac{C A_{\rm rate}}{\eta}
        \sqrt{\frac1\eps-\frac1{\eps_0}}
    \]
    further iterations.  Moreover, \(U(\vmu^{(t)})\le B^2\) throughout the
    local phase.
\end{theorem}

\begin{proof}
    Let
    \[
        \hL_t:=\hL(\vpi^{(t)},\vmu^{(t)}),
        \qquad
        \hL_{t+1/2}:=\hL(\vpi^{(t+1)},\vmu^{(t)}).
    \]
    Again \(\hL_{t+1/2}\le \hL_t\).  We prove the one-step recurrence until
    \(\hL_{t+1/2}\le C\eps\).  In the complementary regime
    \(\hL_{t+1/2}\ge C\eps\), the uniform loss bound gives
    \(L_{t+1/2}:=\L(\vpi^{(t+1)},\vmu^{(t)})\asymp\hL_{t+1/2}\).

    The empirical dual-gap condition transfers to the population weighted KKT
    residual exactly as in the global phase:
    \[
        \sum_i\pi_i^{(t+1)}
        \abs{
            \nabla_{\pi_i}\L(\vpi^{(t+1)},\vmu^{(t)})+1
        }
        \le
        2\eps_\pi+\gamma_\pi
        \le
        c_{\rm kkt}L_{t+1/2}.
    \]
    Thus the population local close-component projection estimate applies at
    the post-weight point.

    Let $\deltacls
        :=
        2\sqrt{\frac{\Xi_{\rm id}}{\pimins}}\,L_{t+1/2}^{1/4}$ and $\tldpi_\ell
        :=
        \sum_{i\in S_\ell(\deltacls)}\pi_i^{(t+1)}$.
    Define the empirical close-component projection
    \[
        \widehat{\mathcal D}_t
        :=
        \sum_{\ell=1}^m
        \sum_{i\in S_\ell(\deltacls)}
        \frac{\pi_\ell^*}{\tldpi_\ell}
        \left\langle
            \nabla_{\vmu_i}\hL(\vpi^{(t+1)},\vmu^{(t)}),
            \vmu_i^{(t)}-\vmu_\ell^*
        \right\rangle .
    \]
    The population version of this projection is at least \(cL_{t+1/2}\).
    The error from replacing \(\nabla_\vmu\L\) by \(\nabla_\vmu\hL\) is
    bounded by
    \[
    \begin{aligned}
        \gamma_\mu
        \left(
        \sum_{\ell=1}^m
        \sum_{i\in S_\ell(\deltacls)}
        \left(\frac{\pi_\ell^*}{\tldpi_\ell}\right)^2
        \norm{\vmu_i^{(t)}-\vmu_\ell^*}^2
        \right)^{1/2}
        &\lesssim
        \gamma_\mu\sqrt{A_{\rm rate}}\,
        L_{t+1/2}^{1/4}
        \le
        cL_{t+1/2}.
    \end{aligned}
    \]
    The last inequality uses \eqref{eq:sample-gamma-choices} and
    \(L_{t+1/2}\gtrsim\eps\).  Therefore
    \(\widehat{\mathcal D}_t\ge cL_{t+1/2}\ge c\hL_{t+1/2}\).

    By Cauchy--Schwarz and the same close-component direction bound,
    \[
        \norm{
            \nabla_\vmu\hL(\vpi^{(t+1)},\vmu^{(t)})
        }_F
        \ge
        cA_{\rm rate}^{-1/2}\hL_{t+1/2}^{3/4}.
    \]
    Applying empirical smoothness to the mean step similarly the two-case analysis gives
    \[
        \hL(\vpi^{(t+1)},\vmu^{(t+1)})
        \le
        \hL_t
        -
        \frac{c\eta}{A_{\rm rate}}\hL_t^{3/2}
    \]
    until \(\hL_t\le C\eps\).

    It remains to preserve the trajectory bound.  Let
    \(\widehat{\vg}_t:=\nabla_\vmu\hL(\vpi^{(t+1)},\vmu^{(t)})\).  The
    empirical descent estimate and the empirical gradient lower bound imply
    the same step-length estimate as in the population local proof:
    \[
        \eta\norm{\widehat{\vg}_t}_F
        \le
        C\sqrt{A_{\rm rate}}\,
        \frac{
            \hL_{t+1/2}
            -
            \hL(\vpi^{(t+1)},\vmu^{(t+1)})
        }{
            \hL_{t+1/2}^{3/4}
        } .
    \]
    Summing this inequality over the local phase and using monotonicity of
    \(\hL\), we get
    \[
    \begin{aligned}
        \sqrt{U(\vmu^{(t)})}
        &\le
        \sqrt{U(\vmu^{(T_1)})}
        +
        C\sqrt{A_{\rm rate}}
        \sum_{s=T_1}^{t-1}
        \frac{
            \hL_{s+1/2}
            -
            \hL(\vpi^{(s+1)},\vmu^{(s+1)})
        }{
            \hL_{s+1/2}^{3/4}
        }
        \le
        \frac{B}{\sqrt2}
        +
        C\sqrt{A_{\rm rate}}\eps_0^{1/4}
        \le B .
    \end{aligned}
    \]
    The final inequality is the local smallness condition from
    \cref{thm: local main}.  Thus the trajectory remains in the concentration
    class.

    Solving the local scalar recurrence gives
    \[
        \hL_t
        \le
        \frac{C A_{\rm rate}^2}
        {\eta^2(t-T_1)^2+A_{\rm rate}^2/\eps_0}.
    \]
    Hence \(\hL_t\le C\eps\) after the claimed number of local iterations.
    At this time, \eqref{eq:uniform-loss-gradient} gives
    \(\L(\vpi^{(t)},\vmu^{(t)})\le C\eps\), because
    \(\gamma_L\le c\eps\) in the local regime.
\end{proof}

\subsection{Proof of \cref{thm: sample complexity}}

\begin{proof}[Proof of \cref{thm: sample complexity}]
    We consider the initialization event from
    \cref{lem: init} and the event in
    \cref{lem:uniform-empirical-event}, with failure probability \(\delta/2\)
    and with \(\gamma_L,\gamma_\pi,\gamma_\mu\) chosen as in
    \eqref{eq:sample-gamma-choices}.  
    We work on this event.
    Choose \(B^2\asymp dn\) and
    \(L_0\asymp d+\log n\) so that
    \(U(\vmu^{(0)})\le B^2/2\) and
    \(\L(\vpi^{(0)},\vmu^{(0)})\le L_0\).  By
    \eqref{eq:uniform-loss-gradient}, after increasing \(L_0\) by a universal
    constant, \(\hL(\vpi^{(0)},\vmu^{(0)})\le L_0+1\).

    If \(\eps\ge\eps_0\), apply \cref{thm:fixed-sample-global} with a sufficiently small constant multiple of \(\eps\).
    This gives \(\L(\vpi^{(T)},\vmu^{(T)})\le \eps\) within
    \(T=O(dn/(\eta\eps))\) iterations.

    If \(\eps<\eps_0\), first apply \cref{thm:fixed-sample-global} with \(\eps_0\), and then
    \cref{thm:fixed-sample-local}, again with the target replaced by a
    sufficiently small constant multiple of \(\eps\), gives
    \[
        \L(\vpi^{(T)},\vmu^{(T)})\le \eps,
        \qquad
        T-T_1
        \le
        \frac{C A_{\rm rate}}{\eta}
        \sqrt{\frac1\eps-\frac1{\eps_0}} .
    \]
    This proves the finite-sample loss and iteration guarantees.

    The recovery statement follows exactly as in the population proof. 
\end{proof}

\subsection{Proof of the uniform empirical event}

We prove \cref{lem:uniform-empirical-event} by a truncated finite-net
argument.  Throughout this proof, for a measurable function \(f\), write
\[
    Pf:=\E_{\vx\sim p^*}f(\vx),
    \qquad
    P_Nf:=\frac1N\sum_{a=1}^N f(\vx_a).
\]
For a scalar random variable \(Y\) under \(P\), let
\[
    \norm{Y}_{\psi_1(P)}
    :=
    \inf\left\{
        t>0:
        P\exp\left(\frac{|Y|}{t}\right)\le 2
    \right\}.
\]

\begin{lemma}[Uniform Bernstein bound via a truncated finite net]
\label{lem:uniform-bernstein-tool}
    Let \(\Theta\) be a parameter set with metric \(d_\Theta\), and let
    \(\mathcal F=\{f_\theta:\theta\in\Theta\}\) be a class of measurable
    functions under \(P=p^*\).  Suppose that
    \(h_\theta:=f_\theta-Pf_\theta\) satisfies
    \[
        \sup_{\theta\in\Theta}\norm{h_\theta}_{\psi_1(P)}\le K .
    \]
    Let \(\mathcal E_x\) be an event in the sample space.  Suppose that, with
    probability at least \(1-\delta/4\), all samples
    \(\vx_1,\ldots,\vx_N\) lie in \(\mathcal E_x\), and that on
    \(\mathcal E_x\),
    \[
        \abs{f_\theta(\vx)-f_{\theta'}(\vx)}
        \le
        L d_\Theta(\theta,\theta')
        \qquad
        \text{for all }\theta,\theta'\in\Theta .
    \]
    Suppose also that the population tail error satisfies
    \[
        \sup_{\theta\in\Theta}
        P\left[
            \abs{f_\theta(\vx)}
            \mathbf 1_{\mathcal E_x^c}
        \right]
        \le \frac{\gamma}{64}.
    \]
    Then, for every \(0<\gamma\le K\), with probability at least
    \(1-\delta\),
    \[
        \sup_{\theta\in\Theta}
        \abs{(P_N-P)f_\theta}
        \le \gamma
    \]
    provided
    \[
        N
        \ge
        C\frac{K^2}{\gamma^2}
        \left[
            \log \mathcal N\!\left(
                \frac{\gamma}{64L},\Theta,d_\Theta
            \right)
            +
            \log\frac1\delta
        \right].
    \]
\end{lemma}

\begin{proof}
    Let \(\mathcal N\) be a \(\gamma/(64L)\)-net of \(\Theta\).  For every
    \(\theta_0\in\mathcal N\), Bernstein's inequality for sub-exponential
    random variables \citep[Theorem~2.8.1]{vershynin2018high} gives
    \[
        \P\left(
            \abs{(P_N-P)f_{\theta_0}}>\frac{\gamma}{2}
        \right)
        \le
        2\exp\left(-cN\frac{\gamma^2}{K^2}\right),
    \]
    because \(f_{\theta_0}-Pf_{\theta_0}=h_{\theta_0}\) and \(\gamma\le K\).
    The stated lower bound on \(N\) makes the union bound over the net at most
    \(\delta/2\).

    On the event that all samples lie in \(\mathcal E_x\), take any
    \(\theta\in\Theta\) and choose \(\theta_0\in\mathcal N\) with
    \(d_\Theta(\theta,\theta_0)\le \gamma/(64L)\).  Then
    \[
        P_N\abs{f_\theta-f_{\theta_0}}
        \le \frac{\gamma}{64}.
    \]
    For the population part,
    \[
    \begin{aligned}
        P\abs{f_\theta-f_{\theta_0}}
        &\le
        P\left[
            \abs{f_\theta-f_{\theta_0}}
            \mathbf 1_{\mathcal E_x}
        \right]
        +
        P\left[
            \abs{f_\theta-f_{\theta_0}}
            \mathbf 1_{\mathcal E_x^c}
        \right]
        \le
        \frac{\gamma}{64}
        +
        P\left[
            \abs{f_\theta}\mathbf 1_{\mathcal E_x^c}
        \right]
        +
        P\left[
            \abs{f_{\theta_0}}\mathbf 1_{\mathcal E_x^c}
        \right]
        \le
        \frac{3\gamma}{64}.
    \end{aligned}
    \]
    Therefore
    \[
        \abs{(P_N-P)f_\theta}
        \le
        \abs{(P_N-P)f_{\theta_0}}
        +
        P_N\abs{f_\theta-f_{\theta_0}}
        +
        P\abs{f_\theta-f_{\theta_0}}
        \le \gamma
    \]
    after adjusting constants.
\end{proof}

\begin{proof}[Proof of \cref{lem:uniform-empirical-event}]
    Set
    \[
        R_{\rm samp}:=
        1+B+\max_{\ell\in[m]}\norm{\vmu_\ell^*}
        \le
        1+B+\Delta_{\max}.
    \]
    We enlarge \(\sfA_{\rm samp}\) so that it dominates a sufficiently large
    polynomial in
    \(d,m,n,\pimins^{-1},R_{\rm samp},\eps_0^{-1}\).  On
    \(\Theta_{2B}\), every model center satisfies
    \[
        \norm{\vmu_i}\le C R_{\rm samp}.
    \]

    We apply \cref{lem:uniform-bernstein-tool} to four classes:
    \[
    \begin{aligned}
        \mathcal H_L
        &:=
        \left\{
            \ell(p_{\vpi,\vmu})-\ell(p^*):
            (\vpi,\vmu)\in\Theta_{2B}
        \right\},\\
        \mathcal H_\mu
        &:=
        \left\{
            \left\langle
                \vV,
                \nabla_\vmu\ell(p_{\vpi,\vmu})
            \right\rangle:
            (\vpi,\vmu)\in\Theta_{2B},\ \norm{\vV}_F\le1
        \right\},\\
        \mathcal H_\pi
        &:=
        \left\{
            \sum_{i=1}^n s_i
            \psi_i^{(\vpi,\vmu)}:
            (\vpi,\vmu)\in\Theta_{2B},\ s_i\in[-1,1]
        \right\},\\
        \mathcal H_{\rm Hess}
        &:=
        \left\{
            \left\langle
                \vV,
                    \nabla_{\vmu\vmu}^2\ell(p_{\vpi,\vmu})
                    \vV
            \right\rangle:
            (\vpi,\vmu)\in\Theta_{2B},\ \norm{\vV}_F\le1
        \right\}.
    \end{aligned}
    \]
    
        We first verify the envelope bounds.  For \(\mathcal H_L\), choose
    \(i_0\) with \(\pi_{i_0}\ge 1/n\).  Since
    \(\norm{\vmu_{i_0}}\le C R_{\rm samp}\),
    $p_{\vpi,\vmu}(\vx)
        \ge
        \frac1n\phi(\vmu_{i_0};\vx)$ and 
    $p^*(\vx)
        \ge
        \pimins\max_{\ell\in[m]}\phi(\vmu_\ell^*;\vx)$.
    Also \(p_{\vpi,\vmu}(\vx),p^*(\vx)\le(2\pi)^{-d/2}\).  Therefore
    \[
        \abs{\ell(p_{\vpi,\vmu}(\vx))-\ell(p^*(\vx))}
        \le
        C\left(
            d+\log\frac n{\pimins}
            +R_{\rm samp}^2+\norm{\vx}^2
        \right).
    \]
    Since \(\vx\sim p^*\) is a finite mixture of Gaussians with true means of
    norm at most \(R_{\rm samp}\), the right-hand side has
    \(\psi_1(P)\)-norm at most \(\sfA_{\rm samp}\).  Thus
    $\sup_{h\in\mathcal H_L}\norm{h}_{\psi_1(P)}
        \le \sfA_{\rm samp}.$

    For \(\mathcal H_\mu\), use
    $\nabla_{\vmu_i}\ell(p_{\vpi,\vmu}(\vx))
        =
        \psi_i(\vx)(\vmu_i-\vx).$
    If \(\norm{\vV}_F\le1\), then
    \[
    \begin{aligned}
        \abs{
            \left\langle
                \vV,
                \nabla_\vmu\ell(p_{\vpi,\vmu}(\vx))
            \right\rangle
        }
        &\le
        \left(\sum_i\psi_i(\vx)\norm{\vV_i}^2\right)^{1/2}
        \left(\sum_i\psi_i(\vx)\norm{\vmu_i-\vx}^2\right)^{1/2}
        \le C(R_{\rm samp}+\norm{\vx}).
    \end{aligned}
    \]
    Hence
        $\sup_{h\in\mathcal H_\mu}\norm{h}_{\psi_1(P)}
        \le \sfA_{\rm samp}.$

    The class \(\mathcal H_\pi\) is bounded because
    $\pi_i\nabla_{\pi_i}\ell(p_{\vpi,\vmu}(\vx))=-\psi_i(\vx)$ and 
    $\abs{\sum_i s_i\psi_i(\vx)}
        \le \sum_i\psi_i(\vx)=1 .$
    Thus \(\sup_{h\in\mathcal H_\pi}\norm{h}_\infty\le2\).

    Finally, for \(\mathcal H_{\rm Hess}\), direct differentiation gives, for
    \(\norm{\vV}_F\le1\),
    \[
        \left\langle
            \vV,\nabla_{\vmu\vmu}^2\ell(p_{\vpi,\vmu}(\vx))\vV
        \right\rangle
        =
        \left(\sum_i\psi_i(\vx)\ip{\vx-\vmu_i,\vV_i}\right)^2
        -
        \sum_i\psi_i(\vx)\ip{\vx-\vmu_i,\vV_i}^2
        +
        \sum_i\psi_i(\vx)\norm{\vV_i}^2 .
    \]
    Therefore
    \[
        \abs{
        \left\langle
            \vV,\nabla_{\vmu\vmu}^2\ell(p_{\vpi,\vmu}(\vx))\vV
        \right\rangle
        }
        \le
        C\left(1+(R_{\rm samp}+\norm{\vx})^2\right),
    \]
    and hence
    $\sup_{h\in\mathcal H_{\rm Hess}}\norm{h}_{\psi_1(P)}
        \le \sfA_{\rm samp}.$

    These envelope bounds also give
    \[
        \sup_\theta
        \norm{f_\theta-Pf_\theta}_{\psi_1(P)}
        \le
        C\sfA_{\rm samp}
    \]
    for the loss, mean-gradient, and Hessian classes, and a universal bound for
    the responsibility class.  Thus the \(\psi_1(P)\)-condition in
    \cref{lem:uniform-bernstein-tool} holds.

        We now verify the finite-net conditions in
    \cref{lem:uniform-bernstein-tool}.  Fix one of the four classes and write
    its uncentered element as \(f_\theta\).  The parameter \(\theta\) consists
    of \((\vpi,\vmu)\), together with the auxiliary direction \(\vV\) for
    \(\mathcal H_\mu,\mathcal H_{\rm Hess}\), or the auxiliary vector
    \(s\in[-1,1]^n\) for \(\mathcal H_\pi\).

    Let
    \[
        R_x^2
        :=
        C\left(
            R_{\rm samp}^2+d+
            \log\frac{N}{\delta}
            +
            \log\frac{C\sfA_{\rm samp}}{\gamma}
        \right),
        \qquad
        \mathcal E_x:=\{\norm{\vx}\le R_x\}.
    \]
    By the Gaussian norm tail bound and a union bound,
    \[
        \P\left(\vx_a\in\mathcal E_x\text{ for all }a\in[N]\right)
        \ge 1-\frac{\delta}{4}.
    \]
    Moreover, for the common envelope
    \[
        F_{\rm samp}(\vx)
        :=
        C\left(
            d+\log\frac n{\pimins}
            +R_{\rm samp}^2+\norm{\vx}^2
        \right),
    \]
    increasing \(C\) gives
    \[
        P\left[
            F_{\rm samp}(\vx)\mathbf 1_{\mathcal E_x^c}
        \right]
        \le
        \frac{\gamma}{C}.
    \]

    On \(\mathcal E_x\), every model density satisfies
    \[
        p_{\vpi,\vmu}(\vx)
        \ge
        \frac1n(2\pi)^{-d/2}
        \exp\left(
            -\frac12(R_x+CR_{\rm samp})^2
        \right).
    \]
    Hence, by differentiating the maps
    \[
        (\vpi,\vmu)\mapsto -\log p_{\vpi,\vmu}(\vx),
        \qquad
        (\vpi,\vmu)\mapsto \psi_i^{(\vpi,\vmu)}(\vx),
    \]
    and then using the displayed responsibility identities for the gradient
    and Hessian integrands, all four uncentered classes are Lipschitz on
    \(\mathcal E_x\) with a common constant
    \[
        L_x:=\exp(CR_x^2).
    \]

    The parameter spaces are covered by fixing the nearest-center assignment
    of each model component, then covering the simplex, the \(n\) bounded
    mean balls, and the auxiliary variables \(\vV\) or \(s\).  Therefore, for
    each of the four classes,
    \[
    \begin{aligned}
        \log \mathcal N\left(
            \frac{\gamma}{64L_x},\Theta_\star
        \right)
        &\le
        n\log m
        +
        C(nd+n)
        \log\frac{C R_{\rm samp}L_x}{\gamma}
        \le
        C\sfA_{\rm samp}
        \log\frac{C\sfA_{\rm samp}}{\gamma\delta}.
    \end{aligned}
    \]
    The term \(\log N\) inside \(R_x^2\) is harmless: the resulting
    requirement has the form
    \[
        N
        \ge
        C\sfA_{\rm samp}\gamma^{-2}
        \log\frac{C\sfA_{\rm samp}}{\gamma\delta},
    \]
    with large enough constant $C$.

    Applying \cref{lem:uniform-bernstein-tool} to
    \(\mathcal H_L,\mathcal H_\mu,\mathcal H_\pi\) with accuracies
    \(\gamma_L,\gamma_\mu,\gamma_\pi\), respectively, gives
    \eqref{eq:uniform-loss-gradient} and \eqref{eq:uniform-pi-gradient}.
    The mean-gradient bound follows because \(\mathcal H_\mu\) includes all
    directions \(\norm{\vV}_F\le1\).  The weight-gradient bound follows from
    \[
    \begin{aligned}
        \sum_i\pi_i
        \abs{
            \nabla_{\pi_i}\hL(\vpi,\vmu)
            -
            \nabla_{\pi_i}\L(\vpi,\vmu)
        }
        &=
        \sum_i\abs{(P_N-P)\psi_i^{(\vpi,\vmu)}}
        =
        \sup_{s_i\in[-1,1]}
        (P_N-P)\sum_i s_i\psi_i^{(\vpi,\vmu)} .
    \end{aligned}
    \]

    Applying the same lemma to \(\mathcal H_{\rm Hess}\) with constant accuracy gives, after increasing \(\sfA_{\rm samp}\),
    \[
        \sup_{(\vpi,\vmu)\in\Theta_{2B}}
        \norm{
            \nabla_{\vmu\vmu}^2\hL(\vpi,\vmu)
            -
            \nabla_{\vmu\vmu}^2\L(\vpi,\vmu)
        }_{\rm op}
        \le A_\eta .
    \]
    Therefore
    \[
        \sup_{(\vpi,\vmu)\in\Theta_{2B}}
        \norm{\nabla_{\vmu\vmu}^2\hL(\vpi,\vmu)}_{\rm op}
        \le \sup_{(\vpi,\vmu)\in\Theta_{2B}}
        \norm{\nabla_{\vmu\vmu}^2\L(\vpi,\vmu)}_{\rm op}
        + A_\eta
        \le 2A_\eta .
    \]
    Taylor's theorem along any segment contained in \(\Theta_{2B}\), followed by
    enlarging constants, gives \eqref{eq:uniform-empirical-smoothness}.

    The sample-size condition in \cref{lem:uniform-empirical-event} is the
    maximum of the four applications above, after absorbing polynomial
    factors into \(\sfA_{\rm samp}\).
\end{proof}

\bibliography{ref}

@article{balakrishnan2017statistical,
  title={STATISTICAL GUARANTEES FOR THE EM ALGORITHM: FROM POPULATION TO SAMPLE-BASED ANALYSIS},
  author={Balakrishnan, Sivaraman and Wainwright, Martin J. and Yu, Bin},
  journal={The Annals of Statistics},
  volume={45},
  number={1},
  pages={77--120},
  year={2017}
}

@InProceedings{daskalakis17TenSteps,
  title = 	 {Ten Steps of {EM} Suffice for Mixtures of Two Gaussians},
  author = 	 {Daskalakis, Constantinos and Tzamos, Christos and Zampetakis, Manolis},
  booktitle = 	 {Proceedings of the 2017 Conference on Learning Theory},
  pages = 	 {704--710},
  year = 	 {2017},
  editor = 	 {Kale, Satyen and Shamir, Ohad},
  volume = 	 {65},
  series = 	 {Proceedings of Machine Learning Research},
  month = 	 {07--10 Jul},
  publisher =    {PMLR},
}

@article{wu2019randomly,
  title={Randomly initialized {EM} algorithm for two-component Gaussian mixture achieves near optimality in $O(\sqrt{n})$ iterations},
  author={Wu, Yihong and Zhou, Harrison H},
  journal={Mathematical Statistics and Learning},
  volume={4},
  number={3},
  year={2021}
}

@article{Weinberger2021TheEA,
  title={The {EM} Algorithm is Adaptively-Optimal for Unbalanced Symmetric Gaussian Mixtures},
  author={Nir Weinberger and Guy Bresler},
  journal={J. Mach. Learn. Res.},
  year={2021},
  volume={23},
  pages={103:1-103:79},
}

@inproceedings{Jin2016LocalMI,
  title={Local Maxima in the Likelihood of Gaussian Mixture Models: Structural Results and Algorithmic Consequences},
  author={Chi Jin and Yuchen Zhang and Sivaraman Balakrishnan and Martin J. Wainwright and Michael I. Jordan},
  booktitle={Neural Information Processing Systems},
  year={2016},
}

@article{Dwivedi2018SingularityMA,
  title={Singularity, misspecification and the convergence rate of {EM}},
  author={Raaz Dwivedi and Nhat Ho and Koulik Khamaru and Michael I. Jordan and Martin J. Wainwright and Bin Yu},
  journal={The Annals of Statistics},
  year={2018},
}

@inproceedings{dasgupta2013two,
author = {Dasgupta, Sanjoy and Schulman, Leonard J.},
title = {A Two-Round Variant of {EM} for Gaussian Mixtures},
year = {2000},
isbn = {1558607099},
publisher = {Morgan Kaufmann Publishers Inc.},
address = {San Francisco, CA, USA},
booktitle = {Proceedings of the 16th Conference on Uncertainty in Artificial Intelligence},
pages = {152–159},
numpages = {8},
series = {UAI '00}
}

@inproceedings{Xu2016GlobalAO,
  title={Global Analysis of Expectation Maximization for Mixtures of Two Gaussians},
  author={Ji Xu and Daniel J. Hsu and Arian Maleki},
  booktitle={Neural Information Processing Systems},
  year={2016},
}

@inproceedings{Dwivedi2019SharpAO,
  title={Sharp Analysis of Expectation-Maximization for Weakly Identifiable Models},
  author={Raaz Dwivedi and Nhat Ho and Koulik Khamaru and Martin J. Wainwright and Michael I. Jordan and Bin Yu},
  booktitle={International Conference on Artificial Intelligence and Statistics},
  year={2019},
}

@inproceedings{Dwivedi2018TheoreticalGF,
  title={Theoretical guarantees for EM under misspecified Gaussian mixture models},
  author={Raaz Dwivedi and Nhat Ho and Koulik Khamaru and Martin J. Wainwright and Michael I. Jordan},
  booktitle={Neural Information Processing Systems},
  year={2018},
}

@article{chen2024local,
  title={Local minima structures in Gaussian mixture models},
  author={Chen, Yudong and Song, Dogyoon and Xi, Xumei and Zhang, Yuqian},
  journal={IEEE Transactions on Information Theory},
  year={2024},
  publisher={IEEE}
}

@InProceedings{kwon_em_2020,
  title = 	 {The {EM} Algorithm gives Sample-Optimality for Learning Mixtures of Well-Separated Gaussians},
  author =       {Kwon, Jeongyeol and Caramanis, Constantine},
  booktitle = 	 {Proceedings of Thirty Third Conference on Learning Theory},
  pages = 	 {2425--2487},
  year = 	 {2020},
  editor = 	 {Abernethy, Jacob and Agarwal, Shivani},
  volume = 	 {125},
  series = 	 {Proceedings of Machine Learning Research},
  month = 	 {09--12 Jul},
  publisher =    {PMLR},
}

@inproceedings{yan_convergence_2017,
 author = {Yan, Bowei and Yin, Mingzhang and Sarkar, Purnamrita},
 booktitle = {Advances in Neural Information Processing Systems},
 editor = {I. Guyon and U. Von Luxburg and S. Bengio and H. Wallach and R. Fergus and S. Vishwanathan and R. Garnett},
 pages = {},
 publisher = {Curran Associates, Inc.},
 title = {Convergence of Gradient EM on Multi-component Mixture of Gaussians},
 volume = {30},
 year = {2017}
}

@article{segol_improved_2021,
  title={Improved convergence guarantees for learning Gaussian mixture models by EM and gradient EM},
  author={Segol, Nimrod and Nadler, Boaz},
  journal={Electronic journal of statistics},
  volume={15},
  number={2},
  pages={4510--4544},
  year={2021},
  publisher={The Institute of Mathematical Statistics and the Bernoulli Society}
}

@article{zhao2020statistical,
  title={Statistical convergence of the EM algorithm on Gaussian mixture models},
  author={Zhao, Ruofei and Li, Yuanzhi and Sun, Yuekai},
  journal={Electronic Journal of Statistics},
  volume={14},
  pages={632--660},
  year={2020}
}

@article{nishiyama2020relations,
  title={On relations between the relative entropy and $\chi$ 2-divergence, generalizations and applications},
  author={Nishiyama, Tomohiro and Sason, Igal},
  journal={Entropy},
  volume={22},
  number={5},
  pages={563},
  year={2020},
  publisher={MDPI}
}

@article{ryu2016primer,
  title={Primer on monotone operator methods},
  author={Ryu, Ernest K and Boyd, Stephen},
  journal={Appl. comput. math},
  volume={15},
  number={1},
  pages={3--43},
  year={2016}
}

@article{anandkumar2014tensor,
  author  = {Animashree Anandkumar and Rong Ge and Daniel Hsu and Sham M. Kakade and Matus Telgarsky},
  title   = {Tensor Decompositions for Learning Latent Variable Models},
  journal = {Journal of Machine Learning Research},
  year    = {2014},
  volume  = {15},
  number  = {80},
  pages   = {2773--2832},
}

@article{stein1981estimation,
  title={Estimation of the mean of a multivariate normal distribution},
  author={Stein, Charles M},
  journal={The Annals of Statistics},
  pages={1135--1151},
  year={1981},
  publisher={JSTOR}
}

@inproceedings{
xu2024toward,
title={Toward Global Convergence of Gradient {EM} for Over-Paramterized Gaussian Mixture Models},
author={Weihang Xu and Maryam Fazel and Simon Shaolei Du},
booktitle={The Thirty-eighth Annual Conference on Neural Information Processing Systems},
year={2024},
}

@inproceedings{zhou2021local,
  title={A local convergence theory for mildly over-parameterized two-layer neural network},
  author={Zhou, Mo and Ge, Rong and Jin, Chi},
  booktitle={Conference on Learning Theory},
  pages={4577--4632},
  year={2021},
  organization={PMLR}
}

@inproceedings{
zhou2024how,
title={How does Gradient Descent Learn Features --- A Local Analysis for Regularized Two-Layer Neural Networks},
author={Mo Zhou and Rong Ge},
booktitle={The Thirty-eighth Annual Conference on Neural Information Processing Systems},
year={2024},
}

@inproceedings{xu2023over,
  title={Over-parameterization exponentially slows down gradient descent for learning a single neuron},
  author={Xu, Weihang and Du, Simon},
  booktitle={The Thirty Sixth Annual Conference on Learning Theory},
  pages={1155--1198},
  year={2023},
  organization={PMLR}
}

@article{gretton2012kernel,
  title={A kernel two-sample test},
  author={Gretton, Arthur and Borgwardt, Karsten M and Rasch, Malte J and Sch{\"o}lkopf, Bernhard and Smola, Alexander},
  journal={The Journal of Machine Learning Research},
  volume={13},
  number={1},
  pages={723--773},
  year={2012},
  publisher={JMLR. org}
}

@inproceedings{hsu2012learningmixturessphericalgaussians,
  title={Learning mixtures of spherical gaussians: moment methods and spectral decompositions},
  author={Hsu, Daniel and Kakade, Sham M},
  booktitle={Proceedings of the 4th conference on Innovations in Theoretical Computer Science},
  pages={11--20},
  year={2013}
}

@inproceedings{liu2022clustering,
  title={Clustering mixtures with almost optimal separation in polynomial time},
  author={Liu, Allen and Li, Jerry},
  booktitle={Proceedings of the 54th Annual ACM SIGACT Symposium on Theory of Computing},
  pages={1248--1261},
  year={2022}
}

@inproceedings{hopkins2017mixturemodelsrobustnesssum,
  title={Mixture models, robustness, and sum of squares proofs},
  author={Hopkins, Samuel B and Li, Jerry},
  booktitle={Proceedings of the 50th Annual ACM SIGACT Symposium on Theory of Computing},
  pages={1021--1034},
  year={2018}
}

@article{marion2023leveraging,
  title={Leveraging the two-timescale regime to demonstrate convergence of neural networks},
  author={Marion, Pierre and Berthier, Rapha{\"e}l},
  journal={Advances in Neural Information Processing Systems},
  volume={36},
  pages={64996--65029},
  year={2023}
}

@inproceedings{kalai2010efficiently,
  title={Efficiently learning mixtures of two Gaussians},
  author={Kalai, Adam Tauman and Moitra, Ankur and Valiant, Gregory},
  booktitle={Proceedings of the forty-second ACM symposium on Theory of computing},
  pages={553--562},
  year={2010}
}

@inproceedings{
takakura2024meanfield,
title={Mean-field Analysis on Two-layer Neural Networks from a Kernel Perspective},
author={Shokichi Takakura and Taiji Suzuki},
booktitle={Forty-first International Conference on Machine Learning},
year={2024},
}

@article{berthier2024learning,
  title={Learning time-scales in two-layers neural networks},
  author={Berthier, Rapha{\"e}l and Montanari, Andrea and Zhou, Kangjie},
  journal={Foundations of Computational Mathematics},
  pages={1--84},
  year={2024},
  publisher={Springer}
}

@book{borkar2008stochastic,
  title={Stochastic approximation: a dynamical systems viewpoint},
  author={Borkar, Vivek S},
  volume={9},
  year={2008},
  publisher={Springer}
}

@article{barboni2025ultra,
  title={Ultra-fast feature learning for the training of two-layer neural networks in the two-timescale regime},
  author={Barboni, Rapha{\"e}l and Peyr{\'e}, Gabriel and Vialard, Fran{\c{c}}ois-Xavier},
  journal={arXiv preprint arXiv:2504.18208},
  year={2025}
}

@article{beck2003mirror,
  title={Mirror descent and nonlinear projected subgradient methods for convex optimization},
  author={Beck, Amir and Teboulle, Marc},
  journal={Operations Research Letters},
  volume={31},
  number={3},
  pages={167--175},
  year={2003},
  publisher={Elsevier}
}

@inproceedings{jaggi2013revisiting,
  title={Revisiting Frank-Wolfe: Projection-free sparse convex optimization},
  author={Jaggi, Martin},
  booktitle={International conference on machine learning},
  pages={427--435},
  year={2013},
  organization={PMLR}
}

@inproceedings{ge2015learning,
  title={Learning mixtures of gaussians in high dimensions},
  author={Ge, Rong and Huang, Qingqing and Kakade, Sham M},
  booktitle={Proceedings of the forty-seventh annual ACM symposium on Theory of computing},
  pages={761--770},
  year={2015}
}

@inproceedings{regev2017learning,
  title={On learning mixtures of well-separated gaussians},
  author={Regev, Oded and Vijayaraghavan, Aravindan},
  booktitle={2017 IEEE 58th Annual Symposium on Foundations of Computer Science (FOCS)},
  pages={85--96},
  year={2017},
  organization={IEEE}
}

@book{liao2012homotopy,
  title={Homotopy analysis method in nonlinear differential equations},
  author={Liao, Shijun},
  volume={153},
  year={2012},
  publisher={Springer}
}

@inproceedings{kane2021robust,
  title={Robust learning of mixtures of gaussians},
  author={Kane, Daniel M},
  booktitle={Proceedings of the 2021 ACM-SIAM Symposium on Discrete Algorithms (SODA)},
  pages={1246--1258},
  year={2021},
  organization={SIAM}
}

@article{liu2023robustly,
  title={Robustly learning general mixtures of gaussians},
  author={Liu, Allen and Moitra, Ankur},
  journal={Journal of the ACM},
  volume={70},
  number={3},
  pages={1--53},
  year={2023},
  publisher={ACM New York, NY}
}

@inproceedings{dasgupta1999learning,
  title={Learning mixtures of Gaussians},
  author={Dasgupta, Sanjoy},
  booktitle={40th Annual Symposium on Foundations of Computer Science (Cat. No. 99CB37039)},
  pages={634--644},
  year={1999},
  organization={IEEE}
}

@inproceedings{kothari2018robust,
  title={Robust moment estimation and improved clustering via sum of squares},
  author={Kothari, Pravesh K and Steinhardt, Jacob and Steurer, David},
  booktitle={Proceedings of the 50th Annual ACM SIGACT Symposium on Theory of Computing},
  pages={1035--1046},
  year={2018}
}

@inproceedings{AndersonBGRV14,
  author       = {Joseph Anderson and
                  Mikhail Belkin and
                  Navin Goyal and
                  Luis Rademacher and
                  James R. Voss},
  title        = {The More, the Merrier: the Blessing of Dimensionality for Learning
                  Large Gaussian Mixtures},
  booktitle    = {Proceedings of 27th Conference on Learning Theory ({COLT}), Barcelona, Spain},
  volume       = {35},
  pages        = {1135--1164},
  publisher    = {JMLR.org},
  year         = {2014},
}

@article{wu1983convergence,
  title={On the convergence properties of the EM algorithm},
  author={Wu, CF Jeff},
  journal={The Annals of statistics},
  pages={95--103},
  year={1983},
  publisher={JSTOR}
}

@article{xu1996convergence,
  title={On convergence properties of the {EM} algorithm for Gaussian mixtures},
  author={Xu, Lei and Jordan, Michael I},
  journal={Neural computation},
  volume={8},
  number={1},
  pages={129--151},
  year={1996},
  publisher={MIT Press One Rogers Street, Cambridge, MA 02142-1209, USA journals-info~…}
}

@article{dempster1977maximum,
  title={Maximum likelihood from incomplete data via the EM algorithm},
  author={Dempster, Arthur P and Laird, Nan M and Rubin, Donald B},
  journal={Journal of the royal statistical society: series B (methodological)},
  volume={39},
  number={1},
  pages={1--22},
  year={1977},
  publisher={Wiley Online Library}
}

@inproceedings{srebro2006investigation,
  title={An investigation of computational and informational limits in gaussian mixture clustering},
  author={Srebro, Nathan and Shakhnarovich, Gregory and Roweis, Sam},
  booktitle={Proceedings of the 23rd international conference on Machine learning},
  pages={865--872},
  year={2006}
}

@article{pearson1894contributions,
  title={Contributions to the mathematical theory of evolution},
  author={Pearson, Karl},
  journal={Philosophical Transactions of the Royal Society of London. A},
  volume={185},
  pages={71--110},
  year={1894},
  publisher={JSTOR}
}

@article{titterington1985statistical,
  title={Statistical analysis of finite mixture distributions},
  author={Titterington, David Michael and Smith, Adrian FM and Makov, Udi E},
  journal={Wiley},
  year={1985}
}

@inproceedings{bhaskara2014smoothed,
  title={Smoothed analysis of tensor decompositions},
  author={Bhaskara, Aditya and Charikar, Moses and Moitra, Ankur and Vijayaraghavan, Aravindan},
  booktitle={Proceedings of the forty-sixth annual ACM symposium on Theory of computing},
  pages={594--603},
  year={2014}
}

@inproceedings{moitra2010settling,
  title={Settling the polynomial learnability of mixtures of gaussians},
  author={Moitra, Ankur and Valiant, Gregory},
  booktitle={2010 IEEE 51st Annual Symposium on Foundations of Computer Science},
  pages={93--102},
  year={2010},
  organization={IEEE}
}

@inproceedings{sanjeev2001learning,
  title={Learning mixtures of arbitrary Gaussians},
  author = {Arora, Sanjeev and Kannan, Ravi},
  booktitle={Proceedings of the thirty-third annual ACM symposium on Theory of computing},
  pages={247--257},
  year={2001}
}

@article{vempala2004spectral,
  title={A spectral algorithm for learning mixture models},
  author={Vempala, Santosh and Wang, Grant},
  journal={Journal of Computer and System Sciences},
  volume={68},
  number={4},
  pages={841--860},
  year={2004},
  publisher={Elsevier}
}

@inproceedings{achlioptas2005spectral,
  title={On spectral learning of mixtures of distributions},
  author={Achlioptas, Dimitris and McSherry, Frank},
  booktitle={International Conference on Computational Learning Theory},
  pages={458--469},
  year={2005},
  organization={Springer}
}

@inproceedings{brubaker2008isotropic,
  title={Isotropic PCA and affine-invariant clustering},
  author={Brubaker, Spencer Charles and Vempala, Santosh},
  booktitle={2008 49th Annual IEEE Symposium on Foundations of Computer Science},
  pages={551--560},
  year={2008},
  organization={IEEE}
}

@inproceedings{diakonikolas2018list,
  title={List-decodable robust mean estimation and learning mixtures of spherical gaussians},
  author={Diakonikolas, Ilias and Kane, Daniel M and Stewart, Alistair},
  booktitle={Proceedings of the 50th Annual ACM SIGACT Symposium on Theory of Computing},
  pages={1047--1060},
  year={2018}
}

@inproceedings{belkin2010polynomial,
  title={Polynomial learning of distribution families},
  author={Belkin, Mikhail and Sinha, Kaushik},
  booktitle={2010 IEEE 51st Annual Symposium on Foundations of Computer Science},
  pages={103--112},
  year={2010},
  organization={IEEE}
}

@inproceedings{hardt2015tight,
  title={Tight bounds for learning a mixture of two gaussians},
  author={Hardt, Moritz and Price, Eric},
  booktitle={Proceedings of the forty-seventh annual ACM symposium on Theory of computing},
  pages={753--760},
  year={2015}
}

@inproceedings{diakonikolas2020small,
  title={Small covers for near-zero sets of polynomials and learning latent variable models},
  author={Diakonikolas, Ilias and Kane, Daniel M},
  booktitle={2020 IEEE 61st Annual Symposium on Foundations of Computer Science (FOCS)},
  pages={184--195},
  year={2020},
  organization={IEEE}
}

@article{diakonikolas2024implicit,
  title={Implicit high-order moment tensor estimation and learning latent variable models},
  author={Diakonikolas, Ilias and Kane, Daniel M},
  journal={arXiv preprint arXiv:2411.15669},
  year={2024}
}

@book{vershynin2018high,
  title={High-dimensional probability: An introduction with applications in data science},
  author={Vershynin, Roman},
  volume={47},
  year={2018},
  publisher={Cambridge university press}
}

\end{document}